\pdfoutput=1

\documentclass{article}

\usepackage[final]{ACL2025}

\usepackage{times}
\usepackage{latexsym}

\usepackage{amsfonts}       
\usepackage{nicefrac}       
\usepackage{microtype}      
\usepackage[table]{xcolor}         

\usepackage{makecell}
\usepackage{epsfig}
\usepackage{graphicx}
\usepackage{amsmath}
\usepackage{amssymb}
\usepackage{subfig}
\usepackage{multirow}
\usepackage{array}
\usepackage{soul}
\usepackage{transparent}
\usepackage{tabularx}
\usepackage{tabularray}
\UseTblrLibrary{booktabs}
\usepackage{longtable}
\usepackage{algorithmic}
\usepackage[ruled,lined,linesnumbered,commentsnumbered]{algorithm2e}
\DeclareMathOperator*{\argmax}{argmax}

\usepackage{url}
\usepackage{subcaption}

\definecolor{lightgray}{gray}{0.95}
\definecolor{ADD}{rgb}{0, 0, 0}
\definecolor{ADD2}{rgb}{0, 0, 0}

\newcommand\eg{\emph{e.g}.} 

\newcommand\ie{\emph{i.e}.}

\SetKwComment{Comment}{// }{}

\makeatletter
\newcommand\myfont{\@setfontsize\myfont{9.2}{9.5}}
\newcommand\fontwo{\@setfontsize\fontwo{8.2}{10.5}}
\newcommand\fonthree{\@setfontsize\fontwo{9.2}}
\makeatother

\usepackage[T1]{fontenc}

\usepackage[utf8]{inputenc}

\usepackage{microtype}

\usepackage{inconsolata}

\usepackage{graphicx}

\title{Sample-Efficient Human Evaluation of Large Language Models via \\Maximum Discrepancy Competition}

\author{Kehua Feng$^{1,2*}$, Keyan Ding$^{1,2}$\thanks{Equal contribution.}, Hongzhi Tan$^3$, Kede Ma$^{4\dag}$, Zhihua Wang$^4$,\\ \textbf{Shuangquan Guo}$^3$, \textbf{Yuzhou Cheng}$^5$, \textbf{Ge Sun}$^5$, \textbf{Guozhou Zheng}$^6$, \textbf{Qiang Zhang}$^{1,2\dag}$, \textbf{Huajun Chen}$^{1,2}$\thanks{Corresponding authors.}\\
  $^1$Zhejiang University  
  $^2$ZJU-Hangzhou Global Scientific and Technological Innovation Center\\
  $^3$Shanghai Electric Group Co., Ltd. Central Academe 
  $^4$City University of Hong Kong\\
  $^5$Shanghai Institute for Advanced Study of Zhejiang University\\
  $^6$Zhoushan-Zhejiang University Ocean Research Center\\
  \texttt{\{kehuafeng, dingkeyan, huajunsir\}@zju.edu.cn} \\
}

\begin{document}
\maketitle
\begin{abstract}
Reliable evaluation of large language models (LLMs) is impeded by two key challenges: objective metrics often fail to reflect human perception of natural language, and exhaustive human labeling is prohibitively expensive. Here, we propose a sample-efficient human evaluation method for LLMs based on the principle of MAximum Discrepancy (MAD) Competition. Our method automatically and adaptively selects a compact set of input instructions that maximize semantic discrepancy between pairs of LLM responses. Human evaluators then perform 
three-alternative forced choices on these paired responses, which are aggregated into a global ranking using Elo rating. We apply our approach to compare eight widely used LLMs across four tasks: scientific knowledge understanding, mathematical reasoning, creative and functional writing, and code generation and explanation. Experimental results show that our sample-efficient evaluation method recovers  ``gold-standard'' model rankings with a handful of MAD-selected instructions, reveals respective strengths and weaknesses of each LLM,  and offers nuanced insights to guide future LLM development. 
Code is available at \url{https://github.com/weiji-Feng/MAD-Eval}.

\end{abstract}

\section{Introduction}

Since the advent of ChatGPT, there has been an unprecedented surge in the development of large language models (LLMs)~\cite{touvron2023llama,qwen,achiam2023gpt,jiang2023mistral,team2023gemini,chen2023large,azaria2024chatgpt}, driven by self-supervised pretraining~\cite{jaiswal2020survey}, supervised fine-tuning~\cite{wang2022self,vicuna2023,xu2023wizardlm} and reinforcement learning~\cite{ouyang2022training}. These models now exhibit remarkable general-purpose capabilities in language generation, understanding, and reasoning. Yet, with so many ``competitive'' LLMs emerging in rapid succession, establishing a reliable, scalable evaluation paradigm that reveals their strengths and weaknesses has become critical~\cite{guo2023survey,li2023evaluation,chang2023survey}.

Traditional evaluation relies on fixed, human-annotated benchmarks---such as MMLU~\cite{hendryckstest2021}, C-Eval~\cite{huang2023ceval} and BIG-bench~\cite{srivastava2022beyond}---and on objective metrics like BLEU~\cite{papineni-etal-2002-bleu} and ROUGE~\cite{lin2004rouge}. However, human annotation is costly and slow, and consequently, these benchmarks can only cover a narrow slice of possible task scenarios. Moreover, standard metrics frequently misalign with human perception of natural language (\eg, faithfulness, fluency, creativity, and semantic equivalence). More importantly, repeated testing on such static benchmarks may invite overfitting~\cite{schaeffer2023pretraining,zhou2023don,grattafiori2024llama}, aptly encapsulated by Goodhart’s Law~\cite{elton2004goodhart}, suggesting that LLMs learn to game specific benchmarks rather than truly improve.

\begin{figure*}[h]
\centering
\includegraphics[width=\linewidth]{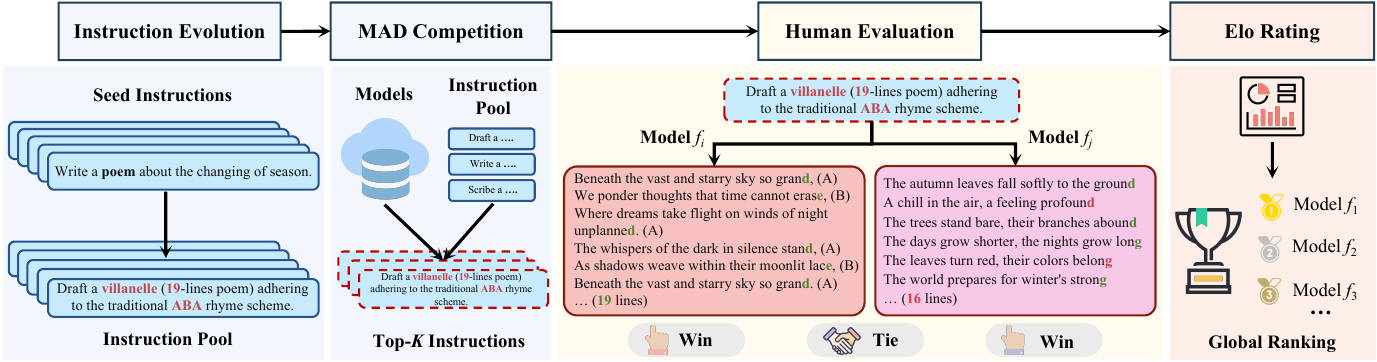}
\vspace{-0.6em}
\caption{Overview of the proposed sample-efficient human evaluation method for comparing LLMs adaptively. Starting from a small set of task-specific seed instructions, we apply an instruction evolution procedure to generate a large-scale pool of diverse instructions. For any two competing LLMs, we then conduct MAD Competition to automatically and adaptively select the top-$K$ instructions (and their corresponding responses) that most effectively distinguish model behaviors. These selected response pairs are presented to human evaluators (along with the input instruction), who express pairwise preferences. Finally, we feed these comparison outcomes into an Elo rating system to produce a global ranking of all evaluated LLMs.
}
\vspace{-0.7em}
\label{fig:pipeline}
\end{figure*}

An alternative is to enlist powerful LLMs themselves as automated judges~\cite{zhang2023wider,zeng2023evaluating}, known as LLM-as-a-judge. For instance, systems such as LIMA \cite{zhou2023lima} and AlpacaFarm \cite{dubois2023alpacafarm} leverage proprietary models like GPT-4 as the judge through API calls, while open-source evaluators, \eg, PandaLM~\cite{wang2023pandalm}, Shepherd~\cite{wang2023shepherd}, AUTO-J~\cite{li2023generative}, Prometheus~\cite{kim2023prometheus,kim2024prometheus}, SaMer~\cite{feng2025samer}, and J1~\cite{whitehouse2025j1} have also been proposed. Although these LLM-based judges offer speed and interpretability, they also come with various biases~\cite{zhu2023judgelm,wang2023large,chen2024humans} toward certain positions (\ie, position bias), specific formats (\ie, format bias), lengthy outputs (\ie, verbosity bias), polished responses (\ie, beauty bias), familiar content encountered during training (\ie, knowledge bias), or self-generated answers (\ie, self-enhancement bias). Additionally, they often struggle in specialized domains like mathematical reasoning and scientific comprehension.

Despite these advances, human evaluation remains the gold standard as natural language is created and consumed by humans. Platforms like Chatbot Arena~\cite{chiang2024chatbot} allow crowdworkers to compare model outputs directly, but they scale poorly in terms of time and cost. This raises a fundamental question: \textit{How can we automatically select the smallest, most informative set of instructions\footnote{In this paper, the terms ``instructions,'' ``prompts,'' and ``queries'' are used interchangeably to denote the inputs provided to LLMs.} from an essentially infinite pool, so that human judgments yield a definitive performance ranking of LLMs with minimal annotation effort}?

In this work, we answer this question by leveraging MAximum Discrepancy (MAD) Competition~\cite{wang2008maximum} from the fields of computational vision and software testing. Starting from a large, self-generated, and unlabeled instruction pool, our method automatically identifies a minimum set of test instructions that maximize the semantic discrepancy of pairs of LLM responses (\ie, where the models disagree the most), and presents only these potential counterexamples to human evaluators. By enforcing diversity among selected instructions, we ensure coverage of varied failure modes. Finally, we aggregate the resulting human preferences via Elo rating~\cite{elo1978rating} to a reliable global ranking. We demonstrate our sample-efficient evaluation approach across four tasks---scientific knowledge understanding, mathematical reasoning, creative and functional writing, and code generation and explanation---on eight widely used LLMs~\cite{du2022glm,ouyang2022training,achiam2023gpt,team2023gemini,xu2023wizardlm,qwen,chiang2024chatbot} under a strict human-labeling budget.

Our contributions are threefold:
\begin{itemize}\setlength\itemsep{.4em}
    \item \textbf{Sample-efficient evaluation}. We present a reliable, scalable method to automatically and adaptively select minimal yet maximally informative test instructions for LLM comparison.
    \item \textbf{Multi-dimensional benchmarking}. We apply our method across four distinct tasks, showing that reliable rankings emerge from only a handful of human judgments.
    \item \textbf{Insights into model behaviors}. By examining counterexamples that lead LLMs to fail, we reveal their relative strengths and weaknesses, offering guidance for future model improvement.
\end{itemize}

\section{Related Work}
Evaluation of LLMs has been addressed through three approaches: standardized benchmark evaluation, automated LLM-based evaluation, and human-centric evaluation. Each paradigm offers distinct insights into model capabilities, yet also bears methodological limitations.

\subsection{Standardized Benchmark Evaluation} 
A rich ecosystem of standardized benchmarks has emerged to quantify LLM performance across knowledge, instruction-following, dialogue, preference alignment, and safety dimensions. Core knowledge evaluation, such as MMLU \cite{hendryckstest2021}, C-Eval \cite{huang2023ceval}, GPQA \cite{rein2024gpqa}, and AGIEval \cite{zhong2023agieval}, use elementary to professional-level examination questions to test factual recall and reasoning under multiple-choice settings. Instruction-following suites like NaturalInstructions \cite{naturalinstructions}, LLMBAR \cite{zeng2023evaluating}, and IFEval \cite{zhou2023instruction} present open-ended tasks formulated in natural language, measuring a model’s ability to execute diverse directives. Conversational benchmarks---xDial-Eval~\cite{zhang2023xdial}, MT-Bench \cite{zheng2023judging}, LongMemEval~\cite{wu2024longmemeval}, and MultiChallenge~\cite{sirdeshmukh2025multichallenge}---simulate multi-turn dialogues to evaluate coherence, context retention, and response appropriateness in conversational interactions. Human preference collections such as HHH Alignment \cite{askell2021general}, PPE \cite{frick2024evaluate}, and JudgeBench \cite{tan2024judgebench} gather pairwise ranking judgments to directly compare model outputs. Safety and robustness tests like AdvGLUE \cite{wang2021adversarial}, DecodingTrust \cite{wang2023decodingtrust}, and SG-Bench \cite{mou2024sg} probe adversarial vulnerabilities and the propensity for harmful content generation.

Despite their breadth, these benchmarks often depend on objective metrics (\eg, BLEU \cite{papineni-etal-2002-bleu}, ROUGE \cite{lin2004rouge}, and BERTScore \cite{zhang2019bertscore}), which correlate weakly with human judgments in open-ended or specialized tasks \cite{novikova-etal-2017-need, wei2024rethinking}. Furthermore, repeated public exposure leads to data contamination and overfitting, while the static nature of fixed test sets limits adaptation to emerging use cases.

\subsection{Automated LLM-based Evaluation}

Inspired by the strong instruction-following capabilities of closed-source models such as GPT-4, several recent studies~\cite{zeng2023evaluating,chan2023chateval,zhou2023lima,zheng2023judging,dubois2023alpacafarm} have repurposed these advanced LLMs as automated judges. In parallel, the community has begun to develop open-source alternatives to reduce reliance on black-box systems. PandaLM \cite{wang2023pandalm} fine-tunes LLaMA \cite{touvron2023llama} to perform pairwise comparisons and generate brief justifications. Shepard \cite{wang2023shepherd} produces critiques across diverse question-answer scenarios, while Prometheus~\cite{kim2023prometheus} incorporates a thousand fine-grained scoring rubrics during training, enabling customization to specific evaluation criteria.
AUTO-J \cite{li2023generative} is trained on user queries paired with LLM-generated responses under fifty-eight real-world scenarios, yielding both critiques and numerical ratings. More recently, SaMer introduces a scenario-aware, multi-dimensional method that delivers both overall scores and fine-grained feedback on LLM replies \cite{feng2025samer}. Despite their flexibility, automated evaluators remain susceptible to various biases, and their lack of domain-specific expertise can undermine reliability in specialized contexts \cite{zheng2023judging, chen2024humans}.

\subsection{Human-Centric Evaluation} 
Human evaluation persists as the gold standard for judging LLMs, particularly when aligning model outputs with subjective preferences. Platforms like Chatbot Arena \cite{chiang2024chatbot} employ large‐scale, crowdsourced ``battles,'' in which human evaluators compare two chatbots on randomly drawn real‐world instructions to produce an Elo‐based ranking. While this approach can yield robust aggregation results, it does not optimize instructions to maximally differentiate competing LLMs, leading to redundant or trivial comparisons and, consequently, wasted human effort. Furthermore, without an explicit mechanism to ensure coverage of diverse failure modes, random instruction sampling risks both sampling bias and under‐representation of important edge cases. Additionally,
running thousands of battles at scale incurs substantial time and financial costs, and crowdsourced judgments introduce noise and variability that must be countered by still more annotations. Dynabench offers another paradigm of ``adversarial'' data collection, requiring users to manually submit potential counterexamples that expose model weaknesses \citep{kiela2021dynabench}. To mitigate these challenges, our study proposes an automated, adaptive sampling strategy based on MAD Competition.

\section{Proposed Sample-Efficient Human Evaluation Method}
In this section, we present our sample-efficient human evaluation method for comparing LLMs adaptively. Building on the principle of MAD Competition~\cite{wang2008maximum}, our approach automatically selects a small yet highly informative subset of test instructions that maximally differentiate model behaviors. Human evaluators then perform pairwise comparisons on the selected response pairs, and we aggregate their judgments into a global ranking via Elo rating. The overall pipeline is depicted in Figure~\ref{fig:pipeline}.

\subsection{Problem Formulation} 
Let $\mathcal{X}$ denote a large, unlabeled pool of instructions, assembled from diverse, practical natural language processing tasks. We wish to compare
$N$ competing LLMs, denoted as $\mathcal{F} = \{ f_n\}_{n=1}^N$, where each $f_n$ produces an output $y_n = f_n(x)$ for any $x \in \mathcal{X}$. Human evaluators operate in a subjective assessment environment  $H$, wherein they can reliably judge the relative quality of two responses for the same input. Under a strict budget on the number of human annotations, our goal is to produce a definitive, global ranking of the LLMs in $\mathcal{F}$ based on only a handful of carefully chosen comparisons.

\subsection{MAD Competition for LLMs} 

Let us first consider a simple case in which we compare two LLMs $f_i$ and $f_j$. According to the principle of MAD Competition, we seek the instruction 
\begin{equation}\label{eq:mad}
    \hat{x} = \argmax_{x \in \mathcal{X}}  D(f_i(x), f_j(x)),
\end{equation}
where $D(\cdot,\cdot)$ represents a distance measure that quantifies the semantic discrepancy between two responses. The comparative analysis of $f_i(\hat{x})$ versus $f_j(\hat{x})$ can produce three distinct outcomes:
\begin{itemize}
    \item $H\left(f_i(\hat{x}), f_j(\hat{x})\right) \approx 1$. The majority of human evaluators prefer $f_i(\hat{x})$ over $f_j(\hat{x})$, making $f_i(\hat{x})$ the clear winner. The chosen $\hat{x}$ thus serves as a counterexample for $f_j$ and is highly informative for ranking the relative performance of two models.
    \item  $H\left(f_i(\hat{x}), f_j(\hat{x})\right) \approx 0$. Conversely, $f_j$ prevails, indicating that human evaluators overwhelmingly favor $f_j(\hat{x})$ over $f_i(\hat{x})$. Again, $\hat{x}$ functions as a counterexample, this time for $f_i$, and maximally discriminates between the two models.
    \item $H(f_i(\hat{x}), f_j(\hat{x})) \approx 0.5$. Here, the evaluators assign similar ratings to both responses, resulting in a tie. Such ties fall into two subcategories:
    \begin{itemize}
        \item High-rating tie. Both $f_i(\hat{x})$ and $f_j(\hat{x})$ receive strong evaluation, suggesting that each model can generate diverse yet satisfactory responses. This scenario reflects real-world task complexity in which multiple plausible outputs exist. Although $\hat{x}$ underscores each model's strengths, it contributes little to distinguishing their overall performance.
        \item Low-rating tie. Both outputs garner poor scores, indicating that each model fails, albeit in different ways, to follow the instruction. In this case, $\hat{x}$ illustrates their respective weaknesses but again offers limited insight for ranking.
    \end{itemize} 
\end{itemize}
Selecting only the $K$ instructions with the most discrepant responses may yield a narrow set of failure cases. To encourage a broader exploration of model behaviors, we add a diversity term. When selecting the $k$-th instruction, we solve
\begin{equation}\label{eq:mad2}
\resizebox{1\linewidth}{!}{$
\hat{x}^{(k)} = \argmax_{x \in \mathcal{X}\setminus\mathcal{I}} D(f_i(x), f_j(x)) +\lambda D(x,\mathcal{I}),
$}
\end{equation}
where $\mathcal{I}=\{\hat{x}^{(k')}\}_{k'=1}^{K-1}$ is the set of previously chosen instructions, $D(x,\mathcal{I})$ measures the maximum discrepancy between $x$ and any element in $\mathcal{I}$, and  $\lambda >0$ balances discrepancy versus diversity.  Once choosing $\hat{x}^{(k)}$, we add it to $\mathcal{I}$ before proceeding to the next iteration.

By iterating this procedure for all $\binom{N}{2}$ model pairs and retaining $K$ instructions per pair, we construct a MAD response set: 
\begin{equation}
    \mathcal{R} = \{\{\{f_i(\hat{x}^{(k)}), f_j(\hat{x}^{(k)})\}_{k=1}^K\}_{i=1}^{N-1}\}_{j=i+1}^N,
\end{equation}
whose size grows only with $O(N^2K)$, independent of the instruction pool $\vert\mathcal{X}\vert$.

\subsection{Ranking Aggregation via Elo Rating}
For each selected instruction $\hat{x}^{(k)}$ and its paired outputs $\{f_i(\hat{x}^{(k)}), f_j(\hat{x}^{(k)})\}$, human evaluators perform a three-alternative forced choice (3-AFC) task, whose outcome is recorded as
\begin{equation}\label{eq:3afc}
w = 
    \begin{cases}
        1, \quad \text{if } f_i \text{wins}\\
        0, \quad \text{if } f_j \text{wins}\\
        0.5, \quad \text{otherwise}. \\
    \end{cases}
\end{equation}
Elo score~\cite{elo1978rating} updates at the $t$-th comparison:
\begin{equation}\label{eq:elo1}
    s_i^{(t)} = s_i^{(t-1)} + \eta\left(w^{(t)}_{ij} - \frac{1}{1 + 10^{-d^{(t-1)}_{ij}}}\right)
\end{equation} 
and 
\begin{equation}\label{eq:elo2}
    s_j^{(t)} = s_j^{(t-1)} + \eta \left(1 - w^{(t)}_{ij} - \frac{1}{1 + 10^{d^{(t-1)}_{ij}}}\right),
\end{equation}
where 
\begin{equation}\label{eq:elo3}
    d_{ij}^{(t-1)} = \frac{s_i^{(t-1)} - s_j^{(t-1)}}{\tau}.
\end{equation}
$\eta$ controls the learning rate and $\tau$ sets the rating scale. $w^{(t)}_{ij}$ is the result of the $t$-th human comparison of $f_i$ and $f_j$ as defined in Eq.~\eqref{eq:3afc}.
${s}^{(0)}=\{s_n^{(0)}\}_{n=1}^{N}$ are the initial ranking scores of $N$ LLMs. 
To mitigate the sensitivity of the online linear update to comparison order, we apply a bootstrapping procedure as suggested in~\citep{chiang2024chatbot}. Specifically, we generate a thousand bootstrap datasets by sampling with replacement from the human judgments, each dataset being the same size as the original. For each bootstrap dataset, we compute the Elo ratings and then average these ratings across all datasets to produce our final ranking. The full procedure is summarized in Algorithm~\ref{alg:framework}.

\subsection{Incorproating New LLMs} 
Integrating an additional LLM $f_{N+1}$ into MAD Competition is both straightforward and cost-effective. The identified MAD response set $\mathcal{R}$, along with their associated human preferences, remains unchanged. One needs only to sample a new collection of $N \times K$ instruction-response pairs to accentuate behavioral differences between $f_{N+1}$ and $\mathcal{F}=\{f_n\}_{n=1}^N$, obtain human preferences on these new pairs, and then update the global ranking via Eqs.~\eqref{eq:elo1} to~\eqref{eq:elo3}.
The complete procedure for adding a new LLM is detailed in Algorithm \ref{alg:newLLM}.

\begin{algorithm}[t]
\small
\caption{\small{Sample-Efficient Human Evaluation of LLMs via MAD Competition}}
\label{alg:framework} 
\KwIn{An instruction pool $\mathcal{X}$, a set of competing LLMs $\mathcal{F}=\left\{f_{n}\right\}_{n=1}^{N}$, and a semantic discrepancy measure $D(\cdot,\cdot)$\\}
\KwOut{Global ranking scores, ${s} \in \mathbb{R}^{N\times 1}$ \\}
$\mathcal{R}\leftarrow\emptyset$

\For{$n \gets 1$ \KwTo $N$}
{
Generate responses $\left\{f_{n}(x)\vert x \in \mathcal{X}\right\}$
}
\For{$i \gets 1$ \KwTo $N-1$}
{
\For{$j \gets i+1$ \KwTo $N$}
{
$\mathcal{I}\leftarrow\emptyset$\hfill\tcp{$f_i$ versus $f_j$}

\For{$k \gets 1$ \KwTo $K$}
{
Select $\hat{x}^{(k)}\in \mathcal{X}\setminus\mathcal{I}$ by solving Eq. \eqref{eq:mad2}

$\mathcal{I}\leftarrow\mathcal{I}\cup\hat{x}^{(k)}$

$\mathcal{R}\leftarrow\mathcal{R}\cup\{f_i(\hat{x}^{(k)}), f_j(\hat{x}^{(k)})\}$
}
}
}
Collect human judgments on $\mathcal{R}$ via 3-AFC

Compute $s$ via Elo rating with bootstrapping
\end{algorithm}

\begin{algorithm}[t]
\small
\caption{\small{Adding a New LLM into MAD Competition}}
\label{alg:newLLM} 
\KwIn{An instruction pool $\mathcal{X}$, global ranking scores of previous $N$ LLMs $s\in\mathbb{R}^{N\times 1}$, a semantic discrepancy measure $D(\cdot,\cdot)$, and a new competing LLM $f_{N+1}$\\}
\KwOut{Global ranking scores, $\tilde{s} \in \mathbb{R}^{(N+1)\times 1}$ \\}
${\mathcal{R}}\leftarrow\emptyset$

Generate responses $\left\{f_{N+1}(x)\vert x \in \mathcal{X}\right\}$

\For{$i \gets 1$ \KwTo $N$}
{
$\mathcal{I}\leftarrow\emptyset$\hfill\tcp{$f_i$ versus $f_{N+1}$}

\For{$k \gets 1$ \KwTo $K$}
{
Select $\hat{x}^{(k)} \in \mathcal{X}\setminus\mathcal{I}$ by solving Eq. \eqref{eq:mad2}

$\mathcal{I}\leftarrow\mathcal{I}\cup\hat{x}^{(k)}$

${\mathcal{R}}\leftarrow{\mathcal{R}}\cup\{f_i(\hat{x}^{(k)}), f_{N+1}(\hat{x}^{(k)})\}$
}
}
Collect human judgments on ${\mathcal{R}}$ via 3-AFC

Compute $\tilde{s}$ via Elo rating with bootstrapping, starting from $s$
\end{algorithm}

\section{Experiments}
In this section, we demonstrate the effectiveness of our sample-efficient human evaluation method on eight widely used LLMs across four distinct tasks. We first describe our experimental setups (Section~\ref{subsec:setup}), then present the resulting model rankings (Section~\ref{subsec:rankresult}), and compare these rankings against established leaderboards (Section~\ref{subsec:existing-lb}). We further evaluate the reliability and efficiency of our method against alternatives (Section~\ref{subsec:sampling_comparsion}), under different key hyperparameter configurations (Section~\ref{subsec:ablation}), and with extended experimentation in real-world scenarios (Section~\ref{subsec:realworld}).

\begin{table*}[t]
\centering
\caption{Global ranking results obtained by our method for eight LLMs on four different tasks.}
\vspace{-0.2em}
\label{tab:rank_global}
\SetTblrInner{rowsep=1.0pt}
\resizebox{\linewidth}{!}{
\begin{tblr}{
colspec = {l|cc|cc|cc|cc|cc},
}
\toprule
\SetRow{rowsep=2pt}
\SetCell[r=2]{l}{Model} & \SetCell[c=2]{c}{Understanding} & & \SetCell[c=2]{c}{Reasoning} & & \SetCell[c=2]{c}{Writing}
 & & \SetCell[c=2]{c}{Coding} & & \SetCell[c=2]{c}{{Overall}} & \\
 \cmidrule[lr]{2-3}\cmidrule[lr]{4-5}\cmidrule[lr]{6-7}\cmidrule[lr]{8-9}\cmidrule[lr]{10-11} & Rank & Elo Rating & Rank & Elo Rating & Rank & Elo Rating & Rank & Elo Rating & Rank & Elo Rating \\
\midrule
GPT-4-Turbo & 2  & 1,065 & 1  & 1,123 & 1  & 1,162 & 1  & 1,103 & 1 & 1,132 \\
Gemini-Pro & 1  & 1,091 & 2  & 1,094 & 2  & 1,097 & 3  & 1,085 & 2 & 1,107 \\
OpenChat-3.5 & 3  & 1,047 & 3  & 1,087 & 3  & 1,025 & 4  & 971 & 3 & 1,035 \\
GPT-3.5-Turbo & 4  & 988 & 4  & 1,069 & 5  & 976 & 2  & 1,095 & 4 & 1,034 \\
WizardLM-13B & 5  & 986 & 8  & 823 & 4  & 1,001 & 6  & 961 & 5 & 937 \\
Qwen-14B-Chat & 6  & 967 & 6  & 939 & 7  & 918 & 5  & 963 & 6 & 932 \\
ChatGLM3-6B & 8 & 924 & 5 & 998 & 8 & 861 & 7 & 958 & 7 & 929 \\
Vicuna-13B & 7  & 932 & 7  & 869 & 6  & 962 & 8  & 865 & 8 & 894 \\

\bottomrule
\end{tblr}
}
\vspace{-1em}
\end{table*}

\subsection{Experimental Setups}\label{subsec:setup}

\begin{figure}[t]
\centering
\includegraphics[width=0.45\textwidth]{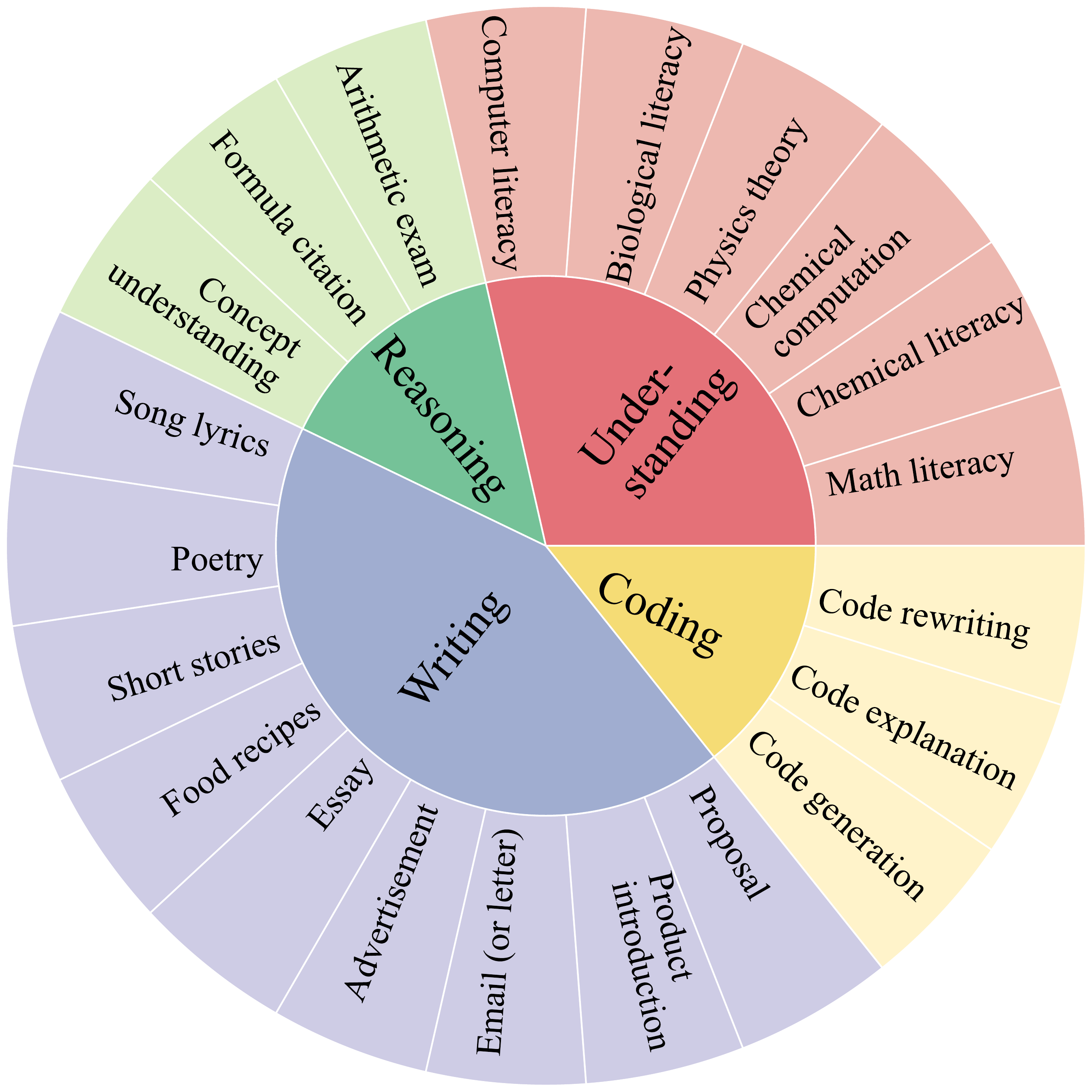}
\caption{Task distribution in our experiment.}
\label{fig:scenario}
\end{figure}

\paragraph{Instruction pool construction}
We curate a large-scale, diverse pool $\mathcal{X}$ of 120K natural language instructions, spanning four core tasks: scientific knowledge understanding, mathematical reasoning, creative and functional writing, and code generation and explanation (see Figure~\ref{fig:scenario}). For each task, we begin with 3K seed instructions drawn from established datasets (\ie, GSM8K \cite{cobbe2021gsm8k}, CAMEL \cite{li2023camel}, AlpacaEval \cite{alpaca_eval}, and CodeAlpaca \cite{codealpaca}) and apply ten rounds of automated instruction evolution~\cite{xu2023wizardlm} to synthesize 30K ``realistic'' instructions that mimic real-world human-chatbot interactions  (see Appendix \ref{ap:pool} for more details). This ensures both breadth across task types and depth in covering potential failure modes.

\paragraph{Model selection} Under a strict human annotation budget, we evaluate eight widely used LLMs, containing three proprietary models: GPT-3.5-Turbo \cite{ouyang2022training}, GPT-4-Turbo \cite{achiam2023gpt} and Gemini-Pro \cite{team2023gemini}, and five open-source models: ChatGLM3-6B \cite{du2022glm}, WizardLM-13B \cite{xu2023wizardlm}, Vicuna-13B \cite{vicuna2023}, OpenChat-3.5 \cite{wang2023openchat}, and Qwen-14B-Chat \cite{qwen} (see Appendix \ref{ap:llm} for their implementations). These span the current spectrum of performance and parameter scales, allowing us to assess both cutting-edge and accessible systems.

\paragraph{Semantic discrepancy measure}
To quantify the semantic discrepancy between paired responses, we embed each using the OpenAI \texttt{text-embedding-ada-002} model and compute cosine dissimilarity~\cite{zhang2019bertscore}. This embedding-based measure to implement $D(\cdot,\cdot)$ in Eq.~\eqref{eq:mad2} reliably captures semantic content discrepancy, and is less prone to model-specific biases.

\paragraph{Human preference collection}\label{collection}
For each unordered model pair $(f_i, f_j)$, we select $K=10$ instructions via MAD Competition, yielding $\binom{8}{2}\times 10=280$ pairwise comparisons. Thirteen STEM-trained postgraduates perform 3-AFC tasks on each paired response, indicating ``win,''  or ``tie.'' These results are then aggregated via Elo rating. More details are given in Appendix \ref{ap:human}.

\subsection{Main Results}\label{subsec:rankresult}

We summarize the evaluation outcomes in Table~\ref{tab:rank_global}, reporting both overall and task-specific rankings. These results illuminate distinct performance patterns across four core tasks.

For \textbf{scientific knowledge understanding},
proprietary models---GPT-4-Turbo, GPT-3.5-Turbo, and Gemini-Pro---dominate this task, reflecting their precise command of domain concepts and robust application of scientific theorems. Remarkably, OpenChat-3.5 (with only 7B parameters) attains a higher ranking than GPT-3.5-Turbo by offering concise yet thorough explanation. In contrast, larger open-source models (\eg, Vicuna-13B) tend to generate redundantly detailed responses.

For \textbf{mathematical reasoning}, our ranking results align closely with the GSM8K leaderboard \cite{cobbe2021gsm8k}, reflecting the source of our instruction pool. Differences in model outputs arise from two primary factors: (1) divergent reasoning paths and (2) arithmetic inaccuracies despite similar reasoning. First, because our generated instructions target problem types and difficulty levels typical of grade-school mathematics, models generally follow singular, concise solution paths; thus, substantial deviations in final answers correspond to noticeably different reasoning trajectories. Second, variations in each model’s computational fidelity can introduce errors in intermediate steps, so even when following analogous logical procedures, minor arithmetic mistakes may propagate to yield incorrect results. Among the models, WizardLM-13B performs relatively poorly. We attribute this to its instruction-evolution training, which uses Alpaca-52K seed data that is not specifically optimized for mathematical tasks \cite{alpaca}. Vicuna-13B exhibits a similar limitation.

For \textbf{creative and functional writing}, the MAD-selected instructions are mostly open-ended, such as ``compose a short story'' or ``craft a holiday recipe'' that invites free-form expression. Consequently, human evaluators consistently favor models that produce longer, more richly detailed outputs over those with terser responses. For example, ChatGLM3-6B generates an average of $221.2$ words per response, whereas GPT-4-Turbo delivers roughly $454.8$ words. This greater verbosity not only provides more descriptive content but also often reflects deeper insights, which in turn drives higher user preference.

For \textbf{code generation and explanation}, human evaluators consider not only the functional correctness of the code but also its fidelity to the given instructions, such as respecting specified line limits, employing designated Python libraries, and conforming to the intended application contexts. We find that LLMs exhibit greater variability in code generation tasks than in code explanation, reflecting the complex interplay between problem specification and implementation. Notably, our human preference rankings correspond closely to established coding benchmarks. For instance, GPT-4-Turbo ($76.83\%$ pass@1), GPT-3.5-Turbo ($74.39\%$ pass@1), and Gemini Pro ($59.76\%$ pass@1) achieve the highest accuracies on HumanEval~\cite{akter2023depth} and are likewise favored in our evaluation method. This alignment underscores the validity of our method to assess LLM performance in realistic coding scenarios.

Table \ref{tab:overview} in the Appendix summarizes each model's strengths and weaknesses across the four evaluated tasks, yielding actionable insights for enhancing response quality. Appendix \ref{ap:experiment} further presents a series of case studies---most notably counterexamples to the otherwise high-performing GPT-4-Turbo---that empirically validate these observations. Moreover, the failure instances uncovered by MAD Competition and confirmed via human evaluation constitute a valuable corpus for developing more reliable LLMs, for example by integrating them into an active learning paradigm \cite{sinha2019variational}.

\begin{figure}[t]
\centering
\includegraphics[width=\linewidth]{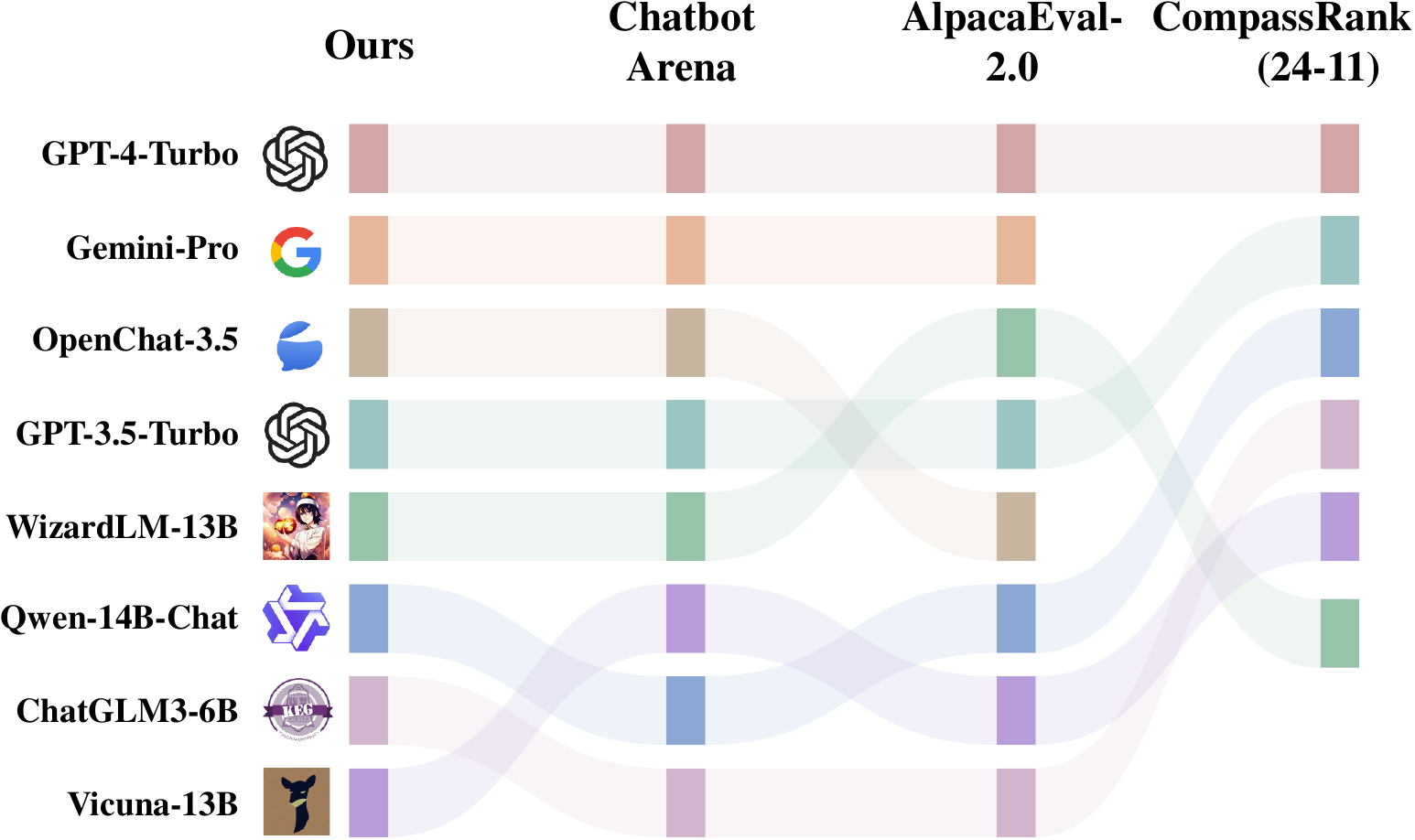}
\caption{Sankey diagram of eight LLMs’ ranking shifts across our sample-efficient human evaluation method, Chatbot Arena, AlpacaEval-2.0, and CompassRank (Nov. 2024 snapshot).}
\label{fig:compare}
\vspace{-1em}
\end{figure}

\begin{table*}[t]
\centering

\caption{Comparison of global ranking results on \textit{mathematical reasoning} using different sampling strategies. Cells with \sethlcolor{red!26}\hl{more saturated} colors indicate more significant ranking discrepancies from the ``gold standard.''}
\label{tab:baselines-reason}
\SetTblrInner{rowsep=1.2pt}
\resizebox{\linewidth}{!}{
\begin{tblr}{
colspec = {l|crcrcrcrcrcrcc},
column{14-15} = {bg=gray!5},     
cell{8-9}{12-13} = {bg=red!4},      
cell{5-9}{2-3} = {bg=red!4},        
cell{4}{2-3} = {bg=red!14},        
cell{3,6,8-9}{4-5} = {bg=red!4},    
cell{4-5}{4-5} = {bg=red!8},        
cell{8-9}{6-7} = {bg=red!4},        
cell{3,6}{6-7} = {bg=red!14},       
cell{6,9}{8-9} = {bg=red!14},      
cell{4-5}{8-9} = {bg=red!4},        
cell{6,9}{10-11} = {bg=red!14},     
}
\toprule
\SetRow{rowsep=2pt}
\SetCell[r=2]{l}{Model} & \SetCell[c=2]{c}{Random} & & \SetCell[c=2]{c}{KL Divergence} & & \SetCell[c=2]{c}{Cross-Entropy} & & \SetCell[c=2]{c}{Anchor Points} & & \SetCell[c=2]{c}{DiffUse} & & \SetCell[c=2]{c}{\makecell{MAD \\Competition}} & & \SetCell[c=2]{c}{\makecell{``Golden'' ranking\\(on GSM8K)}} & \\
    \cmidrule[lr]{2-3}\cmidrule[lr]{4-5}\cmidrule[lr]{6-7}\cmidrule[lr]{8-9}\cmidrule[lr]{10-11}\cmidrule[lr]{12-13}\cmidrule[lr]{14-15} & Rank & Elo & Rank & Elo & Rank & Elo & Rank & Elo & Rank & Elo & Rank & Elo & Rank & Accuracy \\
     \midrule 
GPT-4-Turbo & 1  & 1,028 & 2  & 1,020 & 4  & 983 & 1 & 1,057 & 1 & 1,048 & 1 & 1,157 & 1  & 92.7 \\
OpenChat-3.5 & 5  & 1,000 & 4  & 1,005 & 2  & 1,030 & 3 & 1,044 & 2 & 1,041 & 2 & 1,132 & 2  & 77.3 \\
GPT-3.5-Turbo & 2  & 1,025 & 1  & 1,036 & 3  & 1,025 & 2 & 1,037 & 3 & 1,041 & 3 & 1,079 & 3  & 74.9 \\
ChatGLM3-6B & 3 & 1,007 & 3 & 1,017 & 1 & 1,045 & 7 & 1,023 & 7 & 1,028 & 4 & 1,018 & 4 & 72.3 \\
Qwen-14B-Chat & 4  & 1,007 & 5  & 993 & 5  & 982 & 5 & 1,012 & 5 & 1,012 & 5 & 953 & 5  & 60.1 \\
Vicuna-13B & 7  & 947 & 7  & 957 & 7  & 974 & 6 & 952 & 6 & 942 & 6 & 858 & 7  & 11.3 \\
WizardLM-13B & 6  & 987 & 6  & 972 & 6  & 974 & 4 & 877 & 4 & 886 & 7 & 802 & 6  & 13.5 \\

\bottomrule
\end{tblr}
}
\vspace{-0.6em}
\end{table*}

\subsection{Comparison with Established Leaderboards} \label{subsec:existing-lb}
To validate the reliability of our sample-efficient human evaluation method, we compare its global rankings against three prominent LLM leaderboards: (1) \textit{Chatbot Arena}\footnote{\url{https://huggingface.co/spaces/lmsys/chatbot-arena-leaderboard}}, which aggregates large-scale human preference ``battles'' via Elo rating; (2) \textit{AlpacaEval-2.0}\footnote{\url{https://tatsu-lab.github.io/alpaca_eval/}}, which leverages LLM-based judges to score open-ended instruction-following abilities; and (3) \textit{CompassRank} (Nov. 2024 snapshot)\footnote{\url{https://rank.opencompass.org.cn/leaderboard-llm/?m=24-04}}, which measures performance using standard objective metrics (see Figure~\ref{fig:compare}).

Chatbot Arena draws on extensive, crowdsourced pairwise human judgments, aggregating them via Elo rating to establish a ``gold standard'' benchmark for LLM evaluation. Remarkably, our sample-efficient approach, though engaging only thousands of model ``battles,'' yields rankings that align almost perfectly with those from Chatbot Arena. The lone discrepancy, in Vicuna-13B’s placement, is likely attributable to differences in task sampling, underscoring our method’s ability to faithfully reproduce large-scale human evaluation with minimal annotation effort.

In AlpacaEval 2.0, WizardLM-13B is ranked above OpenChat-3.5 and GPT-3.5-Turbo, in contrast to our human-grounded results. This discrepancy stems from AlpacaEval’s emphasis on assessing LLMs' instruction-following capabilities in unconstrained, open-ended tasks, whereas WizardLM-13B has been fine-tuned on 520K diverse instructions, thus yielding a relative advantage in instruction-following tasks.

When compared to CompassRank, we observe notable shifts for Qwen-14B-Chat and ChatGLM3-6B. Although fine-tuning on MMLU and HumanEval elevates their positions on metric-driven leaderboards, these results alone fail to reflect true human preferences. This gap underscores the necessity of integrating human judgments alongside quantitative metrics to achieve a more comprehensive appraisal of LLM performance.

{\color{ADD}
\subsection{Comparison with Alternative Sampling Strategies} \label{subsec:sampling_comparsion}

To assess the effectiveness of our adaptive sampling strategy based on MAD Competition, we compare it against five alternatives: (1) uniform random sampling, (2) Kullback–Leibler (KL) divergence-based sampling, (3) cross-entropy-based sampling \cite{boubdir2023prompts}, (4) Anchor Points \cite{vivek2023anchor}, and (5) DiffUse \cite{ashury2024label}. Since KL divergence and cross-entropy require access to token log probabilities, we restrict this comparison to seven LLMs that expose such information. All LLMs are evaluated on mathematical reasoning using the GSM8K-derived instruction pool. Human evaluators are instructed to prioritize inference accuracy, and the models’ performance on the original GSM8K test set serves as the ``gold standard.'' As shown in Table~\ref{tab:baselines-reason}, with only $K = 10$ MAD-selected prompts per model pair, our approach reproduces the golden ranking with minimal discrepancy. By contrast, other strategies yield notable ranking errors: KL divergence demotes GPT-4-Turbo below its true position, and cross-entropy erroneously elevates ChatGLM3-6B above stronger baselines such as GPT-3.5-Turbo. These results highlight the effectiveness of MAD Competition in promoting sample efficiency and ranking reliability.

\begin{table}[!t]
\centering
\caption{Comparison of global ranking results under different semantic discrepancy measures.}
\label{tab:rank}
\SetTblrInner{rowsep=1.3pt}
\resizebox{\linewidth}{!}{
\begin{tblr}{
colspec = {l|cr|cr|cr},
cell{3-4}{6-7} = {bg=green!3},
cell{3-4}{2-3} = {bg=green!8}, 
cell{6-7}{6-7} = {bg=red!3}, 
cell{6-7}{4-5} = {bg=red!8}, 
}
\toprule
\SetRow{rowsep=2pt}
 \SetCell[r=2]{l}{Model} & \SetCell[c=2]{c}{BERTScore} & & \SetCell[c=2]{c}{GPT-4-Turbo} & & \SetCell[c=2]{c}{{Default}} & \\
 \cmidrule[lr]{2-3}\cmidrule[lr]{4-5}\cmidrule[lr]{6-7} & Rank & Elo & Rank & Elo & Rank & Elo\\

\midrule
GPT-4-Turbo & 2 & 1,060 &  1 & 1,084 & 1 & 1,162 \\
Gemini-Pro & 1 & 1,061 & 2& 1,040 & 2 & 1,097 \\
OpenChat-3.5 & 3 & 1,020 & 3 & 1,010 & 3 & 1,025\\
WizardLM-13B & 4 & 990  & 5 & 997 & 4 & 1,001\\
GPT-3.5-Turbo & 5 & 989 & 4 & 998 & 5& 976\\
Vicuna-13B & 6 & 982 & 6 & 995 & 6 & 962\\
Qwen-14B-Chat & 7 & 951 & 7 & 974 & 7 & 918\\
ChatGLM3-6B & 8 & 946 & 8 & 902 & 8 & 861\\
\bottomrule
\end{tblr}
}
\vspace{-0.6em}
\end{table}

Beyond quantitative ranking fidelity, we conduct a qualitative analysis of the top-$10$ instructions chosen by each method in the \textit{creative writing} task (see Table \ref{tab:instruct-chosen} in Appendix \ref{ap:experiment}). KL divergence overwhelmingly concentrates on near-homogeneous prompts---nine of ten requests involve poetry---whereas cross-entropy tends to favor academic tasks (\eg, paper outlines and story generation). Random sampling, by its nature, produces a noisy, unpredictable mix of task types with occasional redundancies. In contrast, MAD Competition explicitly incorporates a diversity term, ensuring that each selected instruction probes a distinct aspect of model behavior (see Table~\ref{tab:diversity}). This diversity-aware selection not only minimizes overlap in task categories but also exposes a broader spectrum of failure modes, thereby maximizing the informativeness of each human comparison.

\begin{figure}[t]
\centering
\includegraphics[width=\linewidth]{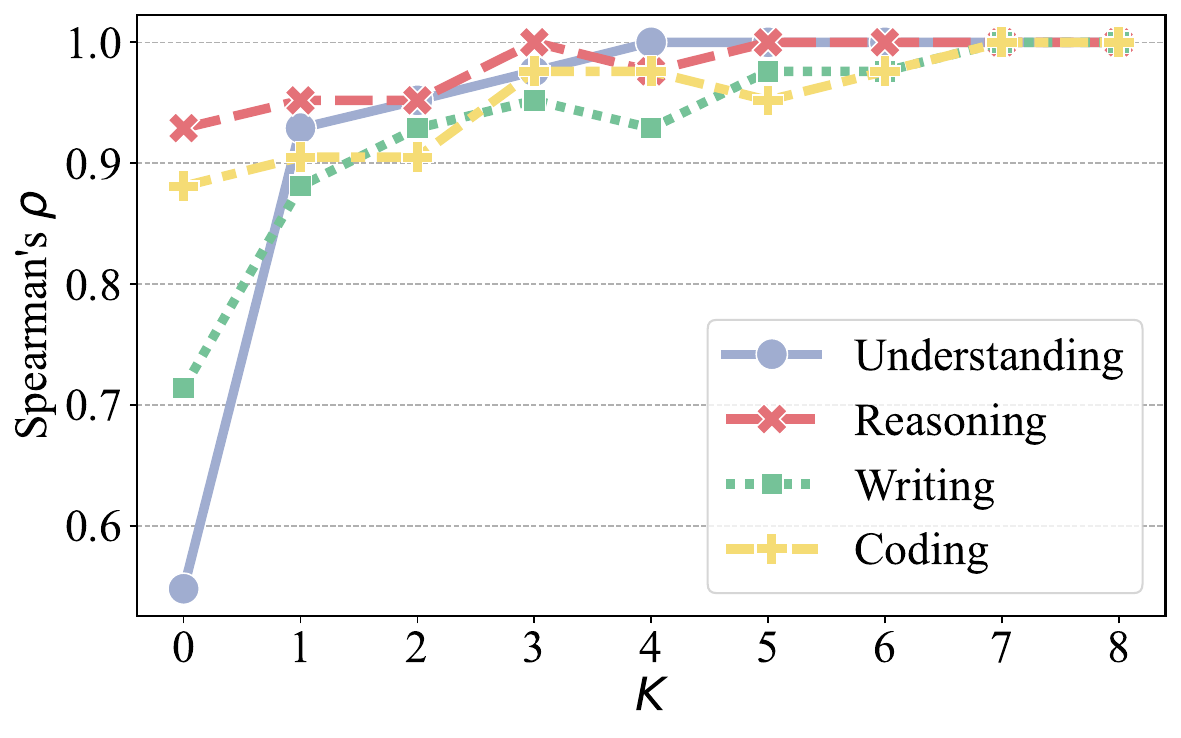} 
\vspace{-2em}
\caption{Spearman's $\rho$ between the global model ranking produced using the default top-$10$ instructions and rankings obtained with reduced prompts ($K\in\{1,\ldots, 9\}$), plotted for each of the four tasks. Correlations exceed $0.95$ for $K \ge 5$ and reach $1.0$ for $K \ge 8$, illustrating the robustness of our sample-efficient evaluation method even under a constrained annotation budget.}
\label{fig:steady}
\vspace{-1em}
\end{figure}

\subsection{Ablation Studies} 
\label{subsec:ablation}

\paragraph{Discrepacny measure sensitivity}
We compare our default \texttt{text-embedding-ada-002} model-based cosine dissimilarity against BERTScore \cite{zhang2019bertscore} and an LLM-as-a-judge approach using GPT-4-Turbo\footnote{We convert semantic similarity measures into discrepancy scores by negating their values.} (see Table~\ref{tab:prompt-gpt4} in the Appendix). Despite their different design principles, all three measures yield near-identical global rankings in the \textit{writing} task (see Table \ref{tab:rank}), demonstrating that the focus of MAD Competition on maximal response discrepancy is effectively captured by diverse semantic discrepancy estimators.

\paragraph{Sample size robustnenss}
We vary $K$ (the number of instructions per model pair) from $1$ to $9$ and compute Spearman rank correlations against the default ranking with $K=10$. Even with as few as five comparisons, correlations exceed $0.95$, reaching a perfect agreement for $K\ge 8$ (see Figure~\ref{fig:steady}). This indicates that reliable model rankings emerge from surprisingly small annotation budgets, and that 
$K=10$ represents a practical compromise between cost and stability.

Importantly, $K$ is treated as a tunable parameter that can vary for each pair of competing LLMs. When two models demonstrate comparable performance, increasing  $K$ allows for additional head-to-head comparisons and thus yields a more robust ranking. Conversely, if one model clearly outperforms the other, it is efficient to reduce 
$K$ (even to zero) to minimize human evaluation effort. This adaptivity parallels the flexible ``battle'' counts used in Chatbot Arena, where the number of matchups between any two models is not fixed but instead reflects their relative similarity.

\paragraph{Diversity weight analysis}
To balance instruction diversity against model discrepancy, we sweep $\lambda \in\{0,0.5,1.0,1.5,2.0\}$ in Eq.~\eqref{eq:mad2}. With $\lambda \le 0.5$, the selected prompts in the \textit{writing} task cluster around a few themes (\eg, poetry), yielding redundant comparisons; at $\lambda =2.0$, diversity increases, but response discrepancy decreases, inflating ``tie'' outcomes. The intermediate value $\lambda = 1.0$ empirically proves optimal, producing an instruction set that preserves informative counterexamples without sacrificing discrimination power.

Taken together, these ablations confirm that our sample-efficient evaluation method is (1) insensitive to reasonable choices of semantic discrepancy measure, (2) robust under reduced annotation budgets, and (3) enhanced by a carefully tuned diversity term.

\begin{table}[!ht]
\centering
\caption{Comparison of global ranking results between our method and MATH500 accuracy.}
\label{tab:math500}
\SetTblrInner{rowsep=1.0pt}
\resizebox{\linewidth}{!}{
\begin{tblr}{
colspec = {l|cc|lr},
cell{5-6}{4-5} = {bg=cyan!4},
cell{13-14}{4-5} = {bg=cyan!4},
}
\toprule
\SetRow{rowsep=2pt}
\SetCell[r=2]{l}{Model} & \SetCell[c=2]{c}{MATH500} & & \SetCell[c=2]{c}{{Ours}} & \\
\cmidrule[lr]{2-3}\cmidrule[lr]{4-5} & Rank & Acc & Rank & Elo \\

\midrule
GPT-4.5 & 1 & 82.6 & 1 & 1,090\\
Claude-3-5-Sonnet-20240620 & 2 & 76.3 & 2 & 1,076\\
GPT-4o-2024-05-13 & 3 & 72.4 & 4 {$\downarrow$} & 1,068\\
GPT-4o-mini-2024-07-18 & 4 & 71.2 & 3 {$\uparrow$} & 1,072\\
GPT-4-Turbo-2024-04-09 & 5 & 68.0 & 5 & 1,067\\
GPT-4-1106-preview & 6 & 60.8 & 6 & 1,061\\
GPT-3.5-Turbo-1106 & 7 & 43.1 & 7 & 1,025\\
Gemma-9b-it & 8 & 34.8 & 8 & 1,012\\
Qwen1.5-14b-chat & 9 & 29.2 & 9 & 1,000\\
Llama-8b-it & 10 & 27.2 & 10 & 977\\
OpenChat-3.5 & 11 & 24.6 & 12 {$\downarrow$} & 935\\
Qwen1.5-7b-chat & 12 & 20.3 & 11 {$\uparrow$} & 958\\
Mistral-7b-it-v0.2 & 13 & 16.4 & 13 & 903\\
ChatGLM3-6b & 14 & 16.0 & 14 & 885\\
Vicuna-13b & 15 & 3.8 & 15 & 870\\

\bottomrule
\end{tblr}
}
\vspace{-0.6em}
\end{table}

\begin{table*}[!t]
\centering
\caption{Comparison of global ranking results between random sampling (with 1K comparisons), 15K subset from Chatbot Arena Conversations,  our method (with 1K comparisons), and the Chatbot Arena leaderboard.}
\label{tab:chatarena}
\SetTblrInner{rowsep=1.0pt}
\resizebox{\linewidth}{!}{
\begin{tblr}{
colspec = {l|rc|rc|rc|rc|rc|rc},
column{12-13} = {bg=gray!5},       
cell{5-6}{8-9} = {bg=red!4},       
cell{12-15}{8-9} = {bg=red!4},
cell{6-7}{10-11} = {bg=red!4},      
cell{12-13}{10-11} = {bg=red!4},
cell{14}{10-11} = {bg=red!8},
cell{5-6,8}{2-3} = {bg=red!4},     
cell{9-11,14,16}{2-3} = {bg=red!8},
cell{7}{2-3} = {bg=red!14},
cell{12-13}{2-3} = {bg=red!20},
cell{6-7,10,15}{4-5} = {bg=red!4},  
cell{5,8}{4-5} = {bg=red!8},
cell{12,14}{4-5} = {bg=red!14},
cell{9,16}{4-5} = {bg=red!20},
cell{11,13}{4-5} = {bg=red!26},
cell{5,8,17}{6-7} = {bg=red!4},     
cell{6-7,10}{6-7} = {bg=red!8},
cell{13}{6-7} = {bg=red!14},
cell{9,11,16}{6-7} = {bg=red!20},
cell{12,15}{6-7} = {bg=red!26},
}
\toprule
\SetRow{rowsep=2pt}
\SetCell[r=2]{l}{Model} & \SetCell[c=2]{c}{\makecell{Random\\(Seed 657)}} & & \SetCell[c=2]{c}{\makecell{Random\\(Seed 216)}} & & \SetCell[c=2]{c}{\makecell{Random\\(Seed 849)}} & & \SetCell[c=2]{c}{\makecell{15K Subset}} & & \SetCell[c=2]{c}{{Ours}} & & \SetCell[c=2]{c}{\makecell{Chatbot Arena\\Leaderboard}} & \\
\cmidrule[lr]{2-3}\cmidrule[lr]{4-5}\cmidrule[lr]{6-7}\cmidrule[lr]{8-9}\cmidrule[lr]{10-11}\cmidrule[lr]{12-13} & Rank & Elo & Rank & Elo & Rank & Elo & Rank & Elo & Rank & Elo & Rank & Elo\\

\midrule
GPT-4 & 1 & 1,119 & 1 & 1,127 & 1 & 1,120 & 1 & 1,254 & 1 & 1,123 & 1 & 1,163\\
Claude-v1 & 2 & 1,095 & 2 & 1,090 & 2 & 1,097 & 2 & 1,196 & 2 & 1,042 & 2 & 1,149\\
GPT-3.5-Turbo & 4 & 1,050 & 5 & 1,024 & 4 & 1,071 & 4 & 1,139 & 3 & 1,044 & 3 & 1,117\\
Claude-Instant-v1 & 3 & 1,083 & 3 & 1,086 & 6 & 1,005 & 3 & 1,164 & 5 & 1,036 & 4 & 1,111\\
Vicuna-13B & 8 & 978 & 4 & 1,079 & 3 & 1,073 & 5 & 1,055 & 4 & 1,042 & 5 & 1,042\\
Vicuna-7B & 7 & 996 & 8 & 977 & 7 & 1,000 & 6 & 1,019 & 6 & 1,019 & 6 & 1,005\\
Koala-13B & 9 & 970 & 11 & 958 & 11 & 956 & 7 & 1,006 & 7 & 1,008 & 7 & 964\\
MPT-7B-Chat & 10 & 967 & 7 & 999 & 9 & 988 & 8 & 946 & 8 & 986 & 8 & 928\\
RWKV-4-Raven-14B & 11 & 963 & 14 & 928 & 13 & 948 & 9 & 938 & 9 & 973 & 9 & 922\\
Alpaca-13B & 5 & 1,046 & 13 & 946 & 15 & 907 & 11 & 906 & 11 & 966 & 10 & 901\\
OAsst-pythia-12B & 6 & 1,011 & 6 & 1,003 & 8 & 993 & 10 & 919 & 12 & 965 & 11 & 893\\
ChatGLM-6B & 14 & 935 & 9 & 961 & 12 & 951 & 13 & 884 & 10 & 968 & 12 & 879\\
FastChat-T5-3B & 13 & 939 & 12 & 952 & 5 & 1,008 & 12 & 894 & 13 & 963 & 13 & 868\\
StableLM-Tuned-Alpha-7B & 12 & 946 & 10 & 958 & 10 & 961 & 14 & 857 & 14 & 915 & 14 & 840\\
Dolly-v2-12B & 15 & 902 & 15 & 914 & 14 & 921 & 15 & 824 & 15 & 902 & 15 & 822\\
\bottomrule
\end{tblr}
}
\vspace{-0.5em}
\end{table*}

\subsection{Further Experimentation}\label{subsec:realworld}
To assess whether our method retains its reliability as the number of compared models grows, we first expand our evaluation to $20$ LLMs. As shown in Table~\ref{tab:twenty_rank} in the Appendix, even at this larger scale, the MAD-derived ranking exhibits a very strong correlation with the Chatbot Arena leaderboard (Spearman’s $\rho = 0.965$ with $p = 5.93\times 10^{-12}$), indicating that our method scales gracefully. Nevertheless, because underlying data distributions may differ between our curated instruction set and the Chatbot Arena benchmark, this single comparison may not fully attest to real-world robustness.

Consequently, we further validate our method on two external, real-world datasets: MATH500~\cite{lightman2023let} and Chatbot Arena Conversations~\cite{zheng2023judging}. For MATH500, we compare model rankings by raw accuracy against those produced by our method with $K = 10$, obtaining an almost perfect concordance (Spearman’s $\rho = 0.993$ with $p = 2.17\times 10^{-13}$ in Table \ref{tab:math500}). For Chatbot Arena Conversations, we sample a subset of 15K dialogues covering fifteen open-source LLMs\footnote{\url{https://huggingface.co/datasets/lmsys/chatbot_arena_conversations}}, ensuring at least $60$ pairwise comparisons per model pair. Using both the full 15K subset and the 1K instances selected by our method (\ie, $\binom{15}{2}\times 10 = 1,050$), we observe Spearman correlations of $0.989$ and $0.986$, respectively (see Table \ref{tab:chatarena}). By contrast, randomly drawing $1,050$ comparisons (three trials with seeds 657, 216, and 849) yields a mean $\rho$ of only $0.791$ (with a standard deviation of $0.069$).

These results demonstrate that (1) our method maintains high fidelity even when scaling to many models, (2) a small, strategically chosen subset of comparisons suffices to replicate full-scale rankings, and (3) our adaptive sampling significantly outperforms na\"{i}ve random selection in both consistency and usability.

\section{Conclusion}
We have introduced a sample-efficient method for human evaluation of LLMs grounded in the principle of MAD Competition. Unlike traditional benchmarks that rely on fixed, manually curated test sets, our method dynamically identifies a minimal set of highly informative instructions that maximize semantic discrepancy between model outputs. By concentrating human judgments on these targeted ``hard'' examples, we achieve reliable global rankings with dramatically fewer annotations. Moreover, the resulting counterexample corpus not only supports accurate model comparison but also serves as rich adversarial data for future model fine-tuning. Our approach is readily extensible to multimodal settings without modifying the core selection and aggregation procedures. Looking ahead, we plan to broaden our coverage by incorporating more LLMs, diversifying evaluation scenarios, and ultimately publishing an open, comprehensive leaderboard that can adapt as new models emerge.

\section*{Limitations}
Our current implementation performs an exhaustive (brute-force) search over the entire instruction pool to identify the top-$K$ discrepant examples for each model pair. While effective for pools on the order of $10^{5}$ instructions, this approach may become computationally prohibitive as pools scale to millions or when comparing large numbers of models. Future work could explore gradient-based or heuristic optimization techniques (\eg, proxy models for discrepancy gradients, Bayesian optimization) to reduce search costs without sacrificing selection quality.

Although we optimize for semantic discrepancy and instruction diversity, we do not explicitly account for the variable cognitive load that different comparisons impose on human evaluators. Certain pairs of responses may be inherently harder to judge---due to subtle semantic differences, open-ended prompts, or required domain expertise---leading to increased annotation time or lower inter-rater agreement. Incorporating a computational measure of human difficulty, for example, estimating decision uncertainty via a small pilot annotation pass, could inform more balanced sample selection and adaptive budgeting of human effort.

Extending our method to dozens or hundreds of competing LLMs still entails $O(N^2K)$ human judgments, which may strain annotation budgets despite our efficiency gains. Coarse-to-fine strategies, such as seeding rankings with a strong LLM judge and then only collecting human judgments for closely ranked subsets, can partially mitigate this cost, but developing fully automated pipelines remains an open challenge.

\section*{Acknowledgements}
This work was supported in part by the National Natural Science Foundation of China (62301480 and 62302433), Zhejiang Provincial ``Jianbing'' ``Lingyan'' Research and Development Program of China (2025C01097 and 2024C01135), and Hangzhou West Lake Pearl Project Leading Innovative Youth Team Project (TD2023017).

\bibliography{ref}

\begin{thebibliography}{77}
\providecommand{\natexlab}[1]{#1}

\bibitem[{Akter et~al.(2023)Akter, Yu, Muhamed, Ou, B{\"a}uerle, Cabrera, Dholakia, Xiong, and Neubig}]{akter2023depth}
Syeda~Nahida Akter, Zichun Yu, Aashiq Muhamed, Tianyue Ou, Alex B{\"a}uerle, {\'A}ngel~Alexander Cabrera, Krish Dholakia, Chenyan Xiong, and Graham Neubig. 2023.
\newblock An in-depth look at {G}emini's language abilities.
\newblock \emph{arXiv:2312.11444}.

\bibitem[{Ashury-Tahan et~al.(2024)Ashury-Tahan, Gera, Sznajder, Choshen, Ein-Dor, and Shnarch}]{ashury2024label}
Shir Ashury-Tahan, Ariel Gera, Benjamin Sznajder, Leshem Choshen, Liat Ein-Dor, and Eyal Shnarch. 2024.
\newblock Label-efficient model selection for text generation.
\newblock \emph{arXiv:2402.07891}.

\bibitem[{Askell et~al.(2021)Askell, Bai, Chen, Drain, Ganguli, Henighan, Jones, Joseph, Mann, DasSarma et~al.}]{askell2021general}
Amanda Askell, Yuntao Bai, Anna Chen, Dawn Drain, Deep Ganguli, Tom Henighan, Andy Jones, Nicholas Joseph, Ben Mann, Nova DasSarma, et~al. 2021.
\newblock A general language assistant as a laboratory for alignment.
\newblock \emph{arXiv:2112.00861}.

\bibitem[{Austin et~al.(2021)Austin, Odena, Nye, Bosma, Michalewski, Dohan, Jiang, Cai, Terry, Le et~al.}]{austin2021program}
Jacob Austin, Augustus Odena, Maxwell Nye, Maarten Bosma, Henryk Michalewski, David Dohan, Ellen Jiang, Carrie Cai, Michael Terry, Quoc Le, et~al. 2021.
\newblock Program synthesis with large language models.
\newblock \emph{arXiv:2108.07732}.

\bibitem[{Azaria et~al.(2024)Azaria, Azoulay, and Reches}]{azaria2024chatgpt}
Amos Azaria, Rina Azoulay, and Shulamit Reches. 2024.
\newblock Chat{GPT} is a remarkable tool—for experts.
\newblock \emph{Data Intelligence}, 6(1):240--296.

\bibitem[{Bai et~al.(2023)Bai, Bai, Chu, Cui, Dang, Deng, Fan, Ge, Han, Huang et~al.}]{qwen}
Jinze Bai, Shuai Bai, Yunfei Chu, Zeyu Cui, Kai Dang, Xiaodong Deng, Yang Fan, Wenbin Ge, Yu~Han, Fei Huang, et~al. 2023.
\newblock Qwen technical report.
\newblock \emph{arXiv:2309.16609}.

\bibitem[{Boubdir et~al.(2023)Boubdir, Kim, Ermis, Fadaee, and Hooker}]{boubdir2023prompts}
Meriem Boubdir, Edward Kim, Beyza Ermis, Marzieh Fadaee, and Sara Hooker. 2023.
\newblock Which prompts make the difference? {Data} prioritization for efficient human {LLM} evaluation.
\newblock \emph{arXiv:2310.14424}.

\bibitem[{Chan et~al.(2023)Chan, Chen, Su, Yu, Xue, Zhang, Fu, and Liu}]{chan2023chateval}
Chi-Min Chan, Weize Chen, Yusheng Su, Jianxuan Yu, Wei Xue, Shanghang Zhang, Jie Fu, and Zhiyuan Liu. 2023.
\newblock {ChatEval}: {T}owards better {LLM}-based evaluators through multi-agent debate.
\newblock \emph{arXiv:2308.07201}.

\bibitem[{Chang et~al.(2024)Chang, Wang, Wang, Wu, Yang, Zhu, Chen, Yi, Wang, Wang et~al.}]{chang2023survey}
Yupeng Chang, Xu~Wang, Jindong Wang, Yuan Wu, Linyi Yang, Kaijie Zhu, Hao Chen, Xiaoyuan Yi, Cunxiang Wang, Yidong Wang, et~al. 2024.
\newblock A survey on evaluation of large language models.
\newblock \emph{ACM Transactions on Intelligent Systems and Technology}, 15(3):1--45.

\bibitem[{Chaudhary(2023)}]{codealpaca}
Sahil Chaudhary. 2023.
\newblock Code {Alpaca}: {A}n instruction-following {LLaMA} model for code generation.
\newblock GitHub repository \url{https://github.com/sahil280114/codealpaca}.

\bibitem[{Chen et~al.(2024)Chen, Chen, Liu, Jiang, and Wang}]{chen2024humans}
Guiming~Hardy Chen, Shunian Chen, Ziche Liu, Feng Jiang, and Benyou Wang. 2024.
\newblock Humans or {LLM}s as the judge? {A} study on judgement biases.
\newblock \emph{arXiv:2402.10669}.

\bibitem[{Chen(2023)}]{chen2023large}
Huajun Chen. 2023.
\newblock Large knowledge model: {P}erspectives and challenges.
\newblock \emph{arXiv:2312.02706}.

\bibitem[{Chia et~al.(2023)Chia, Hong, Bing, and Poria}]{chia2023instructeval}
Yew~Ken Chia, Pengfei Hong, Lidong Bing, and Soujanya Poria. 2023.
\newblock {InstructEval}: {T}owards holistic evaluation of instruction-tuned large language models.
\newblock \emph{arXiv:2306.04757}.

\bibitem[{Chiang et~al.(2023)Chiang, Li, Lin, Sheng, Wu, Zhang, Zheng, Zhuang, Zhuang, Gonzalez et~al.}]{vicuna2023}
Wei-Lin Chiang, Zhuohan Li, Zi~Lin, Ying Sheng, Zhanghao Wu, Hao Zhang, Lianmin Zheng, Siyuan Zhuang, Yonghao Zhuang, Joseph~E Gonzalez, et~al. 2023.
\newblock Vicuna: {A}n open-source chatbot impressing {GPT}-4 with 90\%* {ChatGPT} quality.
\newblock Blog post \url{https://lmsys.org/blog/2023-03-30-vicuna/}.

\bibitem[{Chiang et~al.(2024)Chiang, Zheng, Sheng, Angelopoulos, Li, Li, Zhang, Zhu, Jordan, Gonzalez et~al.}]{chiang2024chatbot}
Wei-Lin Chiang, Lianmin Zheng, Ying Sheng, Anastasios~Nikolas Angelopoulos, Tianle Li, Dacheng Li, Hao Zhang, Banghua Zhu, Michael Jordan, Joseph~E Gonzalez, et~al. 2024.
\newblock Chatbot {A}rena: {A}n open platform for evaluating {LLM}s by human preference.
\newblock \emph{arXiv:2403.04132}.

\bibitem[{Cobbe et~al.(2021)Cobbe, Kosaraju, Bavarian, Chen, Jun, Kaiser, Plappert, Tworek, Hilton, Nakano et~al.}]{cobbe2021gsm8k}
Karl Cobbe, Vineet Kosaraju, Mohammad Bavarian, Mark Chen, Heewoo Jun, Lukasz Kaiser, Matthias Plappert, Jerry Tworek, Jacob Hilton, Reiichiro Nakano, et~al. 2021.
\newblock Training verifiers to solve math word problems.
\newblock \emph{arXiv:2110.14168}.

\bibitem[{Du et~al.(2021)Du, Qian, Liu, Ding, Qiu, Yang, and Tang}]{du2022glm}
Zhengxiao Du, Yujie Qian, Xiao Liu, Ming Ding, Jiezhong Qiu, Zhilin Yang, and Jie Tang. 2021.
\newblock {GLM}: {G}eneral language model pretraining with autoregressive blank infilling.
\newblock \emph{arXiv:2103.10360}.

\bibitem[{Dubois et~al.(2023)Dubois, Li, Taori, Zhang, Gulrajani, Ba, Guestrin, Liang, and Hashimoto}]{dubois2023alpacafarm}
Yann Dubois, Xuechen Li, Rohan Taori, Tianyi Zhang, Ishaan Gulrajani, Jimmy Ba, Carlos Guestrin, Percy Liang, and Tatsunori~B Hashimoto. 2023.
\newblock {AlpacaFarm}: {A} simulation framework for methods that learn from human feedback.
\newblock \emph{arXiv:2305.14387}.

\bibitem[{Elo and Sloan(1978)}]{elo1978rating}
Arpad~E Elo and Sam Sloan. 1978.
\newblock \emph{The {R}ating of {C}hessplayers: Past and {P}resent}.
\newblock Ishi Press International.

\bibitem[{Elton(2004)}]{elton2004goodhart}
Lewis Elton. 2004.
\newblock Goodhart's law and performance indicators in higher education.
\newblock \emph{Evaluation \& Research in Education}, 18(1-2):120--128.

\bibitem[{Feng et~al.(2025)Feng, Ding, Yu, Qu, Chen, Yu, Zhang, Chen et~al.}]{feng2025samer}
Kehua Feng, Keyan Ding, Jing Yu, Yiwen Qu, Zhiwen Chen, Gang Yu, Qiang Zhang, Huajun Chen, et~al. 2025.
\newblock Sa{M}er: {A} scenario-aware multi-dimensional evaluator for large language models.
\newblock In \emph{The Thirteenth International Conference on Learning Representations}.

\bibitem[{Frick et~al.(2024)Frick, Li, Chen, Chiang, Angelopoulos, Jiao, Zhu, Gonzalez, and Stoica}]{frick2024evaluate}
Evan Frick, Tianle Li, Connor Chen, Wei-Lin Chiang, Anastasios~N Angelopoulos, Jiantao Jiao, Banghua Zhu, Joseph~E Gonzalez, and Ion Stoica. 2024.
\newblock How to evaluate reward models for {RLHF}.
\newblock \emph{arXiv:2410.14872}.

\bibitem[{Grattafiori et~al.(2024)Grattafiori, Dubey, Jauhri, Pandey, Kadian, Al-Dahle, Letman, Mathur, Schelten, Vaughan et~al.}]{grattafiori2024llama}
Aaron Grattafiori, Abhimanyu Dubey, Abhinav Jauhri, Abhinav Pandey, Abhishek Kadian, Ahmad Al-Dahle, Aiesha Letman, Akhil Mathur, Alan Schelten, Alex Vaughan, et~al. 2024.
\newblock The {LLaMA} 3 herd of models.
\newblock \emph{arXiv:2407.21783}.

\bibitem[{Guo et~al.(2023)Guo, Jin, Liu, Huang, Shi, Yu, Liu, Li, Xiong, Xiong et~al.}]{guo2023survey}
Zishan Guo, Renren Jin, Chuang Liu, Yufei Huang, Dan Shi, Linhao Yu, Yan Liu, Jiaxuan Li, Bojian Xiong, Deyi Xiong, et~al. 2023.
\newblock Evaluating large language models: {A} comprehensive survey.
\newblock \emph{arXiv:2310.19736}.

\bibitem[{Hendrycks et~al.(2020)Hendrycks, Burns, Basart, Zou, Mazeika, Song, and Steinhardt}]{hendryckstest2021}
Dan Hendrycks, Collin Burns, Steven Basart, Andy Zou, Mantas Mazeika, Dawn Song, and Jacob Steinhardt. 2020.
\newblock Measuring massive multitask language understanding.
\newblock \emph{arXiv:2009.03300}.

\bibitem[{Huang et~al.(2023)Huang, Bai, Zhu, Zhang, Zhang, Su, Liu, Lv, Zhang, Lei et~al.}]{huang2023ceval}
Yuzhen Huang, Yuzhuo Bai, Zhihao Zhu, Junlei Zhang, Jinghan Zhang, Tangjun Su, Junteng Liu, Chuancheng Lv, Yikai Zhang, Jiayi Lei, et~al. 2023.
\newblock {C-Eval}: {A} multi-level multi-discipline chinese evaluation suite for foundation models.
\newblock \emph{arXiv:2305.08322}.

\bibitem[{Husain et~al.(2019)Husain, Wu, Gazit, Allamanis, and Brockschmidt}]{husain2019codesearchnet}
Hamel Husain, Ho-Hsiang Wu, Tiferet Gazit, Miltiadis Allamanis, and Marc Brockschmidt. 2019.
\newblock Codesearchnet challenge: {E}valuating the state of semantic code search.
\newblock \emph{arXiv:1909.09436}.

\bibitem[{Jaiswal et~al.(2020)Jaiswal, Babu, Zadeh, Banerjee, and Makedon}]{jaiswal2020survey}
Ashish Jaiswal, Ashwin~Ramesh Babu, Mohammad~Zaki Zadeh, Debapriya Banerjee, and Fillia Makedon. 2020.
\newblock A survey on contrastive self-supervised learning.
\newblock \emph{arXiv:2011.00362}.

\bibitem[{Jiang et~al.(2023)Jiang, Sablayrolles, Mensch, Bamford, Chaplot, Casas, Bressand, Lengyel, Lample, Saulnier et~al.}]{jiang2023mistral}
Albert~Q Jiang, Alexandre Sablayrolles, Arthur Mensch, Chris Bamford, Devendra~Singh Chaplot, Diego de~las Casas, Florian Bressand, Gianna Lengyel, Guillaume Lample, Lucile Saulnier, et~al. 2023.
\newblock Mistral 7{B}.
\newblock \emph{arXiv:2310.06825}.

\bibitem[{Kiela et~al.(2021)Kiela, Bartolo, Nie, Kaushik, Geiger, Wu, Vidgen, Prasad, Singh, Ringshia et~al.}]{kiela2021dynabench}
Douwe Kiela, Max Bartolo, Yixin Nie, Divyansh Kaushik, Atticus Geiger, Zhengxuan Wu, Bertie Vidgen, Grusha Prasad, Amanpreet Singh, Pratik Ringshia, et~al. 2021.
\newblock {Dynabench}: {R}ethinking benchmarking in {NLP}.
\newblock \emph{arXiv:2104.14337}.

\bibitem[{Kim et~al.(2023)Kim, Shin, Cho, Jang, Longpre, Lee, Yun, Shin, Kim, Thorne et~al.}]{kim2023prometheus}
Seungone Kim, Jamin Shin, Yejin Cho, Joel Jang, Shayne Longpre, Hwaran Lee, Sangdoo Yun, Seongjin Shin, Sungdong Kim, James Thorne, et~al. 2023.
\newblock Prometheus: {I}nducing fine-grained evaluation capability in language models.
\newblock In \emph{The Twelfth International Conference on Learning Representations}.

\bibitem[{Kim et~al.(2024)Kim, Suk, Longpre, Lin, Shin, Welleck, Neubig, Lee, Lee, and Seo}]{kim2024prometheus}
Seungone Kim, Juyoung Suk, Shayne Longpre, Bill~Y Lin, Jamin Shin, Sean Welleck, Graham Neubig, Moontae Lee, Kyungjae Lee, and Minjoon Seo. 2024.
\newblock Prometheus 2: {A}n open source language model specialized in evaluating other language models.
\newblock \emph{arXiv:2405.01535}.

\bibitem[{Kwon et~al.(2023)Kwon, Li, Zhuang, Sheng, Zheng, Yu, Gonzalez, Zhang, and Stoica}]{kwon2023efficient}
Woosuk Kwon, Zhuohan Li, Siyuan Zhuang, Ying Sheng, Lianmin Zheng, Cody~H Yu, Joseph Gonzalez, Hao Zhang, and Ion Stoica. 2023.
\newblock Efficient memory management for large language model serving with pagedattention.
\newblock In \emph{Symposium on Operating Systems Principles}, page 611–626.

\bibitem[{Li et~al.(2023{\natexlab{a}})Li, Hammoud, Itani, Khizbullin, and Ghanem}]{li2023camel}
Guohao Li, Hasan Abed Al~Kader Hammoud, Hani Itani, Dmitrii Khizbullin, and Bernard Ghanem. 2023{\natexlab{a}}.
\newblock {CAMEL}: {C}ommunicative agents for ``mind" exploration of large language model society.
\newblock \emph{arXiv:2303.17760}.

\bibitem[{Li et~al.(2023{\natexlab{b}})Li, Sun, Yuan, Fan, Zhao, and Liu}]{li2023generative}
Junlong Li, Shichao Sun, Weizhe Yuan, Run-Ze Fan, Hai Zhao, and Pengfei Liu. 2023{\natexlab{b}}.
\newblock Generative judge for evaluating alignment.
\newblock \emph{arXiv:2310.05470}.

\bibitem[{Li et~al.(2023{\natexlab{c}})Li, Zhang, Li, You, and Cui}]{li2023evaluation}
Linhan Li, Huaping Zhang, Chunjin Li, Haowen You, and Wenyao Cui. 2023{\natexlab{c}}.
\newblock Evaluation on {C}hat{GPT} for {C}hinese language understanding.
\newblock \emph{Data Intelligence}, 5(4):885--903.

\bibitem[{Li et~al.(2023{\natexlab{d}})Li, Zhang, Dubois, Taori, Gulrajani, Guestrin, Liang, and Hashimoto}]{alpaca_eval}
Xuechen Li, Tianyi Zhang, Yann Dubois, Rohan Taori, Ishaan Gulrajani, Carlos Guestrin, Percy Liang, and Tatsunori~B. Hashimoto. 2023{\natexlab{d}}.
\newblock {AlpacaEval}: {A}n automatic evaluator of instruction-following models.
\newblock GitHub repository \url{https://github.com/tatsu-lab/alpaca_eval}.

\bibitem[{Lightman et~al.(2023)Lightman, Kosaraju, Burda, Edwards, Baker, Lee, Leike, Schulman, Sutskever, and Cobbe}]{lightman2023let}
Hunter Lightman, Vineet Kosaraju, Yuri Burda, Harrison Edwards, Bowen Baker, Teddy Lee, Jan Leike, John Schulman, Ilya Sutskever, and Karl Cobbe. 2023.
\newblock Let's verify step by step.
\newblock In \emph{The Twelfth International Conference on Learning Representations}.

\bibitem[{Lin(2004)}]{lin2004rouge}
Chin-Yew Lin. 2004.
\newblock {ROUGE}: A package for automatic evaluation of summaries.
\newblock In \emph{Text Summarization Branches Out}, pages 74--81.

\bibitem[{Mishra et~al.(2021)Mishra, Khashabi, Baral, and Hajishirzi}]{naturalinstructions}
Swaroop Mishra, Daniel Khashabi, Chitta Baral, and Hannaneh Hajishirzi. 2021.
\newblock Cross-task generalization via natural language crowdsourcing instructions.
\newblock \emph{arXiv:2104.08773}.

\bibitem[{Mou et~al.(2024)Mou, Zhang, and Ye}]{mou2024sg}
Yutao Mou, Shikun Zhang, and Wei Ye. 2024.
\newblock {SG-Bench}: {E}valuating {LLM} safety generalization across diverse tasks and prompt types.
\newblock In \emph{Advances in Neural Information Processing Systems}, pages 123032--123054.

\bibitem[{Novikova et~al.(2017)Novikova, Du{\v{s}}ek, Curry, and Rieser}]{novikova-etal-2017-need}
Jekaterina Novikova, Ond{\v{r}}ej Du{\v{s}}ek, Amanda~Cercas Curry, and Verena Rieser. 2017.
\newblock Why we need new evaluation metrics for {NLG}.
\newblock In \emph{Empirical Methods in Natural Language Processing}, pages 2241--2252.

\bibitem[{OpenAI(2023)}]{achiam2023gpt}
OpenAI. 2023.
\newblock {GPT}-4 technical report.
\newblock \emph{arXiv:2303.08774}.

\bibitem[{Ouyang et~al.(2022)Ouyang, Wu, Jiang, Almeida, Wainwright, Mishkin, Zhang, Agarwal, Slama, Ray et~al.}]{ouyang2022training}
Long Ouyang, Jeffrey Wu, Xu~Jiang, Diogo Almeida, Carroll Wainwright, Pamela Mishkin, Chong Zhang, Sandhini Agarwal, Katarina Slama, Alex Ray, et~al. 2022.
\newblock {T}raining language models to follow instructions with human feedback.
\newblock In \emph{Advances in Neural Information Processing Systems}, pages 27730--27744.

\bibitem[{Papineni et~al.(2002)Papineni, Roukos, Ward, and Zhu}]{papineni-etal-2002-bleu}
Kishore Papineni, Salim Roukos, Todd Ward, and Wei-Jing Zhu. 2002.
\newblock {BLEU}: A method for automatic evaluation of machine translation.
\newblock In \emph{Annual Meeting of the Association for Computational Linguistics}, pages 311--318.

\bibitem[{Rein et~al.(2023)Rein, Hou, Stickland, Petty, Pang, Dirani, Michael, and Bowman}]{rein2024gpqa}
David Rein, Betty~Li Hou, Asa~Cooper Stickland, Jackson Petty, Richard~Yuanzhe Pang, Julien Dirani, Julian Michael, and Samuel~R Bowman. 2023.
\newblock {GPQA}: A graduate-level google-proof {Q\&A} benchmark.
\newblock \emph{arXiv:2311.12022}.

\bibitem[{Schaeffer(2023)}]{schaeffer2023pretraining}
Rylan Schaeffer. 2023.
\newblock Pretraining on the test set is all you need.
\newblock \emph{arXiv:2309.08632}.

\bibitem[{Sinha et~al.(2019)Sinha, Ebrahimi, and Darrell}]{sinha2019variational}
Samarth Sinha, Sayna Ebrahimi, and Trevor Darrell. 2019.
\newblock Variational adversarial active learning.
\newblock In \emph{International Conference on Computer Vision}, pages 5972--5981.

\bibitem[{Sirdeshmukh et~al.(2025)Sirdeshmukh, Deshpande, Mols, Jin, Cardona, Lee, Kritz, Primack, Yue, and Xing}]{sirdeshmukh2025multichallenge}
Ved Sirdeshmukh, Kaustubh Deshpande, Johannes Mols, Lifeng Jin, Ed-Yeremai Cardona, Dean Lee, Jeremy Kritz, Willow Primack, Summer Yue, and Chen Xing. 2025.
\newblock Multi{C}hallenge: {A} realistic multi-turn conversation evaluation benchmark challenging to frontier {LLMs}.
\newblock \emph{arXiv:2501.17399}.

\bibitem[{Srivastava et~al.(2022)Srivastava, Rastogi, Rao, Shoeb, Abid, Fisch, Brown, Santoro, Gupta, Garriga-Alonso et~al.}]{srivastava2022beyond}
Aarohi Srivastava, Abhinav Rastogi, Abhishek Rao, Abu Awal~Md Shoeb, Abubakar Abid, Adam Fisch, Adam~R Brown, Adam Santoro, Aditya Gupta, Adri{\`a} Garriga-Alonso, et~al. 2022.
\newblock Beyond the imitation game: {Q}uantifying and extrapolating the capabilities of language models.
\newblock \emph{arXiv:2206.04615}.

\bibitem[{Tan et~al.(2024)Tan, Zhuang, Montgomery, Tang, Cuadron, Wang, Popa, and Stoica}]{tan2024judgebench}
Sijun Tan, Siyuan Zhuang, Kyle Montgomery, William~Y Tang, Alejandro Cuadron, Chenguang Wang, Raluca~Ada Popa, and Ion Stoica. 2024.
\newblock {JudgeBench}: {A} benchmark for evaluating {LLM}-based judges.
\newblock \emph{arXiv:2410.12784}.

\bibitem[{Taori et~al.(2023)Taori, Gulrajani, Zhang, Dubois, Li, Guestrin, Liang, and Hashimoto}]{alpaca}
Rohan Taori, Ishaan Gulrajani, Tianyi Zhang, Yann Dubois, Xuechen Li, Carlos Guestrin, Percy Liang, and Tatsunori~B. Hashimoto. 2023.
\newblock Stanford {Alpaca}: {A}n instruction-following {LLaMA} model.
\newblock GitHub repository \url{https://github.com/tatsu-lab/stanford_alpaca}.

\bibitem[{Team et~al.(2023)Team, Anil, Borgeaud, Wu, Alayrac, Yu, Soricut, Schalkwyk, Dai, Hauth et~al.}]{team2023gemini}
Gemini Team, Rohan Anil, Sebastian Borgeaud, Yonghui Wu, Jean-Baptiste Alayrac, Jiahui Yu, Radu Soricut, Johan Schalkwyk, Andrew~M Dai, Anja Hauth, et~al. 2023.
\newblock Gemini: {A} family of highly capable multimodal models.
\newblock \emph{arXiv:2312.11805}.

\bibitem[{Touvron et~al.(2023)Touvron, Lavril, Izacard, Martinet, Lachaux, Lacroix, Rozi{\`e}re, Goyal, Hambro, Azhar et~al.}]{touvron2023llama}
Hugo Touvron, Thibaut Lavril, Gautier Izacard, Xavier Martinet, Marie-Anne Lachaux, Timoth{\'e}e Lacroix, Baptiste Rozi{\`e}re, Naman Goyal, Eric Hambro, Faisal Azhar, et~al. 2023.
\newblock {LLaMA}: {O}pen and efficient foundation language models.
\newblock \emph{arXiv:2302.13971}.

\bibitem[{Vivek et~al.(2023)Vivek, Ethayarajh, Yang, and Kiela}]{vivek2023anchor}
Rajan Vivek, Kawin Ethayarajh, Diyi Yang, and Douwe Kiela. 2023.
\newblock Anchor points: {B}enchmarking models with much fewer examples.
\newblock \emph{arXiv:2309.08638}.

\bibitem[{Wang et~al.(2023{\natexlab{a}})Wang, Chen, Pei, Xie, Kang, Zhang, Xu, Xiong, Dutta, Schaeffer et~al.}]{wang2023decodingtrust}
Boxin Wang, Weixin Chen, Hengzhi Pei, Chulin Xie, Mintong Kang, Chenhui Zhang, Chejian Xu, Zidi Xiong, Ritik Dutta, Rylan Schaeffer, et~al. 2023{\natexlab{a}}.
\newblock {DecodingTrust}: {A} comprehensive assessment of trustworthiness in {GPT} models.
\newblock \emph{arXiv:2306.11698}.

\bibitem[{Wang et~al.(2021)Wang, Xu, Wang, Gan, Cheng, Gao, Awadallah, and Li}]{wang2021adversarial}
Boxin Wang, Chejian Xu, Shuohang Wang, Zhe Gan, Yu~Cheng, Jianfeng Gao, Ahmed~Hassan Awadallah, and Bo~Li. 2021.
\newblock Adversarial {GLUE}: {A} multi-task benchmark for robustness evaluation of language models.
\newblock \emph{arXiv:2111.02840}.

\bibitem[{Wang et~al.(2023{\natexlab{b}})Wang, Cheng, Zhan, Li, Song, and Liu}]{wang2023openchat}
Guan Wang, Sijie Cheng, Xianyuan Zhan, Xiangang Li, Sen Song, and Yang Liu. 2023{\natexlab{b}}.
\newblock {OpenChat}: {A}dvancing open-source language models with mixed-quality data.
\newblock \emph{arXiv:2309.11235}.

\bibitem[{Wang et~al.(2023{\natexlab{c}})Wang, Li, Chen, Cai, Zhu, Lin, Cao, Liu, Liu, and Sui}]{wang2023large}
Peiyi Wang, Lei Li, Liang Chen, Zefan Cai, Dawei Zhu, Binghuai Lin, Yunbo Cao, Qi~Liu, Tianyu Liu, and Zhifang Sui. 2023{\natexlab{c}}.
\newblock Large language models are not fair evaluators.
\newblock \emph{arXiv:2305.17926}.

\bibitem[{Wang et~al.(2023{\natexlab{d}})Wang, Yu, Tan, O'Brien, Pasunuru, Dwivedi-Yu, Golovneva, Zettlemoyer, Fazel-Zarandi, and Celikyilmaz}]{wang2023shepherd}
Tianlu Wang, Ping Yu, Xiaoqing~Ellen Tan, Sean O'Brien, Ramakanth Pasunuru, Jane Dwivedi-Yu, Olga Golovneva, Luke Zettlemoyer, Maryam Fazel-Zarandi, and Asli Celikyilmaz. 2023{\natexlab{d}}.
\newblock Shepherd: {A} critic for language model generation.
\newblock \emph{arXiv:2308.04592}.

\bibitem[{Wang et~al.(2023{\natexlab{e}})Wang, Yu, Zeng, Yang, Wang, Chen, Jiang, Xie, Wang, Xie et~al.}]{wang2023pandalm}
Yidong Wang, Zhuohao Yu, Zhengran Zeng, Linyi Yang, Cunxiang Wang, Hao Chen, Chaoya Jiang, Rui Xie, Jindong Wang, Xing Xie, et~al. 2023{\natexlab{e}}.
\newblock {PandaLM}: {A}n automatic evaluation benchmark for {LLM} instruction tuning optimization.
\newblock \emph{arXiv:2306.05087}.

\bibitem[{Wang et~al.(2022)Wang, Kordi, Mishra, Liu, Smith, Khashabi, and Hajishirzi}]{wang2022self}
Yizhong Wang, Yeganeh Kordi, Swaroop Mishra, Alisa Liu, Noah~A Smith, Daniel Khashabi, and Hannaneh Hajishirzi. 2022.
\newblock Self-{I}nstruct: {A}ligning language models with self-generated instructions.
\newblock \emph{arXiv:2212.10560}.

\bibitem[{Wang and Simoncelli(2008)}]{wang2008maximum}
Zhou Wang and Eero~P Simoncelli. 2008.
\newblock Maximum differentiation ({MAD}) competition: A methodology for comparing computational models of perceptual quantities.
\newblock \emph{Journal of Vision}, 8(12):1--13.

\bibitem[{Wei et~al.(2024)Wei, Chen, and Luo}]{wei2024rethinking}
Fangyun Wei, Xi~Chen, and Lin Luo. 2024.
\newblock Rethinking generative large language model evaluation for semantic comprehension.
\newblock \emph{arXiv:2403.07872}.

\bibitem[{Whitehouse et~al.(2025)Whitehouse, Wang, Yu, Li, Weston, Kulikov, and Saha}]{whitehouse2025j1}
Chenxi Whitehouse, Tianlu Wang, Ping Yu, Xian Li, Jason Weston, Ilia Kulikov, and Swarnadeep Saha. 2025.
\newblock J1: {I}ncentivizing thinking in {LLM}-as-a-judge via reinforcement learning.
\newblock \emph{arXiv:2505.10320}.

\bibitem[{Wu et~al.(2024)Wu, Wang, Yu, Zhang, Chang, and Yu}]{wu2024longmemeval}
Di~Wu, Hongwei Wang, Wenhao Yu, Yuwei Zhang, Kai-Wei Chang, and Dong Yu. 2024.
\newblock {LongMemEval}: {B}enchmarking chat assistants on long-term interactive memory.
\newblock \emph{arXiv:2410.10813}.

\bibitem[{Xu et~al.(2023)Xu, Sun, Zheng, Geng, Zhao, Feng, Tao, and Jiang}]{xu2023wizardlm}
Can Xu, Qingfeng Sun, Kai Zheng, Xiubo Geng, Pu~Zhao, Jiazhan Feng, Chongyang Tao, and Daxin Jiang. 2023.
\newblock {WizardLM}: {E}mpowering large language models to follow complex instructions.
\newblock \emph{arXiv:2304.12244}.

\bibitem[{Zeng et~al.(2023)Zeng, Yu, Gao, Meng, Goyal, and Chen}]{zeng2023evaluating}
Zhiyuan Zeng, Jiatong Yu, Tianyu Gao, Yu~Meng, Tanya Goyal, and Danqi Chen. 2023.
\newblock Evaluating large language models at evaluating instruction following.
\newblock \emph{arXiv:2310.07641}.

\bibitem[{Zhang et~al.(2023{\natexlab{a}})Zhang, D'Haro, Tang, Shi, Tang, and Li}]{zhang2023xdial}
Chen Zhang, Luis~F. D'Haro, Chengguang Tang, Ke~Shi, Guohua Tang, and Haizhou Li. 2023{\natexlab{a}}.
\newblock x{D}ial-{Eval}: {A} multilingual open-domain dialogue evaluation benchmark.
\newblock \emph{arXiv:2310.08958}.

\bibitem[{Zhang et~al.(2019)Zhang, Kishore, Wu, Weinberger, and Artzi}]{zhang2019bertscore}
Tianyi Zhang, Varsha Kishore, Felix Wu, Kilian~Q Weinberger, and Yoav Artzi. 2019.
\newblock {BERTScore}: {E}valuating text generation with {BERT}.
\newblock \emph{arXiv:1904.09675}.

\bibitem[{Zhang et~al.(2023{\natexlab{b}})Zhang, Yu, Yu, Lv, Liu, Huang, Xu, and Li}]{zhang2023wider}
Xinghua Zhang, Bowen Yu, Haiyang Yu, Yangyu Lv, Tingwen Liu, Fei Huang, Hongbo Xu, and Yongbin Li. 2023{\natexlab{b}}.
\newblock Wider and deeper {LLM} networks are fairer {LLM} evaluators.
\newblock \emph{arXiv:2308.01862}.

\bibitem[{Zheng et~al.(2023)Zheng, Chiang, Sheng, Zhuang, Wu, Zhuang, Lin, Li, Li, Xing et~al.}]{zheng2023judging}
Lianmin Zheng, Wei-Lin Chiang, Ying Sheng, Siyuan Zhuang, Zhanghao Wu, Yonghao Zhuang, Zi~Lin, Zhuohan Li, Dacheng Li, Eric Xing, et~al. 2023.
\newblock Judging {LLM}-as-a-judge with {MT}-{B}ench and {C}hatbot {A}rena.
\newblock \emph{arXiv:2306.05685}.

\bibitem[{Zhong et~al.(2023)Zhong, Cui, Guo, Liang, Lu, Wang, Saied, Chen, and Duan}]{zhong2023agieval}
Wanjun Zhong, Ruixiang Cui, Yiduo Guo, Yaobo Liang, Shuai Lu, Yanlin Wang, Amin Saied, Weizhu Chen, and Nan Duan. 2023.
\newblock {AGIEval}: {A} human-centric benchmark for evaluating foundation models.
\newblock \emph{arXiv:2304.06364}.

\bibitem[{Zhou et~al.(2023{\natexlab{a}})Zhou, Liu, Xu, Iyer, Sun, Mao, Ma, Efrat, Yu, Yu et~al.}]{zhou2023lima}
Chunting Zhou, Pengfei Liu, Puxin Xu, Srini Iyer, Jiao Sun, Yuning Mao, Xuezhe Ma, Avia Efrat, Ping Yu, Lili Yu, et~al. 2023{\natexlab{a}}.
\newblock {LIMA}: {L}ess is more for alignment.
\newblock \emph{arXiv:2305.11206}.

\bibitem[{Zhou et~al.(2023{\natexlab{b}})Zhou, Lu, Mishra, Brahma, Basu, Luan, Zhou, and Hou}]{zhou2023instruction}
Jeffrey Zhou, Tianjian Lu, Swaroop Mishra, Siddhartha Brahma, Sujoy Basu, Yi~Luan, Denny Zhou, and Le~Hou. 2023{\natexlab{b}}.
\newblock Instruction-following evaluation for large language models.
\newblock \emph{arXiv:2311.07911}.

\bibitem[{Zhou et~al.(2023{\natexlab{c}})Zhou, Zhu, Chen, Chen, Zhao, Chen, Lin, Wen, and Han}]{zhou2023don}
Kun Zhou, Yutao Zhu, Zhipeng Chen, Wentong Chen, Wayne~Xin Zhao, Xu~Chen, Yankai Lin, Ji-Rong Wen, and Jiawei Han. 2023{\natexlab{c}}.
\newblock Don't make your {LLM} an evaluation benchmark cheater.
\newblock \emph{arXiv:2311.01964}.

\bibitem[{Zhu et~al.(2023)Zhu, Wang, and Wang}]{zhu2023judgelm}
Lianghui Zhu, Xinggang Wang, and Xinlong Wang. 2023.
\newblock {JudgeLM}: {F}ine-tuned large language models are scalable judges.
\newblock \emph{arXiv:2310.17631}.

\end{thebibliography}

\clearpage
\appendix

\section{Elo Rating System}
\label{ap:elo}

The Elo rating system \cite{elo1978rating}, devised by Arpad Elo in the 1960s, quantifies the relative skill of two competitors in games like chess or tennis. Each player holds a numerical rating that is revised after every match based on the actual result (win, loss, or tie) versus the expected outcome computed from the rating difference. Upsets---when a lower-rated player defeats a stronger opponent---yield larger point gains, whereas favored players earn fewer points for expected victories. The size of each update is governed by two parameters: the scaling factor $\tau$, which determines how rating differences translate into expected scores, and the K-factor $\eta$, which caps the maximum change per match. In our experiments, we set $\tau = 400$ and $\eta = 4$, consistent with the Chatbot Arena protocol.


\section{Instruction Pool} 
\label{ap:pool}
We construct a large-scale instruction pool $\mathcal{X}$ to serve as unbiased test data for comparing LLMs. Our construction comprises three stages:
\begin{itemize}
    \item \textbf{Task definition}. We identify four capability dimensions---scientific knowledge understanding, mathematical reasoning, creative and functional writing, and code generation and explanation---and include corresponding subtasks (see Figure~\ref{fig:scenario}).
    \item \textbf{Seed instruction collection}. For each task, we sample 3K prompts from established benchmarks:
    \begin{itemize}
        \item \textit{Scientific knowledge understanding} is a task to evaluate the scientific knowledge comprehension and application abilities of LLMs. Questions from CAMEL \cite{li2023camel} cover mathematics, physics, chemistry, biology, and computer science.
        \item \textit{Mathematical reasoning}  is a commonly used task to assess the math problem-solving abilities of LLMs. Problems are sourced from GSM8K \cite{cobbe2021gsm8k}.
        \item \textit{Creative and functional writing} engages in open-ended writing, aiming to satisfy user requirements. Tasks are from AlpacaEval \cite{alpaca_eval} and IMPACT \cite{chia2023instructeval}.
        \item \textit{Code generation and explanation} aims to generate high-quality code snippets based on given instructions. Coding tasks are from CodeSearchNet \cite{husain2019codesearchnet}, MBPP \cite{austin2021program}, and CodeAlpaca \cite{codealpaca}. 
    \end{itemize}
    \item \textbf{Instruction evolution}. Leveraging three strong closed-source models (GPT-4-Turbo, GPT-3.5-Turbo, and Gemini Pro), we perform ten iterative evolutions per seed \cite{xu2023wizardlm}. In each iteration, we prompt models to (i) adapt seed instructions toward more practical or nuanced scenarios (\eg, transforming ``a mundane text abbreviation'' into ``design a mnemonic to aid in memorizing a complex algorithm'') and (ii) impose varied constraints (\eg, ``limit code to $15$ lines,'' ``compose a $1,500$-word article,'' or ``write in Shakespearean style''). For interpretability in human evaluation, evolution prompts for understanding, reasoning, and coding also request exemplar answers.
\end{itemize}
This process yields 30K evolved instructions per task (120K in total), ensuring both diversity and real-world relevance. Figure~\ref{fig:scenario} illustrates the final task distribution, and Tables \ref{tab:prompt-s} to \ref{tab:prompt-c} list our default evolution prompts.

\begin{figure*}[t]
    \centering
    \includegraphics[width=\linewidth]{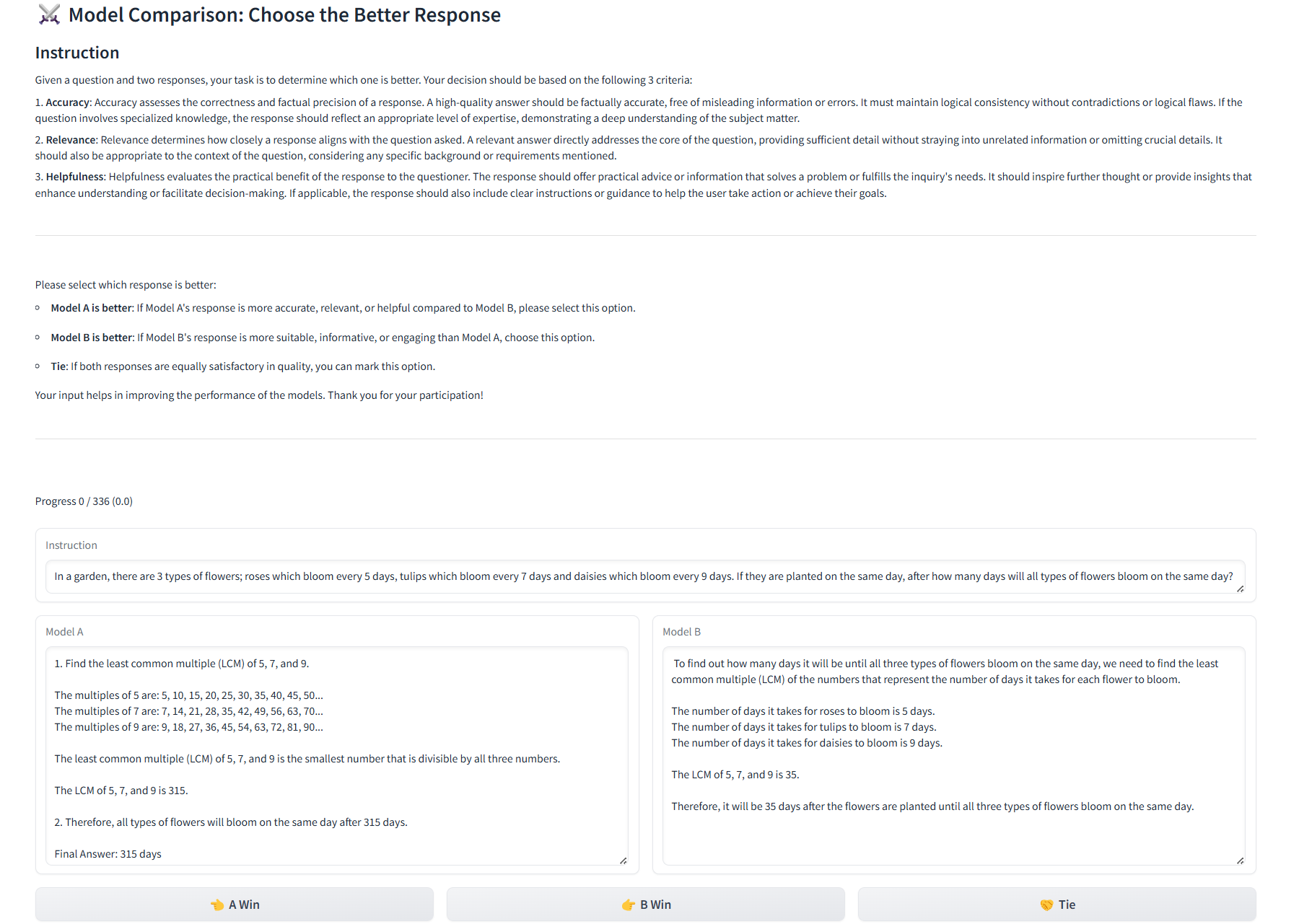} 
    \caption{Graphical user interface for collecting human preferences.}
    \label{fig:interface}
\end{figure*}

\section{Competing LLMs}
\label{ap:llm}
We apply our sample-efficient evaluation method to eight representative LLMs:
\begin{itemize}
    \item \textbf{GPT-4-Turbo} (\ie, GPT-4-1106-preview) and \textbf{GPT-3.5-Turbo} (\ie, GPT-3.5-Turbo-1106) are OpenAI's most advanced proprietary models at the time.
    \item \textbf{Gemini-Pro} (\ie, Gemini-1.0-Pro) \cite{team2023gemini} is Google's multimodal closed-source model, trained jointly on diverse, high-quality multimodal data, demonstrating strong understanding and reasoning capabilities across a variety of specialized domains.
    \item \textbf{OpenChat-3.5} \cite{ouyang2022training} is a 7B-parameter derivative of Mistral-7B \cite{jiang2023mistral} that applies C-RLFT \cite{wang2023openchat} to mixed-quality ShareGPT dialogues (70K in total, including 6K generated by GPT-4) for fine-tuning.
    \item \textbf{WizardLM-13B} (\ie, WizardLM-13B-V1.2) \cite{xu2023wizardlm} is built on LLaMA2-13B \cite{touvron2023llama} and refined via the instruction evolution method \textit{Evol-Instruct}, which expands the 52K seed instructions in Alpaca \cite{alpaca} to a 520K-instruction corpus.
    \item \textbf{Vicuna-13B} (\ie, Vicuna-13B-V1.5) \cite{vicuna2023} is a 13B-parameter model, fine-tuned from LLaMA2-13B \cite{touvron2023llama} on real human-machine dialogue data from ShareGPT.
    \item \textbf{Qwen-14B-Chat} \cite{qwen} is a 14B-parameter model, fine-tuned from Qwen-14B-base that is pretrained on a large-scale, diverse dataset of over three trillion tokens, covering multiple languages such as Chinese and English. 
    \item \textbf{ChatGLM3-6B} \cite{du2022glm} is a 6B-parameter model originated from ChatGLM3-6B-Base, pretrained on over one trillion data.
\end{itemize}

\paragraph{Implementation}
We conduct all experiments in a zero-shot setting. Proprietary models (GPT-4-Turbo, GPT-3.5-Turbo, and Gemini-Pro) are accessed via their official APIs with the temperature set to $0.7$, top-$p$ to 1.0, and a maximum sequence length of $2,048$, leaving all other parameters at their defaults. The five open-source models are deployed locally on two NVIDIA GeForce RTX 4090 GPUs and accelerated using the vLLM method \cite{kwon2023efficient}, with the same inference settings, except that Qwen-14B-Chat is limited to a maximum length of $1,024$ tokens.

\section{Details of Human Evaluation} \label{ap:human}

This section provides a detailed overview of our human evaluation studies.

\subsection{Evaluator Selection Criteria}
All evaluators have graduate‐level STEM training and meet the following academic and technical proficiency criteria:

\begin{itemize}
    \item \textbf{Language proficiency}. Each evaluator meets at least one of the following requirements:
    \begin{itemize}
        \item Native‐level English proficiency.
        \item National College Entrance Examination (NCEE) English score $\ge 125$ and College English Test (CET-6) score $\ge 500$.
    \end{itemize}
    \item \textbf{Disciplinary foundation}. They shall possess:
    \begin{itemize}
        \item High‐school–level mastery of mathematics, physics, chemistry, biology, and formal logic.
        \item General knowledge of computer science.
        \item Proficiency in Python at a professional level.
    \end{itemize}
    \item \textbf{Sustained concentration}. Evaluators commit to offline sessions of at least two hours (with breaks in between), ensuring the accuracy and efficiency of their annotations.

\end{itemize}

\begin{table*}[t]
\color{ADD}
\centering
\caption{Comparison of global ranking results between LLM-based and human evaluation across four tasks.}
\SetTblrInner{rowsep=1.1pt}
\resizebox{0.95\linewidth}{!}{
\begin{tblr}{
colspec = {l|cc|cc|cc|cc|cc},
}

\toprule 
\SetCell[r=2]{l}{Model} & \SetCell[c=2]{c}{Understanding} & & \SetCell[c=2]{c}{Reasoning} & & \SetCell[c=2]{c}{Writing} & & \SetCell[c=2]{c}{Coding} & & \SetCell[c=2]{c}{Overall} &\\
\cmidrule[lr]{2-3}\cmidrule[lr]{4-5}\cmidrule[lr]{6-7}\cmidrule[lr]{8-9}\cmidrule[lr]{10-11} & \makecell{Human\\Rank} & \makecell{GPT-4o\\Rank} & \makecell{Human\\Rank} & \makecell{GPT-4o\\Rank} & \makecell{Human\\Rank} & \makecell{GPT-4o\\Rank} & \makecell{Human\\Rank} & \makecell{GPT-4o\\Rank} & \makecell{Human\\Rank} & \makecell{GPT-4o\\Rank} \\
\midrule 
GPT-4-Turbo & 2 & 2 & 1 & 1 & 1 & 1 & 1 & 1 & 1 & 1 \\
Gemini-Pro & 1 & 1 & 2 & 3 & 2 & 2 & 3 & 3 & 2 & 2 \\
OpenChat-3.5 & 3 & 3 & 3 & 2 & 3 & 3 & 4 & 5 & 3 & 3 \\
GPT-3.5-Turbo & 4 & 4 & 4 & 4 & 5 & 4 & 2 & 2 & 4 & 4 \\
WizardLM-13B & 5 & 5 & 8 & 8 & 4 & 5 & 6 & 6 & 5 & 5 \\
Qwen-14B-Chat & 6 & 6 & 6 & 6 & 7 & 7 & 5 & 4 & 6 & 6 \\
ChatGLM3-6B & 8 & 7 & 5 & 5 & 8 & 8 & 7 & 8 & 7 & 8 \\
Vicuna-13B & 7 & 8 & 7 & 7 & 6 & 6 & 8 & 7 & 8 & 7 \\
\midrule
Spearman's $\rho$ & \SetCell[c=2]{c}{0.9762} & & \SetCell[c=2]{c}{0.9762} & & \SetCell[c=2]{c}{0.9762} & & \SetCell[c=2]{c}{0.9524} & & \SetCell[c=2]{c}{0.9762} &\\
\bottomrule
\end{tblr}
}
\label{tab:ai_vs_human}
  
\end{table*}

\subsection{Pre-Experiment Briefing}

Prior to participation, each annotator receives a comprehensive briefing on the study’s objectives and procedures. Participants are explicitly informed that their annotation outputs would form part of the research dataset and that their continued involvement constitutes voluntary consent.

All annotation data are treated as strictly confidential and used exclusively for scientific analysis. No personally identifiable information is collected, stored, or disclosed, ensuring that participation entails no risk or adverse consequences for the annotators.

\subsection{Graphical User Interface}
Figure \ref{fig:interface} illustrates the interface employed to elicit human preference judgments. Initially, annotators review the task instructions, which (1) describe the core task---choosing the better of two model-generated responses---and (2) define the evaluation criteria (accuracy, relevance, and helpfulness). During the experiment, each annotator examines the two candidate responses alongside the instruction and then records her/his decision by clicking one of three buttons at the bottom of the page: ``A win,'' ``Tie,'' or ``B win.''

\subsection{Human Evaluation Process}
To evaluate model performance, we first identify the ten most discriminative instructions for pairwise comparisons among eight models across four distinct tasks. This selection yields $4\times\binom{8}{2}\times 10 =1,120$ comparisons. Thirteen graduate students with strong STEM backgrounds are recruited to annotate these pairs, each of which receives annotations from at least five different students, so that, on average, each annotator assesses approximately $345$ pairs. The annotation phase lasts about one week. We measure the annotation agreement and observe an average inter-annotator agreement of $83.39\%$. In cases of divergent labels, the final consensus is determined by a majority vote.

\begin{table*}[ht]
\centering
\caption{Global ranking results of twenty LLMs by our LLM-based evaluation method. The gray column denotes the Chatbot Arena leaderboard positions, and the two rankings exhibit strong agreement (Spearman’s $\rho = 0.965$).}
\label{tab:twenty_rank}
\SetTblrInner{rowsep=1.2pt}
\resizebox{\linewidth}{!}{
\begin{tblr}{
colspec = {l|cr|cr|cr|cr|cr|c},
column{12} = {bg=gray!5},
cell{5,11,14,15,20}{2-3} = {bg=red!4},
cell{6,16}{2-3} = {bg=red!8},
cell{4,7,8,18}{2-3} = {bg=red!14},
cell{13}{2-3} = {bg=red!20},
cell{21}{2-3} = {bg=red!26},
cell{17}{2-3} = {bg=red!32},
cell{12}{2-3} = {bg=red!38},
cell{7,10,13,14}{4-5} = {bg=red!0},
cell{11,20}{4-5} = {bg=red!4},
cell{5,16,17,18,21,22}{4-5} = {bg=red!8},
cell{19}{4-5} = {bg=red!14},
cell{4,8,9,15}{4-5} = {bg=red!20},
cell{12}{4-5} = {bg=red!26},
cell{3}{4-5} = {bg=red!32},
cell{10,19,20,21,22}{6-7} = {bg=red!0},
cell{4,12,17,18}{6-7} = {bg=red!4},
cell{8,13,14}{6-7} = {bg=red!8},
cell{5,6,7,15}{6-7} = {bg=red!14},
cell{11}{6-7} = {bg=red!20},
cell{9,16}{6-7} = {bg=red!26},
cell{14,22}{8-9} = {bg=red!0},
cell{8,10,16,19,20}{8-9} = {bg=red!4},
cell{3,11,17,21}{8-9} = {bg=red!8},
cell{9,13,15}{8-9} = {bg=red!14},
cell{4,7,12}{8-9} = {bg=red!20},
cell{6}{8-9} = {bg=red!26},
cell{18}{8-9} = {bg=red!32},
cell{8,13,20,21,22}{10-11} = {bg=red!0},
cell{4,5,9,10,17,18}{10-11} = {bg=red!4},
cell{11,14,16}{10-11} = {bg=red!8},
cell{7,12,15}{10-11} = {bg=red!14},
}

\toprule
\SetCell[r=2]{l}{Model} & \SetCell[c=2]{c}{Understanding} & & \SetCell[c=2]{c}{Reasoning} & & \SetCell[c=2]{c}{Writing} & & \SetCell[c=2]{c}{Coding} & & \SetCell[c=2]{c}{{Overall}} & & Chatbot Arena \\
\cmidrule[lr]{2-3}\cmidrule[lr]{4-5}\cmidrule[lr]{6-7}\cmidrule[lr]{8-9}\cmidrule[lr]{10-11}\cmidrule[lr]{12-12} & Rank & Elo & Rank & Elo & Rank & Elo & Rank & Elo & Rank & Elo & Rank\\
\midrule
GPT-4o-2024-05-13          & 1  & 1,166 & 7  & 1,065 & 1  & 1,106 & 3  & 1,097 & 1  & 1,151 & 1\\
GPT-4o-mini-2024-07-18     & 5  & 1,126 & 6  & 1,078 & 3  & 1,096 & 6  & 1,077 & 3  & 1,144 & 2\\
Claude-3.5-Sonnet-20240620 & 4  & 1,139 & 1  & 1,105 & 6  & 1,054 & 2  & 1,110 & 4  & 1,142 & 3\\
Gemini-1.5-pro-latest      & 6  & 1,110 & 4  & 1,088 & 7  & 1,052 & 9  & 1,044 & 5  & 1,112 & 4\\
GPT-4-Turbo-2024-04-09     & 2  & 1,164 & 5  & 1,078 & 2  & 1,104 & 1  & 1,110 & 2  & 1,147 & 5\\
GPT-4-1106-preview         & 3  & 1,144 & 2  & 1,103 & 4  & 1,088 & 5  & 1,080 & 6  & 1,100 & 6\\
Claude-3-Sonnet-20240229   & 7  & 1,055 & 3  & 1,103 & 12 & 1,022 & 4  & 1,097 & 8  & 1,077 & 7\\
Gemma2-9B-it               & 8  & 1,035 & 8  & 1,030 & 8  & 1,049 & 7  & 1,058 & 9  & 1,069 & 8\\
Llama3.1-8B-it             & 10 & 978   & 10 & 1,009 & 5  & 1,070 & 11 & 1,014 & 7  & 1,079 & 9\\
Llama3-8B-it               & 18 & 878   & 15 & 951   & 11 & 1,023 & 14 & 964   & 13 & 1,001 & 10\\
Gemini-pro                 & 15 & 910   & 11 & 1,006 & 13 & 1,014 & 8  & 1,048 & 11 & 1,021 & 11\\
Qwen1.5-14B-Chat           & 11 & 973   & 12 & 1,000 & 10 & 1,026 & 12 & 998   & 10 & 1,030 & 12\\
OpenChat-3.5               & 12 & 959   & 9  & 1,016 & 16 & 930   & 16 & 911   & 16 & 939   & 13\\
Mistral-7B-it              & 16 & 910   & 16 & 941   & 9  & 1,032 & 15 & 930   & 12 & 1,005 & 14\\
Qwen1.5-7B-Chat            & 9  & 1,016 & 13 & 983   & 14 & 975   & 13 & 977   & 14 & 969   & 15\\
GPT-3.5-Turbo-1106         & 13 & 959   & 14 & 977   & 15 & 952   & 10 & 1,017 & 15 & 942   & 16\\
Wizardlm-13B               & 17 & 909   & 20 & 832   & 17 & 906   & 18 & 881   & 17 & 815   & 17\\
Vicuna-13B                 & 19 & 871   & 19 & 851   & 18 & 889   & 19 & 865   & 18 & 804   & 18\\
Qwen-14B-Chat              & 14 & 924   & 17 & 924   & 19 & 836   & 17 & 909   & 19 & 786   & 19\\
ChatGLM3-6B                & 20 & 773   & 18 & 859   & 20 & 777   & 20 & 814   & 20 & 670   & 20\\
\bottomrule
\end{tblr}
}
\vskip-0.5em
\end{table*}


\section{More Experimental Results}\label{ap:experiment}

\subsection{LLM-based Evaluation}
The proposed sample-efficient evaluation method for LLMs substantially reduces human effort but does not eliminate it, which hampers scalability as the number of models grows. To overcome this bottleneck, we substitute human judges with specialized LLM‐based evaluators guided by carefully crafted prompts (Tables \ref{tab:evaluator_prompt_under} to \ref{tab:evaluator_prompt_code}). As shown in Table \ref{tab:ai_vs_human}, the Spearman's $\rho$ between LLM (in particular, GPT-4o-2024-08-26) and human preferences exceeds $0.95$, confirming that top-performing LLMs (provided that they are excluded from MAD Competition) can reliably emulate human judgments. Exploiting this finding, we then apply our LLM-based method to twenty state‐of‐the‐art LLMs (see Table \ref{tab:twenty_rank}). The resulting ranking exhibits strong agreement with that obtained via the labor‐intensive Chatbot Arena, demonstrating that our approach can scale to large model sets with minimal additional labor and significantly reduced time costs.

\begin{table*}[t]
\color{ADD2}
\centering
\caption{Comparison of global ranking results using different sampling algorithms in the \textit{writing} task.}
\SetTblrInner{rowsep=1.3pt}
\resizebox{\linewidth}{!}{
\begin{tblr}{
colspec = {l|cr|cr|cr|cr|cr|cr|cr},
column{14-15} = {bg=gray!5},
cell{4,7}{2-3} = {bg=red!4},         
cell{5}{2-3} = {bg=red!8},            
cell{8}{2-3} = {bg=red!20},          
cell{7}{4-5} = {bg=red!4},             
cell{4-6}{4-5} = {bg=red!8},          
cell{8}{4-5} = {bg=red!14},         
cell{7}{6-7} = {bg=red!4},         
cell{4-6}{6-7} = {bg=red!8},        
cell{8}{6-7} = {bg=red!14},        
cell{4-5,7}{8-9} = {bg=red!4},      
cell{8-9}{8-9} = {bg=red!8},         
cell{6}{8-9} = {bg=red!14},          
cell{4-6,9}{10-11} = {bg=red!4},     
cell{7}{10-11} = {bg=red!8},          
cell{8}{10-11} = {bg=red!20},        
}
\toprule 
\SetCell[r=2]{l}{Model} & \SetCell[c=2]{c}{Random} & & \SetCell[c=2]{c}{KL Divergence} & & \SetCell[c=2]{c}{Cross-Entropy} & & \SetCell[c=2]{c}{Anchor Points} & & \SetCell[c=2]{c}{DiffUse} & & \SetCell[c=2]{c}{\makecell{MAD \\ Competition}} & & \SetCell[c=2]{c}{\makecell{``Golden'' ranking\\(Chatbot Arena)}} & \\
\cmidrule[lr]{2-3}\cmidrule[lr]{4-5}\cmidrule[lr]{6-7}\cmidrule[lr]{8-9}\cmidrule[lr]{10-11}\cmidrule[lr]{12-13}\cmidrule[lr]{14-15} & Rank & Elo & Rank & Elo & Rank & Elo & Rank & Elo & Rank & Elo & Rank & Elo & Rank & Accuracy \\
\midrule 
GPT-4-Turbo    & 1 & 1,080 & 1 & 1,075 & 1 & 1,046 & 1 & 1,127 & 1 & 1,129 & 1 & 1,086 & 1 & 1,250\\
OpenChat-3.5   & 3 & 993   & 4 & 1,005 & 4 & 1,026 & 3 & 1,011 & 3 & 1,028 & 2 & 1,028 & 2 & 1,091\\
WizardLM-13B   & 5 & 985   & 5 & 988   & 5 & 995   & 2 & 1,037 & 4 & 989   & 3 & 1,022 & 3 & 1,068\\
GPT-3.5-Turbo  & 4 & 988   & 2 & 1,033 & 2 & 1,035 & 7 & 941   & 5 & 942   & 4 & 1,010 & 4 & 1,059\\
Vicuna-13B     & 6 & 983   & 6 & 974   & 6 & 942   & 6 & 944   & 7 & 931   & 5 & 990   & 5 & 1,042\\
Qwen-14B-Chat  & 2 & 1,038 & 3 & 1,007 & 3 & 1,030 & 4 & 993   & 2 & 1,044 & 6 & 954   & 6 & 1,035\\
ChatGLM3-6B    & 7 & 932   & 7 & 919   & 7 & 925   & 5 & 946   & 6 & 938   & 7 & 910   & 7 & 955\\

\bottomrule
\end{tblr}
}
\label{tab:baselines-write}
\vspace{-0.5em}  
\end{table*}

\subsection{Sampling Algorithm Comparison}\label{tab:algorithm}
In Section~\ref{subsec:sampling_comparsion}, we evaluate our adaptive sampling method based on MAD Competition against three alternatives in the \textit{reasoning} (Table \ref{tab:baselines-reason}) and \textit{writing} (Table \ref{tab:baselines-write}) tasks. Below is a concise summary of each baseline and its configuration:
\begin{itemize}
    \item \textbf{DiffUse}~\cite{ashury2024label}
    \begin{itemize}
        \item \textbf{Method}: Cluster the embedding‐difference vectors between each pair of model responses; estimate the expected ranking by drawing instructions from each cluster.
        \item \textbf{Setting}: $10$ clusters per model pair; $3$ prompts per cluster $\rightarrow$ $\binom{7}{2}\times10\times3=630$ instructions.
    \end{itemize}
        \item \textbf{Anchor Points}~\cite{vivek2023anchor}
        \begin{itemize}
            \item \textbf{Method}: Apply $K$-Medoids to select a small set of ``anchor'' instructions.
            \item \textbf{Setting}: $K=10$ anchors per model pair $\rightarrow$ $\binom{7}{2}\times 10 = 210$ instructions.
        \end{itemize}
        \item \textbf{KL and Cross-Entropy Sampling}~\cite{boubdir2023prompts}
        \begin{itemize}
            \item \textbf{Method}: Identical to MAD Competition's protocol except that semantic discrepancy is measured by KL divergence or cross‐entropy on token log probabilities.
            \item \textbf{Setting}: $10$ prompts per model pair $\rightarrow$ $\binom{7}{2}\times 10 = 210$ instructions.
        \end{itemize}
\end{itemize}

Despite their differing sample budgets, all alternatives fall short of MAD Competition’s accuracy in recovering the golden ranking (Tables \ref{tab:baselines-reason} and \ref{tab:baselines-write}). We attribute their failures to two key limitations: inefficient sampling and biased/uninformative discrepancy measures. DiffUse demands a relatively large sample count simply to approximate its cluster‐based expectation, and performance degrades sharply if fewer samples are used. The Anchor Points method likewise needs far more than $10$ anchors per pair to represent complex response distributions adequately. Moreover, its medoid selection biases toward dataset geometry rather than evaluation relevance. KL divergence and cross-entropy on token log probabilities do not reliably reflect substantive differences in response quality and thus yield rankings no better than random sampling.
In contrast, MAD Competition achieves high‐fidelity ranking estimates with a minimal, unbiased sample set, demonstrating both statistical efficiency and robustness across tasks.

\begin{figure*}[!t]
    \centering
    \subfloat[Overall]{\includegraphics[width=0.38\linewidth]{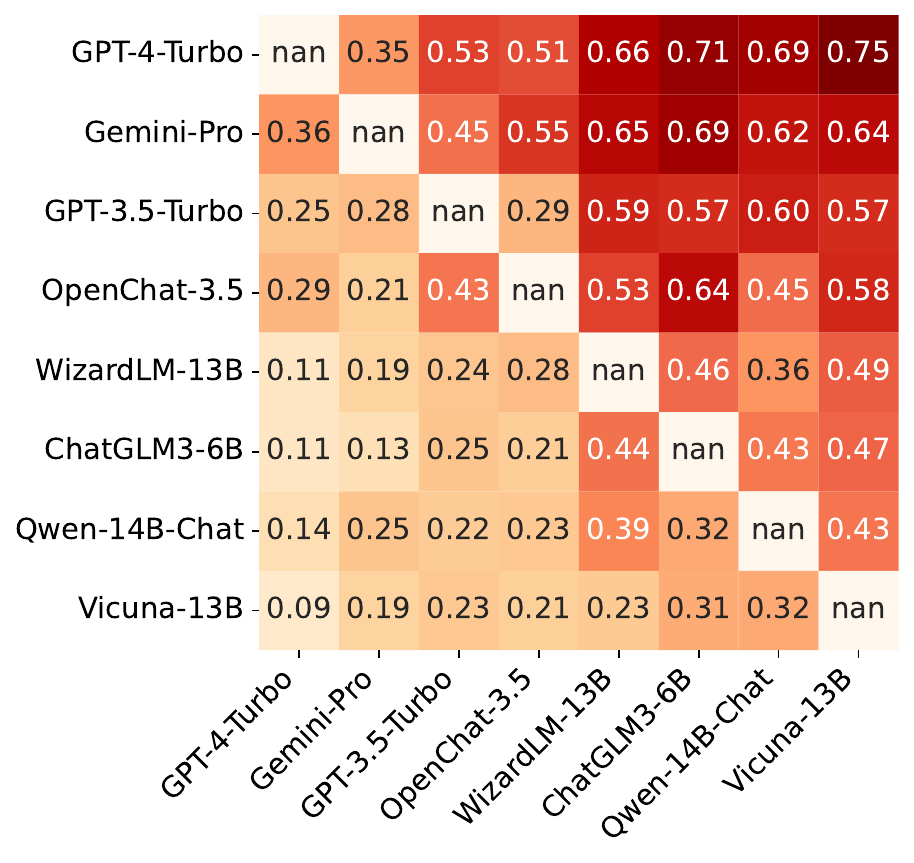}}\\
    \subfloat[Scientific knowledge understanding]{\includegraphics[width=0.38\linewidth]{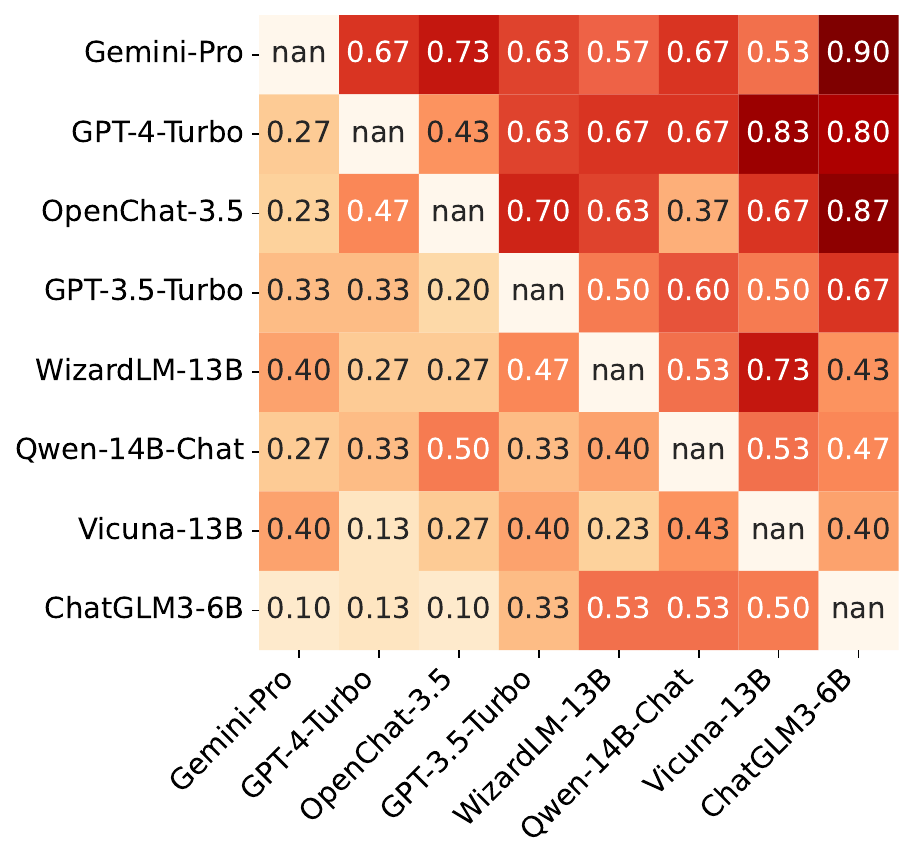}}\hspace{1em}
    \subfloat[Mathematical reasoning]{\includegraphics[width=0.38\linewidth]{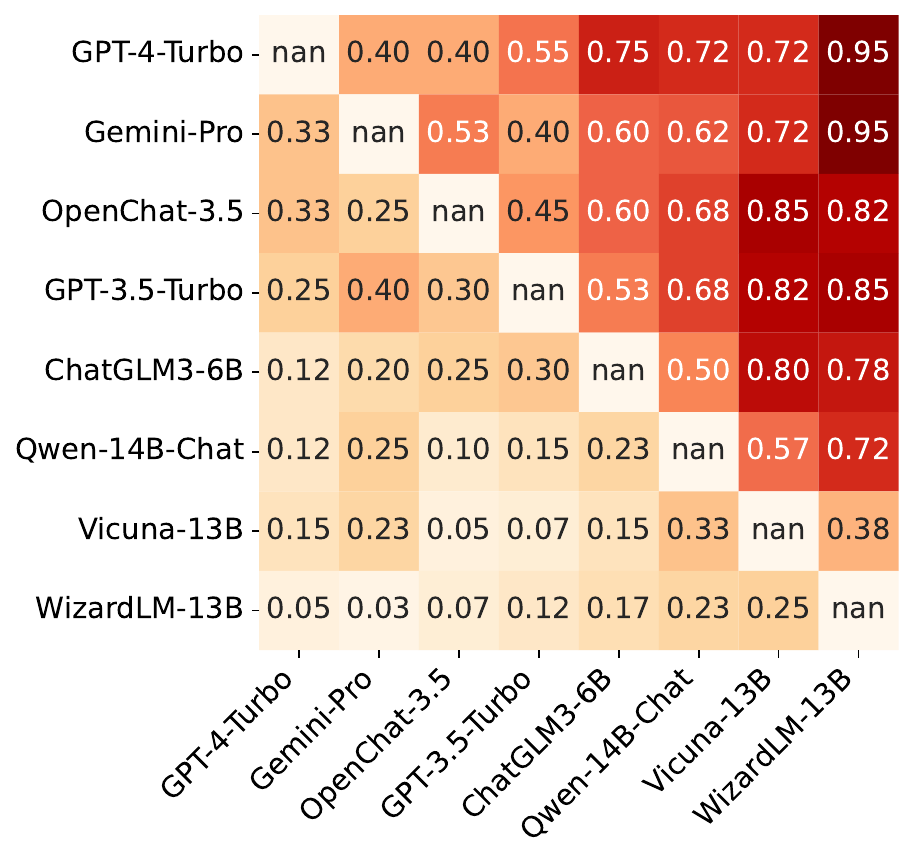}}\\
    \subfloat[Creative and functional writing]{\includegraphics[width=0.38\linewidth]{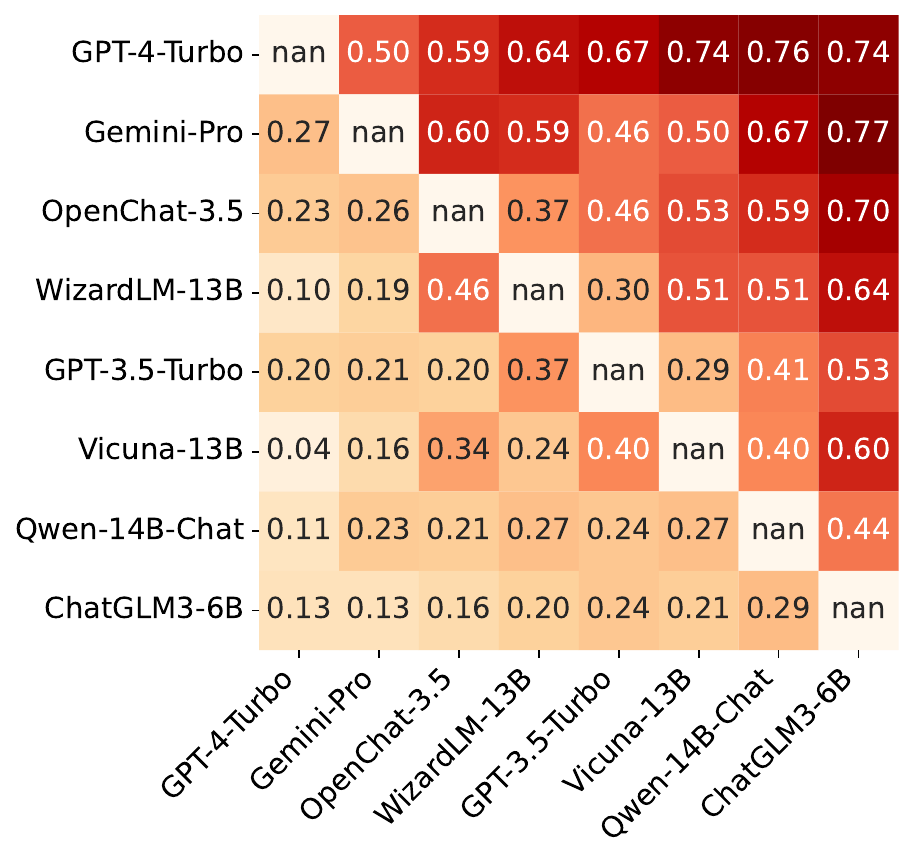}}\hspace{1em}
    \subfloat[Code generation and explanation]{\includegraphics[width=0.38\linewidth]{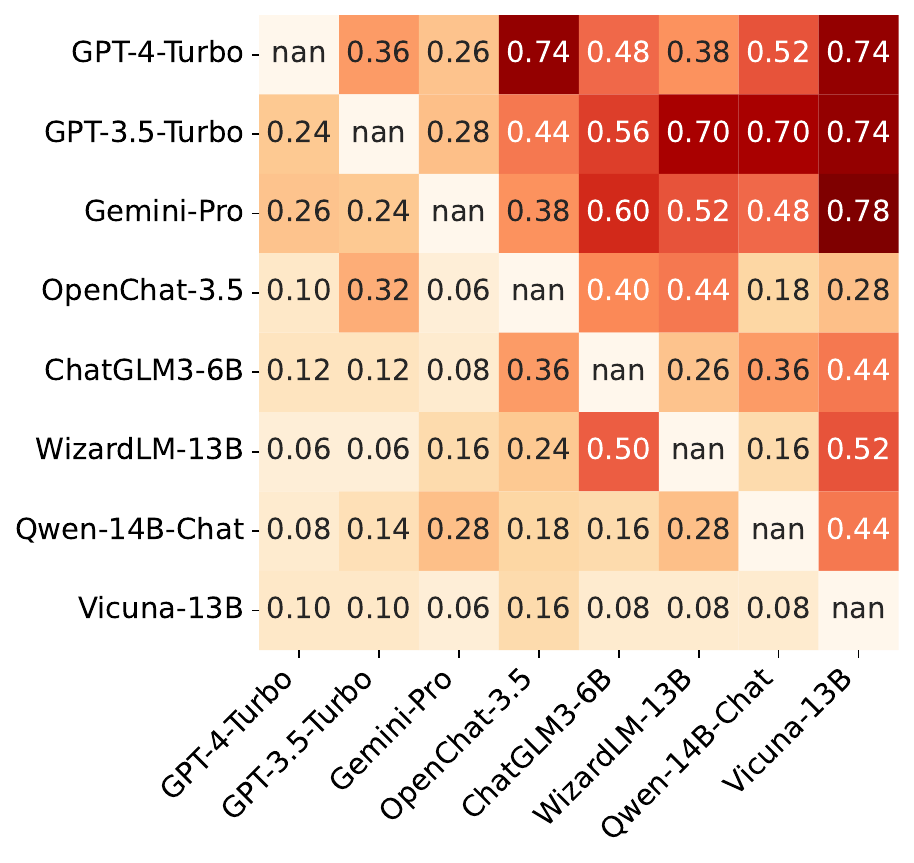}}
    \caption{Heatmaps of pairwise win rates between model pairs for the overall evaluation and four individual tasks. Each cell shows the proportion of head-to-head victories for the model on the vertical axis against the model on the horizontal axis; darker red indicates a higher win rate. Note that tied outcomes do not contribute to either direction.}
    \label{fig:pairwise}
\end{figure*}

\begin{table*}[t]
\centering
 \renewcommand\arraystretch{1.25} 
    \small
    \caption{Some strengths and weaknesses of LLMs discovered by our sample-efficient evaluation method.}
	\begin{tabularx}{\textwidth}{>{\hsize=0.25\hsize}X
                                     >{\hsize=0.68\hsize}X
                                     >{\hsize=0.8\hsize}X} 
		\toprule
\makecell[c]{Model} & \makecell[c]{Strength} & \makecell[c]{Weakness} \\ 
\midrule
GPT-4-Turbo 
& 
{\sethlcolor{white!20!green!30}\hl{Reasoning}}: clear reasoning logic.
\newline
{\sethlcolor{white!20!magenta!30}\hl{Writing}}: powerful writing capability.
\newline
{\sethlcolor{white!20!cyan!30}\hl{Coding}}: high success rate in code execution.
&  
{\sethlcolor{yellow!50}\hl{Others}}: laziness, slight deviation in instruction comprehension.
\\ \hline
Gemini-Pro \& OpenChat-3.5
& 
{\sethlcolor{white!20!green!30}\hl{Reasoning}}: clear reasoning logic.
\newline
{\sethlcolor{white!20!magenta!30}\hl{Writing}}: strong writing capability.
&  
{\sethlcolor{white!20!green!30}\hl{Reasoning}}: limited ability in complex arithmetic computations.
\newline
{\sethlcolor{white!20!cyan!30}\hl{Coding}}: insufficient accuracy in writing complex code.
\\ \hline
GPT-3.5-Turbo
& 
{\sethlcolor{white!20!cyan!30}\hl{Coding}}: strong coding proficiency.
& 
{\sethlcolor{white!20!orange!30}\hl{Understanding}}: short reply length, lack of detailed analysis.
\newline
{\sethlcolor{yellow!50}\hl{Others}}: laziness.
\\ \hline
WizardLM-13B
& 
{\sethlcolor{white!20!magenta!30}\hl{Writing}}: relatively strong writing capability.
& 
{\sethlcolor{white!20!green!30}\hl{Reasoning}}: unclear reasoning logic, weak arithmetic ability.
\newline
{\sethlcolor{white!20!cyan!30}\hl{Coding}}: limited coding proficiency.
\\ \hline
Qwen-14B
&
{\sethlcolor{white!20!cyan!30}\hl{Coding}}: relatively high success rate in execution.
& 
{\sethlcolor{white!20!green!30}\hl{Reasoning}}: limited ability in complex arithmetic computations.
\newline
{\sethlcolor{white!20!magenta!30}\hl{Writing}}: short reply length, simple content.
\newline
{\sethlcolor{yellow!50}\hl{Others}}: (somewhat) laziness.
\\ \hline
ChatGLM3-6B
& 
{\sethlcolor{white!20!green!30}\hl{Reasoning}}: relatively rigorous logic, \newline relatively accurate arithmetic operations.
& 
{\sethlcolor{white!20!orange!30}\hl{Understanding}}: limited knowledge recall, lack of detailed explanation.
\newline
{\sethlcolor{white!20!green!30}\hl{Reasoning}}: limited ability in complex arithmetic computations.
\newline
{\sethlcolor{white!20!magenta!30}\hl{Writing}}: short reply length, simple content.
\newline
{\sethlcolor{white!20!cyan!30}\hl{Coding}}: low success rate in code execution, disregard for instruction requirements.
\newline
{\sethlcolor{yellow!50}\hl{Others}}: laziness.
\\ \hline
Vicuna-13B
& 
{\sethlcolor{white!20!magenta!30}\hl{Writing}}: exceptional instruction-following capability.
& 
{\sethlcolor{white!20!orange!30}\hl{Understanding}}: lack of expertise in explanation, moderate knowledge reserve.
\newline
{\sethlcolor{white!20!green!30}\hl{Reasoning}}: limited computational and reasoning capabilities.
\newline
{\sethlcolor{white!20!cyan!30}\hl{Coding}}: low success rate in code execution, disregard for instruction requirements.
\\
    \bottomrule
	\end{tabularx}
 \label{tab:overview}%
\end{table*}

\begin{table*}[t]
    \centering
    \caption{Instruction evolution prompt for \textit{scientific knowledge understanding}.}
    \vspace{-0.1cm}
    \small
    \rowcolors{1}{lightgray}{lightgray}
    \begin{tabularx}{\textwidth}{|X|}
\hline
\vspace{0.1cm}
You are a brilliant assistant. Your goal is to draw inspiration from \#Given Prompt\# to create a brand-new prompt which is used to evaluate the \textbf{domain knowledge} of a college human student.\newline
The new prompt must be reasonable, unambiguous, and must be understood by humans.\newline
Your response should include `new\_prompt' and `answer' in the following format:\newline
\{\{\newline
  "new\_prompt": "The new prompt for domain knowledge. Try your best to focus on subject basic knowledge and theorems.",\newline
  "answer": "The answer of the new prompt."\newline
\}\}\newline
{Output the response in JSON}.\newline
\newline
The new \#Created Prompt\# should also belong to the similar domain as \#Given Prompt\#.\newline
The question in \#Created Prompt\# should focus on subject basic knowledge and theorems that can evaluate a college student.\newline
The LENGTH and difficulty level of \#Created Prompt\# should be similar to that of \#Given Prompt\#.\newline
\newline
\#Given Prompt\#:\newline
\{instruction\}\newline
\newline
\#Created Prompt\#:\newline
Now, output your response with `new\_prompt' and `answer' in the above format:
\vspace{0.2cm}\\
\hline
    \end{tabularx}
    \label{tab:prompt-s}
\end{table*}%

\begin{table*}[!htbp]
    \centering
    \caption{Instruction evolution prompt for \textit{mathematical reasoning}.}
    \small
    \rowcolors{1}{lightgray}{lightgray}
    \begin{tabularx}{\textwidth}{|X|}
\hline
\vspace{0.1cm}
You are a brilliant assistant. Your goal is to draw inspiration from the question and the associated answer in \#Given Prompt\# to create a brand-new prompt which is used to evaluate the {\textbf{math reasoning capability}} of a primary school student.\newline
The new prompt must be reasonable, unambiguous, and must be understood by humans.\newline
Your response should include `question' and `answer' in the following format:\newline
\{\{\newline
  "question": "The new question for math reasoning. Try your best to follow the same difficulty level and a similar length.",\newline
  "answer": "The step-by-step answer of the new question."\newline
\}\}\newline
{Output the response in JSON}.\newline
\newline
The new question and the step-by-step answer in \#Created Prompt\# should follow the same format as \#Given Prompt\#.\newline
The question in \#Created Prompt\# should focus on grade school math problems.\newline
The LENGTH and difficulty level of \#Created Prompt\# should be similar to that of \#Given Prompt\#.\newline
\newline
\#Given Prompt\#:\newline
Question:\newline
\{instruction\}\newline
Answer:\newline
\{output\}\newline
\newline
\#Created Prompt\#:\newline
Now, output your response with `question' and `answer' in the above format:
\vspace{0.2cm}\\
\hline
    \end{tabularx}
    \label{tab:prompt-r}
\end{table*}%
\begin{table*}[!htbp]
    \centering
    \caption{Instruction evolution prompt for \textit{creative and functional writing}.}
    \small
    \rowcolors{1}{lightgray}{lightgray}
    \begin{tabularx}{\textwidth}{|X|}
\hline
\vspace{0.1cm}
You are a brilliant assistant. Your goal is to draw inspiration from \#Given Prompt\# to create a brand-new prompt with SAME FORMAT (i.e., same structure and number of sentences) which is used to evaluate the {\textbf{writing ability}} of a human student. The new prompt must be reasonable and must be easily understood by humans.\newline
You should design a more rare scenerio or topic that are totally DIFFERENT from \#Given Prompt\# but has practical significance.\newline
The LENGTH and difficulty level of \#Created Prompt\# should be extremely similar to that of \#Given Prompt\#.\newline
`\#Given Prompt\#', `\#Created Prompt\#', `given prompt' and `created prompt' are not allowed to appear in \#Created Prompt\#.\newline
\newline
\#Given Prompt\#:\newline
\{instruction\}\newline
\newline
\#Created Prompt\#:\newline
The new prompt is: \newline
\\
\hline
    \end{tabularx}
    \label{tab:prompt-w}
\end{table*}%

\begin{table*}[ht]
\centering
\caption{Instruction evolution prompt for \textit{code generation and explanation}.}
\small
\rowcolors{1}{lightgray}{lightgray}
\begin{tabularx}{\textwidth}{|X|}
\hline
\vspace{0.1cm}
You are a brilliant assistant. Your goal is to add some constraints to the Python question in \#Given Prompt\# to make the rewritten prompt a bit more challenging. The rewritten prompt is used to evaluate the {\textbf{coding ability}} of a human student. The rewritten prompt must be reasonable and must be easily understood by humans.\newline
Your response should include `new\_prompt' and `answer' in the following format:\newline
\{\{\newline
  "new\_prompt": "The rewritten Python prompt with two constraints.",\newline
  "answer": "The Python code and brief code explanation to the rewritten prompt. Ten to Thirty lines are recommended."\newline
\}\}\newline
{Output in JSON}.\newline
\newline
The rewritten prompt should be different from the scenarios in \#Given Prompt\#. You should add two of the following constraints to the rewritten prompt:\newline
1. Limit the lines of the code, e.g., `Write a $10$-line Python code'. Ten to thirty lines are recommended and preferred.\newline
2. Require high efficiency (time or space) of the code, e.g., `Write a sufficiently efficient Python code'.\newline
3. Force the use of a certain Python library, e.g., `Write a Python code ... using the library NumPy'.\newline
4. Limit the complexity of the code, e.g., `Write a Python code ... easy to read for freshman'.\newline
\newline
The rewritten prompt can only add ten to twenty words into \#Given Prompt\#. You should try your best not to make the new prompt become verbose.\newline
\newline
\#Given Prompt\#:\newline
\{instruction\}\newline
\newline
\#Response\#:\newline
Now, choose two constraints and create a new prompt: 
\vspace{0.2cm}\\
\hline
\end{tabularx}
\label{tab:prompt-c}
\end{table*}%


\begin{table*}[ht]
\color{ADD}
    \centering
    \captionsetup{labelfont={color=ADD},textfont={color=ADD}}
    \caption{Prompt used by LLM-based evaluators for the  \textit{understanding} task.}
    \small
    \rowcolors{1}{lightgray}{lightgray}
    \begin{tabularx}{\textwidth}{|X|}

\hline
\vspace{0.1cm}

Please act as an impartial judge and evaluate the quality of the responses provided by two AI assistants to the user question displayed below. You are given a reference answer to the user question. You should choose the assistant that follows the user's instructions and answers the user's question better. Your evaluation should consider the following factors:\newline\newline

1. \textbf{Accuracy}: Whether the answer is correct based on the reference answer.\newline
2. \textbf{Core Knowledge Coverage}: Whether the answer covers the core knowledge points of the question.\newline
3. \textbf{Logical Consistency}: Whether the answer is logically clear and presents the arguments coherently.\newline
4. \textbf{Clarity of Expression}: Whether the answer uses precise language and is easy to understand.\newline
5. \textbf{Relevance}: Whether the answer stays focused on the question without deviating from the topic.
\newline\newline
If both assistants provide correct answers that cover the core knowledge points, it should be considered a tie (C), regardless of the length or detail of the responses. Only choose a winner if one assistant's response is clearly superior in terms of the evaluation factors.
\newline\newline
Begin your evaluation by comparing the two responses and provide a short explanation. Avoid any position biases and ensure that the order in which the responses are presented does not influence your decision. Do not allow the length of the responses to influence your evaluation. Be as objective as possible. Output "A" if assistant A is better, "B" if assistant B is better, and "C" stands for a tie. Output your final verdict by strictly following this format:\newline
\{\newline
  "explanation": "The short explanation of your evaluation.",\newline
  "winner": "A, B or C."\newline
\}\newline
Output this format in JSON.
\newline\newline
[The Start of User Question]\newline
\{question\}\newline
[The End of User Question]\newline
\newline
[The Start of Reference Answer]\newline
\{answer\}\newline
[The End of Reference Answer]\newline
\newline
[The Start of Assistant A's Answer]\newline
\{response\_1\}\newline
[The End of Assistant A's Answer]\newline
\newline
[The Start of Assistant B's Answer]\newline
\{response\_2\}\newline
[The End of Assistant B's Answer]

\vspace{0.2cm}
\\
\hline

\end{tabularx}
\label{tab:evaluator_prompt_under}
\end{table*}

\begin{table*}[ht]
\color{ADD}
    \centering
    \captionsetup{labelfont={color=ADD},textfont={color=ADD}}
    \caption{Prompt used by LLM-based evaluators for the \textit{reasoning} task.}
    \small
    \rowcolors{1}{lightgray}{lightgray}
    \begin{tabularx}{\textwidth}{|X|}

\hline
\vspace{0.1cm}
Please act as an impartial judge and evaluate the quality of the responses provided by two AI assistants to the user question displayed below. You are given a reference answer to the user question. You should choose the assistant that follows the user's instructions and answers the user's question better. Your evaluation should consider the following factors:
\newline\newline
1. \textbf{Accuracy}: Whether the answer is correct based on the reference answer.\newline
2. \textbf{Logical Consistency}: Whether the answer is logically clear and presents the arguments coherently.\newline
3. \textbf{Clarity of Expression}: Whether the answer uses precise language and is easy to understand.
\newline\newline
If both assistants provide the same final answer, it should be considered a tie (C), regardless of the length or detail of the responses. Only choose a winner if one assistant's response is clearly superior in terms of the evaluation factors, e.g., one achieves the correct answer while the other fails.
\newline\newline
Begin your evaluation by comparing the two responses and provide a short explanation. Avoid any position biases and ensure that the order in which the responses are presented does not influence your decision. Do not allow the length of the responses to influence your evaluation. Be as objective as possible. Output "A" if assistant A is better, "B" if assistant B is better, and "C" stands for a tie. Output your final verdict by strictly following this format:\newline
\{\newline
  "explanation": "The short explanation of your evaluation.",\newline
  "winner": "A, B or C."\newline
\}\newline
Output this format in JSON.
\newline\newline
[The Start of User Question]\newline
\{question\}\newline
[The End of User Question]\newline
\newline
[The Start of Reference Answer]:\newline
\{answer\}\newline
[The End of Reference Answer]\newline
\newline
[The Start of Assistant A's Answer]\newline
\{response\_1\}\newline
[The End of Assistant A's Answer]\newline
\newline
[The Start of Assistant B's Answer]\newline
\{response\_2\}\newline
[The End of Assistant B's Answer]

\vspace{0.2cm}
\\
\hline

\end{tabularx}
\label{tab:evaluator_prompt_reason}
\end{table*}

\begin{table*}[ht]
\color{ADD}
    \centering
    \captionsetup{labelfont={color=ADD},textfont={color=ADD}}
    \caption{Prompt used by LLM-based evaluators for the \textit{writing} task.}
    \small
    \rowcolors{1}{lightgray}{lightgray}
    \begin{tabularx}{\textwidth}{|X|}

\hline
\vspace{0.1cm}
Please act as an impartial judge and evaluate the quality of the responses provided by two AI assistants to the user question displayed below. You should choose the assistant that follows the user's instructions and answers the user's question better. 
Your evaluation should consider the following factors:
\newline\newline
1. \textbf{Accuracy}: How well the response matches the user's instructions and stays on topic.\newline
2. \textbf{Creativity}: The uniqueness of the perspective and the ability to engage the reader.\newline
3. \textbf{Logic and Structure}: Clarity of structure with a clear beginning and ending, and logical flow of ideas.\newline
4. \textbf{Language Expression}: Clarity of language, richness of vocabulary, and appropriate use of rhetorical devices.\newline
5. \textbf{Detail and Depth}: Provision of sufficient details to support main points and depth of exploration of the topic.
\newline\newline
Begin your evaluation by comparing the two responses and provide a short explanation. Avoid any position biases and ensure that the order in which the responses are presented does not influence your decision. Do not allow the length of the responses to influence your evaluation. Be as objective as possible. Output "A" if assistant A is better, "B" if assistant B is better, and "C" stands for a tie. A tie should be determined based on the following criteria:
\newline\newline
1. \textbf{Similar Quality}: When the two responses are close in performance without any significant difference. Performance is measured by quality, and is not related to the length of the responses.\newline
2. \textbf{Similar Core Content Match}: When the performance of both responses is close to meeting the main requirements of the user's instructions.
\newline
3. \textbf{Complementary Strengths}: When each response excels (or fails) in some different aspects, but their overall quality is comparable.
\newline\newline
DO NOT make the judgment too strict. You can output "C" just because they are similar.
\newline
Output your final verdict by strictly following this format:\newline
\{\newline
  "explanation": "The short explanation of your evaluation.",\newline
  "winner": "A, B or C."\newline
\}\newline
Output this format in JSON.
\newline\newline
[The Start of User Question]\newline
\{question\}\newline
[The End of User Question]\newline
\newline
[The Start of Assistant A's Answer]\newline
\{response\_1\}\newline
[The End of Assistant A's Answer]\newline
\newline
[The Start of Assistant B's Answer]\newline
\{response\_2\}\newline
[The End of Assistant B's Answer]

\vspace{0.2cm}
\\
\hline

\end{tabularx}
\label{tab:evaluator_prompt_write}
\end{table*}

\begin{table*}[ht]
\color{ADD}
    \centering
    \captionsetup{labelfont={color=ADD},textfont={color=ADD}}
    \caption{Prompt used by LLM-based evaluators for the \textit{coding} task.}
    \small
    \rowcolors{1}{lightgray}{lightgray}
    \begin{tabularx}{\textwidth}{|X|}

\hline
\vspace{0.1cm}
Please act as an impartial judge and evaluate the quality of the responses provided by two AI assistants to the user question displayed below. You should choose the assistant that follows the user's instructions and answers the user's question better. Your evaluation should consider the following factors:\newline\newline 
1. \textbf{Accuracy}: Whether the code correctly implements the user's requirements or the explanation is accurate.\newline
2. \textbf{Level of Detail}: Whether the code or explanation is detailed enough to meet the user's needs.\newline
3. \textbf{Logical Consistency}: Whether the code structure or explanation logic is clear and consistent.\newline
4. \textbf{Code Quality}: The quality of the code, including readability, efficiency, and maintainability.\newline
5. \textbf{Creativity and Reasonableness}: Whether the code implementation or explanation shows creativity and is reasonable.
\newline\newline
Begin your evaluation by comparing the two responses and provide a short explanation. Avoid any position biases and ensure that the order in which the responses are presented does not influence your decision. Do not allow the length of the responses to influence your evaluation. Be as objective as possible. Output "A" if assistant A is better, "B" if assistant B is better, and "C" stands for a tie. A tie should be determined based on the following criteria:
\newline\newline
- If both assistants' responses are good across all criteria without significant quality differences, it should be judged as a tie.\newline
- If both responses have significant errors (despite that the errors may be different), it should be judged as a tie.\newline
- If both responses have no significant differences in core accuracy and logical consistency, even if there are slight differences in detail or code quality, it should be judged as a tie.\newline
- If both responses follow different but reasonable, effective approaches, it should be judged as a tie.
\newline\newline
Output your final verdict by strictly following this format:\newline
\{\newline
  "explanation": "The short explanation of your evaluation.",\newline
  "winner": "A, B or C."\newline
\}\newline
Output this format in JSON.
\newline\newline
[The Start of User Question]\newline
\{question\}\newline
[The End of User Question]\newline
\newline
[The Start of Assistant A's Answer]\newline
\{response\_1\}\newline
[The End of Assistant A's Answer]\newline
\newline
[The Start of Assistant B's Answer]\newline
\{response\_2\}\newline
[The End of Assistant B's Answer]

\vspace{0.2cm}
\\
\hline

\end{tabularx}
\label{tab:evaluator_prompt_code}
\end{table*}

\subsection{Pairwise Comparison Results}
\label{ap:pair-rank}
Figure~\ref{fig:pairwise} presents the pairwise comparison results. Across both the overall assessment and each individual task, GPT-4-Turbo and Gemini-Pro emerge as the top two models, outperforming all other competitors by a substantial margin.

\subsection{Case Studies} 
\label{ap:case}
In this subsection, we present illustrative cases that corroborate the trends summarized in Table~\ref{tab:overview}.

\paragraph{Scientific knowledge understanding}  
Table~\ref{tab:case-s-1} compares proprietary and open-source models on domain-specific questions. Proprietary systems consistently outperform open-source counterparts, reflecting their larger knowledge repositories and more effective retrieval mechanisms. In Table~\ref{tab:case-s-2}, both OpenChat-3.5 and GPT-3.5-Turbo correctly identify the key scientific facts, yet humans prefer OpenChat-3.5 because it delivers richer and deeper analytical commentary.

\paragraph{Mathematical reasoning}  
Table~\ref{tab:case-r-1} highlights how divergent reasoning paths between WizardLM-13B and OpenChat-3.5 can yield different answers on simple arithmetic tasks. This variability underscores WizardLM’s relative deficit in systematic, step-by-step deduction. Table~\ref{tab:case-r-2} further reveals that minor lapses in arithmetic precision by WizardLM-13B lead to incorrect final results, indicating a need to strengthen its core computational routines.

\paragraph{Creative and functional writing}  
Table~\ref{tab:case-w-1} demonstrates clear human preferences for responses that incorporate substantive content (\eg, illustrative examples or nuanced argumentations), suggesting that response length correlates positively with perceived quality.

\paragraph{Code generation and explanation}  
Although evaluators consider readability and adherence to instructions, accuracy remains paramount. In Table~\ref{tab:case-c-1}, Vicuna-13B annotates its code thoroughly, but human raters prefer Gemini-Pro because it alone delivers reliably correct, executable solutions within the specified constraints.

\paragraph{Counterexamples of GPT-4-Turbo}  
Despite its top ranking overall, GPT-4-Turbo exhibits notable shortcomings in certain contexts:
\begin{itemize}
  \item \textbf{Algorithm explanation} (Table~\ref{tab:counterexample-s-1}). OpenChat-3.5’s inclusion of a complete Dijkstra implementation renders its explanation more intuitive and actionable for human readers, suggesting that model performance should be judged not only by correctness but also by pedagogical clarity.
  \item \textbf{Instruction comprehension} (Tables~\ref{tab:counterexample-r-1} and \ref{tab:counterexample-w-1}). GPT-4-Turbo occasionally misinterprets prompts---for example, labeling Rosalind Franklin’s contributions as ``underappreciated,'' contrary to historical consensus---indicating room to improve contextual sensitivity and factual alignment.
  \item \textbf{Code constraint} (Table~\ref{tab:counterexample-c-1}). GPT-4-Turbo sometimes generates solutions that exceed prescribed length limits or contain subtle logical errors, highlighting persistent challenges in enforcing user-specified coding guidelines.
  \item \textbf{Instruction adherence} (Table~\ref{tab:counterexample-w-2}). At times, GPT-4-Turbo omits direct answers or skims over critical request elements, a phenomenon we describe as ``response laziness,'' underscoring the importance of robust instruction-following mechanisms.
\end{itemize}

These counterexamples illustrate that, although GPT-4-Turbo leads on average, future high-performance LLMs must dynamically adapt response formats to specific tasks, enhance fine-grained comprehension, and ensure both factual and procedural fidelity.

\begin{table*}[!htbp]
	\centering
    \captionsetup{labelfont={color=ADD},textfont={color=ADD}}
    \caption{Top-$10$ instructions chosen by different sampling strategies.}
    \scriptsize
	\begin{tabularx}{\textwidth}{>{\centering}cXXXX}
		\toprule
\makecell{\small Order} & \makecell{\small Random} & \makecell{\small KL Divergence} & \makecell{\small Cross-Entropy} & \makecell{\small MAD Competition} \\ 
\midrule
\multicolumn{1}{c}{1} 
& You are asked to offer valuable \textcolor[RGB]{0,163,245}{\textbf{suggestions}}, relevant statistics, or elucidation on the issue outlined below...
& Craft a \textcolor{red}{\textbf{limerick}} centered around a family embarking on a submarine voyage to explore the Mariana Trench, capturing the whimsical tone of Lewis Carroll.
& Compose four \textcolor[RGB]{41,114,244}{\textbf{essays}} on the subject of climate change adaptation strategies, each with a minimum of 300 words.
& You have been tasked with offering informed perspectives, analyses, or elucidations to \textcolor{green}{\textbf{address the following question}}...\\ \hline
2 
& Compose a compelling \textcolor[RGB]{41,114,244}{\textbf{essay}} on the importance of learning a second language.
& Can you compose \textcolor{red}{\textbf{limericks}} for various renewable energy sources? Begin with solar power.
& Develop a Market \textcolor[RGB]{245,196,0}{\textbf{Analysis}} Report, following APA referencing style, with six sections evaluating a different emerging technology...
& ... \textcolor[RGB]{0,163,245}{\textbf{Suggest}} a few alternative historical items that could temporarily fill this space and enhance the exhibition's narrative.
\\ \hline
3 
& You have been assigned to \textcolor[RGB]{245,196,0}{\textbf{explore}} the environmental implications of a modern dilemma...
& Compose a \textcolor{red}{\textbf{limerick}} that uses alliteration to add humor to the described situation.\newline A clumsy chef spills the soup.
& Craft a 1900-word science fiction \textcolor[RGB]{154,56,215}{\textbf{story}} set in an underwater city.
& Craft a one-page \textcolor[RGB]{41,114,244}{\textbf{essay}} discussing the importance of preserving linguistic diversity in the age of global communication.
\\ \hline
4 
& Draft a dialogue for a historical fiction \textcolor[RGB]{154,56,215}{\textbf{novella}}.
& Craft a \textcolor{red}{\textbf{poem}} about Maya and Leo launching an initiative to clean up the ocean, using advanced robotic technology to tackle the problem of marine pollution...
& Craft four separate evaluations on the impact of technological advancements on various educational methodologies, ensuring that each evaluation consists of at least 300 words.
& Envision \textcolor[RGB]{248,136,47}{\textbf{adapting}} the discovery of penicillin into an interactive escape room game.
\\ \hline
5 
& Craft a 500-word \textcolor[RGB]{41,114,244}{\textbf{essay}} on the significance of preserving historical languages that are on the verge of extinction.
& Compose a \textcolor{red}{\textbf{poem}} in the voice of Edgar Allan Poe.
& Craft a 1900-word adventure \textcolor[RGB]{154,56,215}{\textbf{story}} set in the midst of a perilous journey across the Sahara Desert in the early 20th century.
& Develop a compelling and nuanced \textcolor[RGB]{154,56,215}{\textbf{backstory}} for my protagonist, a cunning diplomat in a fantasy realm...
\\ \hline
6 
& Craft a \textcolor[RGB]{154,56,215}{\textbf{tale}} about an ethnomusicologist who discovers a remote village where the people communicate using a complex system of rhythm and drumming.
& Compose a \textcolor{red}{\textbf{quatrain}} about a group of scientists traveling to the Arctic Circle to study the aurora borealis, reflecting the mystical style of J.R.R. Tolkien.
& Compose a 1900-word historical fiction \textcolor[RGB]{154,56,215}{\textbf{narrative}} situated in a secluded mountain monastery.
& Draft a \textcolor{red}{\textbf{villanelle}} utilizing the traditional ABA ABA ABA ABA ABA ABAA rhyme scheme.
\\ \hline
7 
& Your assignment is to showcase your proficiency in effective and engaging \textcolor{green}{\textbf{marketing writing}} as described below...
& Compose a \textcolor{red}{\textbf{sonnet}} about a group of friends on a hiking adventure in the Swiss Alps...
& Write four unique \textcolor[RGB]{41,114,244}{\textbf{essays}} evaluating the impact of technological advancements on various educational methodologies, ensuring that each essay consists of at least 300 words.
& Reflect on the influence of a scientific discovery you encountered in the past month. Strive to make your evaluation thoughtful and significant.
\\ \hline
8 
& Compose a 500-word \textcolor[RGB]{41,114,244}{\textbf{essay}} examining the difficulties faced when introducing a new public healthcare system in a developing country.
& Compose a \textcolor{red}{\textbf{sonnet}} following the Shakespearean rhyme pattern.
& Compose four distinct \textcolor[RGB]{41,114,244}{\textbf{essays}} analyzing the influence of climate change on different agricultural practices...
& Greetings, could you help me formulate my introduction \textcolor{green}{\textbf{speech}} for the local art gallery opening?
\\ \hline
9
& You are assigned to \textcolor[RGB]{245,196,0}{\textbf{explore}} the following environmental issue by detailing arguments for various viewpoints...
& Craft a \textcolor{red}{\textbf{sonnet}} detailing the adventure of Leo and Harper as they venture into the depths of an unexplored cave system to ...
& Pretend you are an innovative urban planner who has been tasked to present at a global conference on sustainable cities. Compose an \textcolor[RGB]{41,114,244}{\textbf{essay}}...
& You have been selected to demonstrate your expertise in critical \textcolor[RGB]{245,196,0}{\textbf{analysis}} for the situation described below...
\\ \hline
10
& ...Would you be able to supply me with a detailed \textcolor[RGB]{0,163,245}{\textbf{proposal}}?
& Demonstrate your creative \textcolor[RGB]{154,56,215}{\textbf{narrative}} skills with this task...
& Draft a 2000 word \textcolor[RGB]{245,196,0}{\textbf{analysis}} on the influence of community-based recycling programs on reducing municipal waste.
& Compose a brief \textcolor[RGB]{154,56,215}{\textbf{narrative}} beginning with the provided opening line...
\\

    \bottomrule
	\end{tabularx}
 \label{tab:instruct-chosen}%
\end{table*}

\begin{table*}[!htbp]
	\centering
    \caption{Top-$10$ instructions chosen by MAD Competition with and without the diversity term.}
    \small
	\begin{tabularx}{\textwidth}{cXX}
		\toprule
Order & \textbf{without} Diversity & \textbf{with} Diversity \\ 
\midrule
1 
& You have been assigned to evaluate the following technological issue by considering opinions from diverse standpoints.\newline Is the widespread implementation of autonomous vehicles on public roads justifiable?
& You have been assigned to evaluate the following technological issue by considering opinions from diverse standpoints.\newline Is the widespread implementation of autonomous vehicles on public roads justifiable? \\ \hline
2 
& Devise a catchy mnemonic to remember the key elements.\newline An astronomical method for categorizing, observing, and explaining Variable Star Observation Parameters (VSOP).
& I am particularly fascinated by the traditional sounds of Mongolia. Could you provide me with the sheet music for a classic Morin Khuur tune?\\ \hline
3 
& I am particularly fascinated by the traditional sounds of Mongolia. Could you provide me with the sheet music for a classic Morin Khuur tune?
& Devise a catchy mnemonic to remember the key elements.\newline An astronomical method for categorizing, observing, and explaining Variable Star Observation Parameters (VSOP).\\ \hline
4 
& Draft a {\color{red}villanelle} utilizing the traditional ABA ABA ABA ABA ABA ABAA rhyme scheme.
& Draft a {\color{blue}villanelle} utilizing the traditional ABA ABA ABA ABA ABA ABAA rhyme scheme.\\ \hline
5 
& Compose a {\color{red}sonnet} following the Shakespearean rhyme pattern.
& Designing a mascot for your environmental conservation campaign is a crucial step in engaging with the community. Reflecting on the goals and fundamental purpose of your initiative, you are required to conceive a variety of appealing mascot concepts. Take some time to ponder your ideas.\\ \hline
6 
& Can you compose {\color{red}sonnets} that reflect the distinct flavors of various cuisines? Begin with Italian pasta dishes.
& Design a program for a high school science fair. Assign a precise duration for each presentation included in the event.\\ \hline
7 
& Draft a dialogue for a historical fiction novella.
& Write an essay discussing the three main economic theories that explain market behavior.\\ \hline
8 
& Compose a brief narrative in 10-15 lines that encapsulates the experience of conducting a scientific field research in a remote rainforest during a significant weather phenomenon.
& As a dedicated librarian, I cherish the opportunity to introduce young readers to classic literature. I'm currently seeking to diversify the selection of adventure stories in our children's section that highlight female protagonists. Could you recommend a novel featuring a young heroine that is suitable for readers aged 8 to 12?\\ \hline
9
& Are you able to concoct {\color{red}haikus} about different forms of transportation? Start with bicycles.
& Hello, could you craft a narrative in the style of a screenplay that features interactions among characters, set within the universe of Greek mythology, including the figures, Hermes, Apollo, and Medusa.\\ \hline
10
& Hello, could you craft a narrative in the style of a screenplay that features interactions among characters, set within the universe of Greek mythology, including the figures, Hermes, Apollo, and Medusa.
& Outline the progression of major milestones in the field of artificial intelligence in healthcare during the year 2022.\\

    \bottomrule
	\end{tabularx}
 \label{tab:diversity}%
\end{table*}
\begin{table*}[!htbp]
    \centering
    \caption{Prompt used by GPT-4-Turbo as a semantic similarity measure.}
    \small
    \rowcolors{1}{lightgray}{lightgray}
    \begin{tabularx}{\textwidth}{|X|}
\hline
\vspace{0.1cm}
Given two responses, you are asked to evaluate the similarity between the two responses. Your evaluation should be based on the following metrics:\newline
1. \textbf{Task and Theme}: Analyze whether this text pair addresses the same open-ended task and theme. If the tasks or themes are not entirely the same, analyze their similarities and differences.\newline
2. \textbf{Emotion and Semantics}: Conduct an emotion and semantic analysis of this text pair, determining their similarity in emotional polarity (positive, negative, and neutral) and semantic polarity.\newline
3. \textbf{Content Quality}: Analyze the similarity in content quality between this text pair, considering aspects such as incomplete content, nonsensical statements, lack of details, etc.\newline
4. \textbf{Details}: Analyze the similarity in the details of the text content, such as the approach and steps in solving mathematical problems, plot details in stories, etc.\newline
5. \textbf{Language Expression}: Analyze the similarity in the language expression between the two responses, including language style, vocabulary, syntax, rhetorical devices, etc.\newline
You should write an explanation carefully about your evaluation using ALL the metrics above [1]. DO NOT forget any metric in your explanation.\newline
\newline
Next, your similarity evaluation will be integrated into a two-decimal score between 0 and 1 by strictly adhering to the following scoring rubric:\newline
1. \textbf{0.00-0.20}: Extremely low similarity. It means responses met one of the following conditions: (1) The tasks they solved are not the same, or their themes are totally different, (2) They have opposite semantic polarity, (3) They have opposite emotional polarity, (4) There is a huge difference in content quality, i.e. one of the responses contains obvious incompleteness, or one of the responses is nonsense because it simply repeats some sentences.\newline
2. \textbf{0.21-0.40}: Relatively low similarity. It means responses met one of the following conditions: (1) The tasks they solved are the same, their themes have few differences, but are nearly the same, (2) They have relatively low similarity in content quality due to one of the responses contains few repetitive sentences, (3) They have completely different details or linguistic expression, (4) They have relatively low similarity in semantic polarity or emotional polarity.\newline
3. \textbf{0.41-0.60}: Medium similarity. It means responses met all the following conditions: (1) Responses solve exactly the same task, (2) They are similar in theme, semantic polarity, and emotional polarity, (3) There are still some differences in linguistic expression and details.\newline
4. \textbf{0.61-0.80}: Relatively high similarity. It means responses met all the following conditions: (1) Responses solve exactly the same task, and have exactly the same theme, semantic polarity, and emotional polarity, (2) There are few differences in content, linguistic expression, and details, resulting in relatively high similarity.\newline
5. \textbf{0.81-1.00}: Extremely similar. It means basically no difference can be found.\newline
DO NOT VIOLATE ANY DETAIL IN THE SCORING RUBRIC. You should obey that if any of the conditions of `extremely low similarity' and `relatively low similarity' are met, the score segment must be selected.\newline
\newline
You should carefully compare the similarity evaluation with the scoring criteria and select the most appropriate score range from the scoring rubric, then you should provide a specific score based on your evaluation. You should also describe why you selected this score range and how you scored in your explanation [2]. Your explanation MUST be more than 100 words! \newline
\newline
You must only respond with the following two components:\newline
- explanation: "The explanation of your evaluation using all 5 metrics in [1] and scoring in [2], more than 100 words.",\newline
- score: "The similarity score, a two-decimal score."\newline
{Output in JSON} containing the explanation and score.\newline
\newline
Now, please complete the following evaluation:\newline
[response 1]:\newline
\{response\_1\}\newline
\newline
[response 2]:\newline
\{response\_2\}
\vspace{0.2cm}\\
\hline
    \end{tabularx}
    \label{tab:prompt-gpt4}
\end{table*}%
%
\begin{table*}[!htbp]
    \centering
    \caption{For \textit{scientific knowledge understanding}, Gemini-Pro demonstrates significantly better understanding and application capabilities of scientific knowledge, compared to the open-source model Vicuna-13B.}
    \small
    \begin{tabularx}{\textwidth}{XX}
        \toprule
\multicolumn{2}{p{13.5cm}}{\textbf{Instruction:} \newline What are the boundary conditions typically applied to solve the standard diffusion equation in one dimension?} \\ 
\midrule

\textbf{Vicuna-13B Response:\newline} 
The standard one-dimensional diffusion equation is given by:\newline 
$$\frac{\partial c}{\partial t} = D \frac{\partial^2 c}{\partial x^2}$$\newline 
where $c$ is the concentration of diffusing species, $t$ is time, $x$ is position, and $D$ is the diffusion constant.\newline 
The boundary conditions for this equation depend on the specific problem being solved. However, some common boundary conditions are:\newline
1. {\color{red}Periodic boundary conditions}: $c(x, t) = c(x + L, t)$\newline 
2. {\color{red}Reflective boundary conditions}: $c(x, t) = c(x, 0)$\newline
3. {\color{red}Absorbing boundary conditions}: $c(x, t) = 0$\newline
4. {\color{red}Mixed boundary conditions}: a combination of reflective and absorbing boundary conditions.\newline 
It is important to note that the choice of boundary conditions can affect the solution of the diffusion equation, and the appropriate boundary conditions must be chosen based on the specific problem being solved.
& 
\textbf{Gemini-Pro Response:\newline}  
1. {\color{green!80!black}\textbf{Dirichlet boundary condition}}: This condition specifies the value of the solution at the boundaries. It is also known as the fixed boundary condition. Mathematically, it can be expressed as:\newline
$$u(a) = g_1, \quad u(b) = g_2$$\newline    
where $a$ and $b$ are the boundaries of the domain, $u(a)$ and $u(b)$ are the prescribed values of the solution at the boundaries, and $g_1$ and $g_2$ are constants or given functions.\newline
2. {\color{green!80!black}\textbf{Neumann boundary condition}}: This condition specifies the value of the normal derivative of the solution at the boundaries. It is also known as the flux boundary condition. Mathematically, it can be written as:\newline
$$\frac{\partial u}{\partial x}(a) = h_1, \quad \frac{\partial u}{\partial x}(b) = h_2$$\newline
where $h_1$ and $h_2$ are constants or given functions.\newline
3. {\color{green!80!black}\textbf{Robin boundary condition}}: This condition is a combination of the Dirichlet and Neumann boundary conditions. It specifies both the value of the solution and the value of its normal derivative at the boundaries. Mathematically, it can be expressed as:\newline
$$u(a) = g_1, \quad  \frac{\partial u(b)}{\partial x} + \alpha u(b) = h_2$$\newline
where $\alpha$ is a constant.\newline
4. \textbf{Periodic boundary condition}: This condition is used when the solution is expected to have a periodic behavior. It requires the solution to have the same value at the two boundaries, i.e.,\newline
$$u(a) = u(b)$$\newline
These are the common boundary conditions used to solve the standard diffusion equation in one dimension. The specific boundary condition to be applied depends on the physical problem being solved.
\\ \midrule
\multicolumn{2}{X}{\textbf{Human Preference}: Gemini-Pro}\\
\bottomrule
    \end{tabularx}
    \label{tab:case-s-1}
\end{table*}

\begin{table*}[!htbp]
    \centering
    \caption{For \textit{scientific knowledge understanding}, OpenChat-3.5 provides fine-grained explanation while encompassing the required core knowledge, garnering preferences from the majority of human subjects, compared to GPT-3.5-Turbo.}
    \small
    \begin{tabularx}{\textwidth}{XX}
        \toprule
\multicolumn{2}{p{13.5cm}}{\textbf{Instruction:} \newline How does Le Chatelier's principle explain the effect of changing the concentration of a reactant on the position of equilibrium in a chemical reaction? Provide a specific example with the relevant balanced chemical equation and explain your answer.} \\ 
\midrule

\textbf{OpenChat-3.5 Response:\newline} 
Le Chatelier's principle states that when a change is applied to a system at equilibrium, the system will adjust itself to counteract the change in order to regain its original state of equilibrium. In the context of a chemical reaction at equilibrium, this means that if a change is made to one of the reactants or products, the position of the equilibrium will shift in a way that minimizes the effect of the change.\newline One example of this can be observed in the reaction of hydrogen gas (H2) and iodine gas (I2) to form hydrogen iodide (HI). The balanced chemical equation for this reaction is:\newline H2(g) + I2(g) $\leftrightharpoons$ 2HI(g)\newline Initially, the reaction is at equilibrium, and the concentrations of H2, I2, and HI are in a specific ratio. Now, let's say we increase the concentration of hydrogen gas (H2) by adding more H2 to the system. According to Le Chatelier's principle, the system will respond by shifting the equilibrium position to consume some of the excess hydrogen gas and minimize its effect on the reaction.\newline To do this, the forward reaction (H2 + I2 → 2HI) will become more favorable, as it consumes hydrogen gas and forms more HI. This will eventually lead to a new equilibrium state where the concentration of H2 is lower than it was before the change. Conversely, if we were to decrease the concentration of hydrogen gas, the reaction would shift in the opposite direction to consume more HI and form more H2, again restoring the original concentration of H2.\newline In summary, Le Chatelier's principle explains how changing the concentration of a reactant affects the position of equilibrium in a chemical reaction. By increasing the concentration of a reactant, the equilibrium position will shift to consume more of that reactant and minimize its effect. Conversely, decreasing the concentration of a reactant will cause the equilibrium position to shift in the opposite direction, consuming less of that reactant and minimizing its effect.
& 
\textbf{GPT-3.5-Turbo Response:\newline}  
Le Chatelier's principle states that if a system at equilibrium is disturbed by a change in temperature, pressure, or concentration of a component, the system will shift its position of equilibrium to counteract the effect of the disturbance.\newline Let's consider the reaction between nitrogen dioxide (NO2) and dinitrogen tetroxide (N2O4):\newline 2NO2(g) $\leftrightharpoons$ N2O4(g)\newline If we increase the concentration of NO2, according to Le Chatelier's principle, the system will respond by shifting the equilibrium to the left to counteract the increase in NO2 concentration. As a result, more N2O4 will be formed to consume the excess NO2, thereby reducing the concentration of NO2 and restoring equilibrium.\newline Conversely, if we decrease the concentration of NO2, the system will shift the equilibrium to the right to produce more NO2 and consume N2O4, again restoring equilibrium.\newline In both cases, Le Chatelier's principle explains how the position of equilibrium shifts in response to changes in the concentration of a reactant in order to maintain a constant equilibrium condition.
\\ \midrule
\multicolumn{2}{X}{\textbf{Human Preference}: OpenChat-3.5}\\
\bottomrule
    \end{tabularx}
    \label{tab:case-s-2}
\end{table*}%

\begin{table*}[!htbp]
    \centering
    \caption{For \textit{mathematical reasoning}, WizardLM-13B takes an incorrect reasoning path, leading to wrong solutions, in contrast to OpenChat-3.5.}
    \small
    \begin{tabularx}{\textwidth}{XX}
        \toprule
\multicolumn{2}{p{13.5cm}}{\textbf{Instruction:} \newline John wants to build a rectangular garden. He has 60 feet of fencing material. If he wants the length of the garden to be twice the width, how long will each side of the garden be?} \\ 
\midrule

\textbf{OpenChat-3.5 Response:\newline} 
Let's assume the width of the garden is W feet.\newline
Since the length of the garden is twice the width, the length will be 2W feet.\newline 
The perimeter of a rectangle is given by the formula: P = 2(length + width)\newline
In this case, the perimeter is 60 feet, so we can write the equation as:\newline
{\color{green!80!black}60 = 2(2W + W)}\newline
Simplifying the equation, we get: 60 = 2(3W)\newline
{\color{green!80!black}60 = 6W}\newline 
Dividing both sides by 6, we get:\newline
{\color{green!80!black}W = 10}\newline 
So, the width of the garden is 10 feet.\newline
{\color{green!80!black}Since the length is twice the width, the length is 2 * 10 = 20 feet}.\newline
Therefore, each side of the garden will be 10 feet and the other side will be 20 feet.
& 
\textbf{WizardLM-13B Response:\newline}  
To solve this problem, we can use the Pythagorean theorem, which states that in a right-angled triangle, the square of the length of the hypotenuse (the side opposite the right angle) is equal to the sum of the squares of the other two sides. In this case, the length of one side of the garden is equal to the sum of the squares of the other two sides.\newline
Let's denote the width of the garden as 'a' and the length as 'b'. We know that 'b' is equal to 2*'a', so we can rewrite the equation as follows:\newline
$b^2 = 2\times a^2$\newline
{\color{red}Now, we can use the Pythagorean theorem to find the value of 'a'. Let's denote 'c' as the hypotenuse of a right-angled triangle formed by the width and the length of the garden. Then, we can write the equation as:\newline
$c^2 = a^2 + b^2$\newline
Substituting the given values, we get:\newline
$c^2 = 60^2$\newline}
Solving for 'a', we get: $a = \sqrt{60^2}$\newline
Therefore, the width of the garden is $\sqrt{60^2}$ feet.
\\ \midrule
\multicolumn{2}{X}{\textbf{Human Preference}: OpenChat-3.5}\\
\bottomrule
\end{tabularx}
    \label{tab:case-r-1}%
\end{table*}%

\begin{table*}[!htbp]
    \centering
    \caption{For \textit{mathematical reasoning}, WizardLM-13B makes an arithmetic error during the intermediate reasoning process, leading to the eventual collapse of the final result, compared to GPT-3.5-Turbo.}
    \small
    \begin{tabularx}{\textwidth}{XX}
        \toprule
\multicolumn{2}{p{13.5cm}}{\textbf{Instruction:} \newline A bakery sells cupcakes and cookies. Each cupcake costs \$3 and each cookie costs \$2. In one day, the bakery makes \$120 from selling cupcakes and \$80 from selling cookies. How many items did the bakery sell in total in one day?} \\ 
\midrule

\textbf{WizardLM-13B Response:\newline} 
First, we need to determine the total number of items sold by the bakery in one day. We know that the bakery sold 40 cupcakes and 20 cookies. To find the total number of items sold, we can simply add the number of cupcakes and cookies together.\newline
{\color{green!80!black}Step 1: Add the number of cupcakes and cookies}.\newline
{\color{green!80!black}Cupcakes: 40}\newline
{\color{red}Cookies: 20}\newline
{\color{green!80!black}Step 2: Calculate the total number of items sold}.\newline
{\color{green!80!black}Total items sold} = {\color{green!80!black}Cupcakes + Cookies} = {\color{green!80!black}40} + {\color{red}20} = {\color{red}60}\newline
So, the bakery sold 60 items in total in one day.\newline
The answer is: 60.
& 
\textbf{GPT-3.5-Turbo Response:\newline}  
Sure! Let's solve the problem step by step.\newline
Let's represent the number of cupcakes sold as C and the number of cookies sold as K.\newline
We are given the following information:\newline
1. Each cupcake costs \$3, and the bakery makes \$120 from selling cupcakes. This can be represented as 3C = 120.\newline
2. Each cookie costs \$2, and the bakery makes \$80 from selling cookies. This can be represented as 2K = 80.\newline
Now, let's solve the equations to find the values of C and K.\newline
From the first equation, we can solve for C:\newline
3C = 120\newline
{\color{green!80!black}C = 120 / 3 = 40}\newline
From the second equation, we can solve for K:\newline
2K = 80\newline
{\color{green!80!black}K = 80 / 2 = 40}\newline
So, the bakery sold 40 cupcakes and 40 cookies in one day. To find the total number of items sold, we add the number of cupcakes and cookies:\newline
{\color{green!80!black}Total items = 40 cupcakes + 40 cookies = 80 items}\newline
Therefore, the bakery sold a total of 80 items in one day.
\\ \midrule
\multicolumn{2}{X}{\textbf{Human Preference}: GPT-3.5-Turbo}\\
\bottomrule
    \end{tabularx}
    \label{tab:case-r-2}%
\end{table*}%

\begin{table*}[!htbp]
    \centering
    \caption{For \textit{writing}, the response of GPT-4-Turbo has richer content, while ChatGLM3-6B's response not only fails to meet the instruction requirements, but is also lacking in content.}
    \small
    \begin{tabularx}{\textwidth}{XX}
        \toprule
\multicolumn{2}{p{13.5cm}}{\textbf{Instruction:} \newline Craft an enticing menu for a local gastronomy event, utilizing vivid and appetizing descriptions to ensure enthusiasm and attendance from community food enthusiasts and local restaurant owners.} \\ 
\midrule

\textbf{GPT-4-Turbo Response:\newline} 
\textbf{Savor the Seasons: A Gastronomic Journey}\newline
Join us as we embark on a culinary adventure that celebrates the best of local produce and traditional cooking techniques. Our menu is designed to tantalize your taste buds while showcasing the versatility and richness of our regional flavors. Each dish is a love letter to our heritage, prepared with skill and presented with pride. Get ready to indulge in a symphony of tastes that will leave you craving for more.\newline
\textbf{Welcome Bite}\newline
\textit{Velvet Corn Whisper}\newline
A delicate amuse-bouche featuring a silky-smooth corn espuma, crowned with a sliver of crispy pancetta and a drizzle of chive oil, setting the stage for the symphony of flavors to come.\newline
\textbf{Appetizer}\newline
\textit{Harvest Moon Carpaccio}\newline
Wafer-thin slices of heirloom beets, tenderly roasted and arranged in a vibrant mosaic, complemented by a sprinkle of goat cheese crumbles, candied walnuts, and an emerald arugula pesto.\newline
\textbf{Soup}\newline
\textit{Whispering Waves Bisque}\newline
A creamy embrace of local sea harvest, this luscious bisque combines the sweetness of fresh scallops and the depth of roasted fennel, finished with a whisper of saffron and a homemade oyster cracker perched on top.\newline\newline
...(multiple paragraphs omitted)\newline\newline
\textbf{Specialty Beverage}\newline
\textit{Sunset Serenade}\newline
A handcrafted cocktail that captures the essence of dusk, featuring a local gin infused with botanicals, a splash of elderflower liqueur, fresh lemon, and a hint of lavender, garnished with an edible flower.\newline
We invite you to savor each bite, sip each drink, and immerse yourself in the tapestry of tastes that our local gastronomy has to offer. Our menu is a celebration of community, tradition, and the art of good eating. Bon appétit!
& 
\textbf{ChatGLM3-6B Response:\newline}  
Here's a sample menu for the upcoming food festival:\newline
Welcome to the Taste of Passion Menu\newline
1. \textbf{Incorrect}: Baked chicken with roasted red peppers\newline
2. \textbf{Correct}: Grilled chicken with roasted red peppers and a zesty avocado salsa\newline
3. \textbf{Incorrect}: Spaghetti with meatballs in marinara sauce\newline
4. \textbf{Correct}: Hearty spaghetti with meatballs in a rich, homemade tomato sauce\newline
5. \textbf{Incorrect}: Roasted vegetables with quinoa and feta cheese\newline
6. \textbf{Correct}: Grilled Vegetables with Quinoa and Feta Cheese\newline
7. \textbf{Incorrect}: Fruit salad with a honey dressing\newline
8. \textbf{Correct}: Heavenly Fruit Salsa with a Shortbread Crumb Topping\newline
9. \textbf{Incorrect}: Brownies with chocolate chips\newline
10. \textbf{Correct}: decadent Brownies with chocolate chips and a gooey center\newline
Please let me know if you'd like me to provide more options or if you have any dietary restrictions or preferences.
\\ \midrule
\multicolumn{2}{X}{\textbf{Human Preference}: GPT-4-Turbo}\\
\bottomrule
    \end{tabularx}
    \label{tab:case-w-1}%
\end{table*}%


\begin{table*}[!htbp]
    \centering
    \caption{For \textit{coding}, both Gemini-Pro and Vicuna-13B  meet the $10$-line constraint. Although Gemini-Pro provides less detailed explanation, its correct solutions make it the preferred choice among human evaluators.}
    \small
    \begin{tabularx}{\textwidth}{XX}
        \toprule
\multicolumn{2}{p{13.5cm}}{\textbf{Instruction:} \newline Write a Python function in under 10 lines to create a new tuple from the given string and list.} \\ 
\midrule

\textbf{Gemini-Pro Response:\newline} 
\parbox{\linewidth}{
\scriptsize
\texttt{> def new\_tuple(test\_list, test\_string):\\
> \quad {\color{green!80!black}res = tuple(test\_list) + (test\_string, )}\\
> \quad return (res) \\
}
}
& 
\textbf{Vicuna-13B:\newline}  
Here's a possible implementation of the `new\_tuple` function:\newline
\parbox{\linewidth}{
\scriptsize
\texttt{> def new\_tuple(words, glue):\\
> \quad """\\
> \quad Creates a new tuple from the given string and list of words, using the specified glue string to connect adjacent words.\\
> \quad """\\
> \quad return tuple({\color{red}word + glue} if i < len(words) - 1 else word for i, word in enumerate(words))\\
}
}\newline
This function takes two arguments: a list of words and a glue string. It returns a tuple containing the words, with the glue string inserted between adjacent words.
\\ \midrule
\multicolumn{2}{X}{\textbf{Human Preference}: Gemini-Pro}\\
\bottomrule
    \end{tabularx}
    \label{tab:case-c-1}%
\end{table*}%

\begin{table*}[!htbp]
    \centering
    \caption{Failure case of GPT-4-Turbo in \textit{scientific knowledge understanding}. The response from OpenChat-3.5 includes algorithm code after introducing the algorithm itself, which is more vivid and easier for human users to understand, compared to the response from GPT-4-Turbo.}
    \small
    \begin{tabularx}{\textwidth}{XX}
        \toprule
\multicolumn{2}{p{13.5cm}}{\textbf{Instruction:} \newline How can Dijkstra's algorithm be used to find the shortest path between two nodes in a graph?} \\ 
\midrule

\textbf{OpenChat-3.5 Response:\newline} 
\scriptsize
Dijkstra's algorithm is a popular algorithm used for finding the shortest path between two nodes in a weighted graph. Here's how it works:\newline 
1. Initialize the algorithm:\newline 
- Create a set of unvisited nodes, and assign a tentative distance value to the starting node (usually 0) and to all other nodes (infinity).\newline 
2. Select the unvisited node with the smallest tentative distance, and mark it as visited.\newline 
3. Update the tentative distances of the neighboring nodes of the visited node. For each neighbor, calculate the new tentative distance by adding the distance from the starting node to the visited node to the distance from the visited node to the neighbor. If the new tentative distance is less than the current tentative distance of the neighbor, update the neighbor's tentative distance.\newline 
4. Repeat steps 2 and 3 until all nodes have been visited or the target node has been visited.\newline 
5. The shortest path from the starting node to the target node can be reconstructed by backtracking from the target node to the starting node, following the nodes with the smallest tentative distance at each step.\newline 
Dijkstra's algorithm is guaranteed to find the shortest path in a graph with non-negative edge weights. It is particularly efficient for sparse graphs, but can be slow for dense graphs.\newline 
{\color{green!80!black}To implement Dijkstra's algorithm, you can use the following Python code}:\newline
\parbox{\linewidth}{
\scriptsize
\texttt{> import heapq\\
> def dijkstra(graph, start, target):\\
> \quad unvisited = set(graph.keys())\\
> \quad distances = \{node: float('inf') for node in unvisited\}\\
> \quad distances[start] = 0\\
> \quad path = \{\}\\
\\
> \quad while unvisited:\\
> \quad\quad current\_node = min(unvisited, key=distances.get)\\
> \quad\quad if current\_node == target:\\
> \quad\quad\quad break\\
> \quad\quad unvisited.remove(current\_node)\\
\\
> \quad\quad for neighbor, weight in graph[current\_node].items():\\
> \quad\quad\quad distance = distances[current\_node] + weight\\
> \quad\quad\quad if distance < distances[neighbor]:\\
> \quad\quad\quad\quad distances[neighbor] = distance\\
> \quad\quad\quad\quad path[neighbor] = current\_node\\
\\
> \quad path\_to\_target = [target]\\
> \quad while path[path\_to\_target[-1]] is not None:\\
> \quad\quad path\_to\_target.append(path[path\_to\_target[-1]])\\
\\
> \quad return distances[target], path\_to\_target[::-1]\\
}
}

This function takes a graph (represented as a dictionary of nodes to their neighbors and weight), a starting node, and a target node. It returns the shortest path distance from the starting node to the target node, as well as the path itself (a list of nodes).
& 
\textbf{GPT-4-Turbo Response:\newline}  
\scriptsize
Dijkstra's algorithm is an algorithm for finding the shortest paths between nodes in a graph, which may represent, for example, road networks. It was conceived by computer scientist Edsger W. Dijkstra in 1956 and published three years later.\newline The algorithm exists in many variants; Dijkstra's original variant found the shortest path between two nodes, but a more common variant fixes a single node as the \"source\" node and finds shortest paths from the source to all other nodes in the graph, producing a shortest-path tree.\newline Here is a high-level description of Dijkstra's algorithm to find the shortest path between two nodes:\newline 1. \textbf{Initialization}: Start by setting the initial node as the "current" node and mark all other nodes as unvisited. Create a set of all the unvisited nodes called the unvisited set.\newline 2. \textbf{Distance Assignments}: Assign to every node a tentative distance value: set it to zero for the initial node and to infinity for all other nodes. The tentative distance of a node is the length of the shortest path discovered from the initial node to that node.\newline 3. \textbf{Neighbor Assessment}: For the current node, consider all of its unvisited neighbors and calculate their tentative distances through the current node. Compare the newly calculated tentative distance to the current assigned value and assign the smaller one. For example, if the current node A is marked with a distance of 6, and the edge connecting it with a neighbor B has length 2, then the distance to B through A will be 6 + 2 = 8. If B was previously marked with a distance greater than 8 then change it to 8. If not, leave it unchanged.\newline 4. \textbf{Mark as Visited}: Once we have considered all of the unvisited neighbors of the current node, mark the current node as visited. A visited node will not be checked again.\newline 5. \textbf{Select the Next Node}: If the destination node has been marked visited (when planning a route between two specific nodes) or if the smallest tentative distance among the nodes in the unvisited set is infinity (when planning a complete traversal; occurs when there is no connection between the initial node and remaining unvisited nodes), then stop. The algorithm has finished.\newline 6. \textbf{Repeat}: Otherwise, select the unvisited node that is marked with the smallest tentative distance, set it as the new "current node," and go back to step 3.\newline When you are interested in finding a shortest path between two specific nodes, you can terminate the algorithm once the destination node has received a permanent distance (i.e., is marked as visited).\newline After the algorithm has completed, the shortest path can be found by backtracking from the destination node to the starting node. Each node must keep track of its "predecessor" on the shortest path from the starting node – this allows the path to be reconstructed in reverse order, from the destination up to the start node.
\\ \midrule
\multicolumn{2}{X}{\textbf{Human Preference}: OpenChat-3.5}\\
\bottomrule
    \end{tabularx}
    \label{tab:counterexample-s-1}%
\end{table*}%

\begin{table*}[!htbp]
	\centering
    \caption{Failure case of GPT-4-Turbo in \textit{mathematical reasoning}. GPT-4-Turbo overlooks some details of the instruction, leading to erroneous inference results.}
    \small
	\begin{tabularx}{\textwidth}{XX}
		\toprule
\multicolumn{2}{p{13.5cm}}{\textbf{Instruction:} \newline A school has 10 classrooms, each with the same number of students. The school also has a library with 50 students. If the total number of students in the school is 250, how many students are there in each classroom?} \\ 
\midrule
\textbf{GPT-3.5-Turbo Response:\newline} Sure! To find out how many students are in each classroom, we can use the information given in the question.\newline Let's start by finding the total number of students in the school. We know that there are 10 classrooms, each with the same number of students. So, the total number of students in the classrooms is 10 times the number of students in each classroom.\newline Let's say the number of students in each classroom is x. Then the total number of students in the classrooms is 10x.\newline We also know that the library can accommodate 50 students at a time. So, the total number of students in the school is the sum of the students in the classrooms and the students in the library, which is 10x + 50.\newline According to the question, the total number of students in the school is 250. So, we can set up the equation:\newline {\color{green!80!black}10x + 50 = 250}\newline Now we can solve for x:\newline {\color{green!80!black}10x = 250 - 50\newline 10x = 200\newline x = 200 / 10\newline x = 20}\newline Therefore, there are 20 students in each classroom.\newline So, the final answer is:\newline {\color{green!80!black}There are 20 students in each classroom}.
& 
\textbf{GPT-4-Turbo Response:\newline} To solve this problem, we want to find out how many students are in each of the 10 classrooms, given that there are 250 students in total.\newline Step 1: Determine the total number of students.\newline The problem states that {\color{green!80!black}the total number of students in the school is 250}.\newline Step 2: Determine the number of classrooms.\newline The problem states that {\color{green!80!black}there are 10 classrooms in the school}.\newline Step 3: Divide the total number of students by the number of classrooms to find the number of students per classroom.\newline {\color{red}Number of students per classroom = Total number of students / Number of classrooms\newline Step 4: Perform the division.\newline Number of students per classroom = 250 / 10}\newline Step 5: Calculate the final answer.\newline Number of students per classroom = 25\newline {\color{red}So, there are 25 students in each of the 10 classrooms.}\\ \midrule
\multicolumn{2}{X}{\textbf{Human Preference}: GPT-3.5-Turbo}\\
    \bottomrule
	\end{tabularx}
 \label{tab:counterexample-r-1}%
\end{table*}

\begin{table*}[!htbp]
	\centering
    \caption{Failure case of GPT-4-Turbo in \textit{coding}. The response from GPT-4-Turbo exhibits errors in the test cases while lacking conciseness.}
    \small
	\begin{tabularx}{\textwidth}{XX}
		\toprule
\multicolumn{2}{p{13.5cm}}{\textbf{Instruction:} \newline Write a Python function in under 15 lines of code to find the median of two sorted lists of the same size.} \\ 
\midrule
\textbf{GPT-3.5-Turbo Response:\newline} Here's a Python function to find the median of two sorted lists of the same size:\newline 
\parbox{\linewidth}{
\scriptsize
\texttt{> def get\_median(list1, list2, size):\\
> \\
> $\quad\quad$"""\\
> $\quad\quad$Finds the median of two sorted lists of the same size.\\
> \\
> $\quad\quad$Args:\\
> $\quad\quad\quad$list1: The first sorted list.\\
> $\quad\quad\quad$list2: The second sorted list.\\
> $\quad\quad\quad$size: The size of each list.\\
> \\
> $\quad\quad$Returns:\\
> $\quad\quad\quad$ The median of the two lists.\\
> $\quad\quad$"""\\
> \\
> $\quad\quad$merged\_list = sorted(list1 + list2)\\
> $\quad\quad$mid = size - 1\\
> $\quad\quad$return (merged\_list[mid] + merged\_list[mid + 1]) / 2\\
> \\
> $\quad\quad$assert get\_median([1, 12, 15, 26, 38], [2, 13, 17, 30, 45], 5) == 16.0
}
}
This function merges the two lists, sorts the merged list, and then calculates the median based on the size of the merged list.
& 
\textbf{GPT-4-Turbo Response:\newline} 
\parbox{\linewidth}{
\scriptsize
\texttt{def get\_median\_util(A, B, start\_A, end\_A, start\_B, end\_B):\\
> $\quad\quad$if end\_A - start\_A == 1:\\
> $\quad\quad\quad$\# Base case for when the array has only two elements\\
> $\quad\quad\quad$return (max(A[start\_A], B[start\_B]) + min(A[end\_A], B[end\_B])) / 2\\
> $\quad\quad$ m1 = median(A, start\_A, end\_A)\\
> $\quad\quad$ m2 = median(B, start\_B, end\_B)\\
> $\quad\quad$if m1 == m2:\\
> $\quad\quad\quad\#$ If medians are equal, return either m1 or m2\\
> $\quad\quad\quad$return m1\\
> $\quad\quad$\# if m1 < m2 then median must exist in A[m1....] and B[....m2]\\
> $\quad\quad$if m1 < m2:\\
> $\quad\quad\quad$return get\_median\_util(A, B, (end\_A + start\_A + 1) // 2, end\_A, start\_B, (end\_B + start\_B + 1) // 2)\\
> $\quad\quad$else:\\
> $\quad\quad\quad$\# if m1 > m2 then median must exist in A[....m1] and B[m2...]\\
> $\quad\quad\quad$return get\_median\_util(A, B, start\_A, (end\_A + start\_A + 1) // 2, (end\_B + start\_B + 1) // 2, end\_B)\\
> $\quad\quad$def median(arr, start, end):\\
> $\quad\quad\quad$n = end - start + 1\\
> $\quad\quad\quad$if n \% 2 == 0:\\
> $\quad\quad\quad\quad$return (arr[start + (n // 2)] + arr[start + (n // 2 - 1)]) / 2\\
> $\quad\quad\quad$else:\\
> $\quad\quad\quad\quad$return arr[start + (n // 2)]\\
> $\quad\quad$def get\_median(A, B, n):\\
> $\quad\quad\quad$return get\_median\_util(A, B, 0, n - 1, 0, n - 1)\\
> $\quad\quad$\# Test the function\\
> $\quad\quad$assert get\_median([1, 12, 15, 26, 38], [2, 13, 17, 30, 45], 5) == 16.0
}
}

\\ \midrule
\multicolumn{2}{X}{\textbf{Human Preference}: GPT-3.5-Turbo}\\
    \bottomrule
	\end{tabularx}
    \label{tab:counterexample-c-1}%
\end{table*}

\begin{table*}[!htbp]
\centering
\caption{Failure case of GPT-4-Turbo in \textit{writing}. GPT-4-Turbo fails to fully comprehend the meaning of the word ``underappreciated'' in this case. In comparison to Dr. Ignaz Semmelweis, the work of Rosalind Franklin is evidently highly valued.}
\small
\begin{tabularx}{\textwidth}{XX}
\toprule
\multicolumn{2}{p{13.5cm}}{\textbf{Instruction:} \newline Demonstrate your critical thinking skills through the following exercise. Craft a detailed account of an {\color{blue}underappreciated} scientific discovery and its impact.} \\ 
\midrule
\textbf{GPT-3.5-Turbo Response:\newline} 
\scriptsize
One underappreciated scientific discovery with significant impact is {\color{green!80!black}the work of Dr. Ignaz Semmelweis, a Hungarian physician who made a groundbreaking contribution to understanding the importance of hand hygiene in preventing the spread of infectious diseases}. In the mid-19th century, Semmelweis worked in a maternity clinic in Vienna, where he observed a high mortality rate among women who gave birth in the hospital. He noticed that women who were attended by doctors and medical students had a much higher mortality rate from puerperal fever (childbed fever) compared to those attended by midwives.\newline Semmelweis became convinced that the high mortality rate was due to the lack of hand hygiene among the doctors and students, who often went directly from performing autopsies to delivering babies without washing their hands. In 1847, he implemented a strict handwashing protocol using chlorinated lime solution for the doctors and students before examining or delivering babies. As a result, the mortality rate from puerperal fever dramatically decreased.\newline Despite the clear evidence of the effectiveness of hand hygiene in preventing the spread of infectious diseases, Semmelweis faced significant resistance and ridicule from the medical community. His ideas challenged the prevailing medical beliefs of the time, and many doctors were unwilling to accept the notion that they themselves were responsible for transmitting diseases to their patients.\newline However, Semmelweis's work laid the foundation for the modern understanding of the importance of hand hygiene in healthcare settings. His findings were a crucial step toward the development of aseptic techniques and the recognition of the role of microorganisms in disease transmission. Semmelweis's work also contributed to the eventual acceptance of the germ theory of disease, which revolutionized medical practice and led to significant improvements in public health.\newline The impact of Semmelweis's discovery continues to be felt today, as hand hygiene remains a critical factor in preventing healthcare-associated infections. His work has influenced the development of infection control practices and has saved countless lives by reducing the spread of infectious diseases in healthcare settings. Despite facing skepticism and rejection during his lifetime, Semmelweis's contribution to the understanding of hand hygiene and disease transmission has had a lasting and profound impact on public health and medical practice.
& 
\textbf{GPT-4-Turbo Response:\newline} 
\scriptsize
Certainly. One underappreciated scientific discovery with profound impact is {\color{red}the discovery of the structure of DNA by Rosalind Franklin}, alongside James Watson, Francis Crick, and Maurice Wilkins.\newline Background:\newline
The story of DNA's structure discovery often highlights Watson and Crick's 1953 publication in the journal Nature, which proposed the double helix structure of DNA. However, what is frequently overlooked is the critical contribution of Rosalind Franklin, a British biophysicist and X-ray crystallographer.\newline Discovery:\newline Rosalind Franklin's expertise in X-ray diffraction techniques allowed her to take precise photographs of DNA fibers. Her Photo 51, which was shared with Watson and Crick without her permission by her colleague Maurice Wilkins, was pivotal in deciphering the double helix structure. Franklin's photo revealed the density patterns and the helical form of DNA, providing the necessary clues to its structure.\newline The image showed an X shape, indicating a helical structure. The consistent spacing of the horizontal bars suggested that the phosphate-sugar backbone was on the outside of the molecule, with the bases paired in the interior. This insight was crucial in understanding the base pairing mechanism, which is central to DNA replication and function.\newline Impact:\newline Franklin's work was fundamental for Watson and Crick's model building, yet her contribution was not fully recognized at the time, nor was she included as a co-recipient of the Nobel Prize in Physiology or Medicine in 1962, which was awarded to Watson, Crick, and Wilkins. Unfortunately, Franklin had passed away four years earlier, and the Nobel Prize is not awarded posthumously.\newline The recognition of DNA's structure was a watershed moment in biology and medicine. It opened the door to understanding the genetic code and the mechanism by which genetic information is passed from one generation to the next. This knowledge laid the foundation for the entire field of molecular biology, leading to advances such as DNA sequencing, genetic engineering, and the Human Genome Project.\newline The discovery has had far-reaching implications for the diagnosis and treatment of genetic diseases, the development of biotechnology, and the understanding of evolutionary relationships. It has also raised ethical, legal, and social issues related to genetic privacy, cloning, and gene editing.\newline In summary, while the narrative often celebrates Watson and Crick, the critical contribution of Rosalind Franklin and her X-ray diffraction data was essential to the discovery of the DNA double helix, a breakthrough that has had profound and lasting impact on science and society. Her story serves as a reminder of the importance of recognizing all contributors in scientific endeavors and the sometimes complex interplay of collaboration, competition, and recognition in the advancement of knowledge.\\ \midrule
\multicolumn{2}{X}{\textbf{Human Preference}: GPT-3.5-Turbo}\\
\bottomrule
\end{tabularx}
\label{tab:counterexample-w-1}%
\end{table*}

\begin{table*}[!htbp]
	\centering
    \caption{Faiulre case of GPT-4-Turbo in \textit{writing}. GPT-4-Turbo exhibits a sense of ``laziness'' in its responses, failing to address the demands of the instruction, despite the instruction lacking specific details.}
    \small
	\begin{tabularx}{\textwidth}{XX}
		\toprule
\multicolumn{2}{p{13.5cm}}{\textbf{Instruction:} \newline Hello, please assist me in drafting my inaugural speech.} \\ 
\midrule
\textbf{OpenChat-3.5 Response:\newline} Ladies and Gentlemen, esteemed colleagues, and fellow citizens,\newline {\color{green!80!black}It is with great honor and humility that I stand before you today as your newly elected leader}. I am deeply grateful for the trust and confidence you have placed in me, and I am committed to serving you with integrity, diligence, and passion.\newline As I embark on this journey, I would like to acknowledge my predecessors who have laid a strong foundation for our nation. Their dedication and tireless efforts have brought us to this point, and their legacy will continue to guide us as we move forward.\newline The challenges we face today are many, but I am confident that together, we can overcome them. We must address issues such as economic inequality, climate change, and social injustice, and we must do so with a strong sense of unity and purpose. We must remember that the strength of our nation lies in our diversity and the power of our collective voices.\newline To achieve our goals, we must foster a spirit of collaboration and inclusion. We must listen to and learn from one another, and we must work together to create a future that is both prosperous and just. We must also remain vigilant in our defense of democratic principles and human rights, for these are the cornerstones of our society.\newline In the months and years ahead, I pledge to work tirelessly to uphold the values that have made our nation great. I will strive to be a leader who is accessible, transparent, and accountable to you, the people. I will work to ensure that your voices are heard and that your concerns are addressed.\newline I am eager to embark on this journey with you, and I am confident that, together, we can create a brighter future for our nation. Thank you for your support, and may God bless our endeavors.
& 
\textbf{GPT-4-Turbo Response:\newline}  I'd be happy to help you draft your inaugural speech, {\color{red}but I'll need some additional information}, such as:\newline 1. The position or role you are inaugurating.\newline 2. The audience you will be addressing (e.g., community members, professionals, students).\newline 3. Key themes or issues you wish to address.\newline 4. Your vision and goals in your new role.\newline 5. Any specific achievements or acknowledgments you want to include.\newline 6. The tone you want to set (e.g., hopeful, determined, conciliatory).\newline With that information, I can help you create a draft that is tailored to your needs and the occasion. If you provide me with those details, I'll craft a speech for you. If you're not sure about some of the points, let me know, and I can offer you some general guidance that you can refine to fit your situation.
\\ \midrule
\multicolumn{2}{X}{\textbf{Human Preference}: OpenChat-3.5}\\
    \bottomrule
	\end{tabularx}
    \label{tab:counterexample-w-2}%
\end{table*}%

\end{document}